\documentclass{article}

\usepackage[preprint]{neurips_2026}

\usepackage[utf8]{inputenc} 
\usepackage[T1]{fontenc}    
\usepackage{hyperref}       
\usepackage{url}            
\usepackage{booktabs}       
\usepackage{amsfonts}       
\usepackage{nicefrac}       
\usepackage{microtype}      
\usepackage{xcolor}         
\usepackage{graphicx}
\usepackage{amsmath, amssymb, amsfonts}
\usepackage{amsthm} 
\usepackage{mathtools}
\usepackage{algorithm}
\usepackage{algorithmic}
\usepackage{booktabs}
\usepackage{multirow}
\usepackage{tabularx}
\usepackage{csquotes}
\usepackage{adjustbox}
\usepackage{subcaption} 
\usepackage{bm}
\usepackage{xcolor}
\usepackage{hyperref}
\usepackage{pgfplotstable}
\pgfplotsset{compat=1.18}
\usepackage{xfp}
\usepackage{siunitx}
\usepackage{placeins}
\usepackage{pdflscape}

\theoremstyle{plain}

\theoremstyle{definition}

\theoremstyle{remark}

\usepackage{xfp}

\newcolumntype{C}{>{\centering\arraybackslash}p{2.8cm}}

\title{Sampling from Neural Operator Priors to Solve Partial Differential Equations}

\author{%
  Andrew Millard \thanks{Lead Author, correspondance to \texttt{andmi73@liu.se}.} \thanks{Department of Computer and Information Science, Linköping University, Sweden.}\\
  \And 
  Joshua Murphy \thanks{Department of Political Science, Trinity College Dublin, Ireland.}
  \And
  Zheng Zhao$^\dagger$ \\
}

\begin{document}

\maketitle

\begin{abstract}
    Guided sampling with sparse observations and PDE residuals has recently been combined with neural operator priors to solve partial differential equations in function space, their natural setting. We pair such a prior with pseudo-bootstrap sequential Monte Carlo (pBS SMC), a guided sampling framework that accommodates proposals without tractable densities, and introduce a new proposal by adapting the DPM++ algorithm to the guidance setting. Across five benchmark PDE systems this proposal attains the best error--time trade-off at every step count tested, and wrapping it in pBS gives the lowest error on most inverse-problem settings, with the largest gains when few sampling steps are used.
\end{abstract}


\section{Introduction and Related Work}
\label{sec:introduction}

Classical numerical solvers for partial differential equations \citep[PDEs,][]{evans2022partial, leveque2007finite} scale poorly in dimension, parameter count, and nonlinearity. Machine learning offers a faster alternative: PINNs \citep{raissi2019physics} and Physics-Informed Neural Operators \citep[PINO,][]{li2024physics} penalise PDE residuals during training, while operator models \citep{li2021fourier, lu2019deeponet} learn mappings between coefficient and solution fields. Purely data-driven generative models \citep{yang2023denoising, kohl2023benchmarking} instead learn PDE structure from data alone, imposing no PDE-specific constraint and so giving no guarantee of physically admissible solutions. A third option combines the two: pretrain a diffusion model \citep{song2021scorebasedgenerativemodelingstochastic} on PDE data as a prior, then use residuals and observations at inference time to guide sampling \citep{zhao2025conditional, shysheya2024-on-PDE-DA}. Diffusion Posterior Sampling \citep[DPS,][]{chung2022diffusion} guides on observations alone; \citet{huang2024diffusionpdegenerativepdesolvingpartial} additionally incorporate the PDE residual.

\citet{millard2026particleguideddiffusionmodelspartial} embed guidance within a Sequential Monte Carlo \citep[SMC,][]{del2006sequential} framework. Their best-performing proposal, a second-order stochastic integrator (SOSaG), admits no tractable density, precluding standard SMC weights. They therefore introduce a pseudo-bootstrap (pBS) variant, re-weighting particles by the ratio of approximate likelihoods at consecutive steps, giving a target related to but distinct from the diffusion posterior recovered by exact schemes such as the Twisted Diffusion Sampler \citep{wu2023-TDS}.

Both approaches train on a fixed-resolution grid, whereas PDE solutions are better described as elements of a function space. \citet{yao2026guided} therefore replace the finite-dimensional diffusion model with a U-shaped Neural Operator \citep[UNO,][]{rahman2023uno} prior acting directly on function space, generalising Tweedie's formula accordingly, and report lower error in fewer sampling steps. The two target different bottlenecks: pBS corrects point-estimate guidance bias via particles even without a tractable proposal density, while the UNO prior ensures the model itself respects function-space structure. We pair them. 

Our contributions are threefold. We combine pBS guided sampling with a UNO diffusion prior, solving PDE inverse problems via function-space, particle-guided diffusion. We then modify DPM++ \citep{Lu_2025}, a second-order solver that recycles previous-step estimates rather than re-evaluating them, to accept guidance (G-DPM++); its proposal density is likewise intractable, so pBS is again essential to its use within a particle-guided sampler. Finally, across PDE benchmarks, G-DPM++ achieves a better error-versus-time trade-off than SOSaG and the Heun integrator of \citet{yao2026guided}, and pBS reduces error for guided proposals, with gains most pronounced in the small-sampling-step limit.
\section{Problem Formulation and Background}

\paragraph{PDEs and Neural Operators.} \label{sec:pde_background}

We consider both static and time-dependent PDEs. In the static case, for a spatial domain $\Omega \subset \mathbb{R}^d$, a PDE is written $f(c, a, u) = 0$ for $c \in \Omega$, subject to $u(c) = g(c)$ on $\partial\Omega$, where $c$ is a spatial coordinate, $u \in \mathcal{U} \subset H^1(\Omega)$ is the solution field, and $a \in \mathcal{A} \subset L^2(\Omega)$ is a coefficient field parameterising the PDE. This implicitly defines a solution operator $\mathcal{G} : \mathcal{A} \to \mathcal{U}$, infinite-dimensional in general. The time-dependent extension, in which $\mathcal{G}$ maps an initial condition $a_0$ to a space-time trajectory in $L^2([0,\mathcal{T}]; \mathcal{U})$, is given in Appendix~\ref{app:pde_details}; both Darcy flow and the incompressible Navier--Stokes equations are instances. Throughout, $\tau$ denotes PDE time and $\mathcal{T}$ the PDE horizon, distinct from the diffusion time $t$ and horizon $T$; discrete diffusion steps are indexed by $k \in \{0, 1, \ldots, K\}$. Our goal is to recover $a$ (or $a_0$) and the corresponding solution from sparse observations.

Neural operators \citep{JMLR:v24:21-1524} approximate $\mathcal{G}$ by a parametric map defined at the function level rather than on a fixed grid, providing \emph{discretisation invariance}: the same trained model can be evaluated on any discretisation of $\Omega$ without retraining. We use the U-shaped Neural Operator \citep[UNO;][]{rahman2023uno}, a multi-resolution architecture building on the Fourier Neural Operator \citep[FNO;][]{li2021fourier}, as the backbone denoiser of a function-space diffusion model, a combination we refer to as the \emph{neural operator diffusion prior}.

\paragraph{Guided Sampling in Function Space.} \label{sec:diffusion_background}

We consider diffusion models \citep{song2021scorebasedgenerativemodelingstochastic} in an infinite-dimensional setting on the separable Hilbert space $\mathcal{H} = L^2(\Omega) \times L^2(\Omega)$, with joint state $x = (a, u) \in \mathcal{H}$, following the EDM formulation \citep{karras2022elucidatingdesignspacediffusionbased}. The data distribution is corrupted by Gaussian noise with standard deviation $\sigma_t \in [0, \sigma_{\text{max}}]$, giving the reference measure $p_{\text{ref}}(x) \coloneqq \mathcal{N}(0, \sigma_{\text{max}}^2 \mathcal{C})$, where $\mathcal{C}: \mathcal{H} \to \mathcal{H}$ is a symmetric, positive-definite, trace-class covariance operator (Appendix~\ref{app:fn_space_remarks}). The score is estimated via a neural operator denoiser $\mathcal{D}_\theta$ through $\nabla_{x_t} \log p(x_t, \sigma_t) = \sigma_t^{-2}(x_t - \mathcal{D}_\theta(x_t, \sigma_t))$.

We solve PDEs using a diffusion model pretrained on the joint distribution of coefficients and solutions~\citep{huang2024diffusionpdegenerativepdesolvingpartial}, with PDE residuals and sparse observations $a_{\text{obs}}, u_{\text{obs}}$ supplied as a condition $y$ during inference. Following \citet{huang2024diffusionpdegenerativepdesolvingpartial, millard2026particleguideddiffusionmodelspartial, mammadovdiffusion}, we take $p(y \mid x)$ to be a Gaussian \emph{PDE residual likelihood},
\begin{equation}
    \log p(y \mid x) = - \tfrac{\beta}{n} \|u_{\text{obs}} - \mathcal{M}_u \odot x\|^2_2 - \tfrac{\gamma}{n} \|a_{\text{obs}} - \mathcal{M}_a \odot x\|^2_2 - \tfrac{\omega}{m} \|f(c, \tau, x)\|^2_2 + \text{const},
    \label{eq:likelihoods_expanded}
\end{equation}
where $\mathcal{M}_{u}, \mathcal{M}_{a}$ are binary observation masks, $n$ the number of observations, $m$ the number of grid points, and $\beta, \gamma, \omega > 0$ tunable weights. Guided sampling then draws $x_T \sim p_{\text{ref}}$ and simulates the conditional reverse-time SDE
\begin{equation}
    dx_t = -2\dot{\sigma}_t \sigma_t \big[\nabla_{x_t} \log p_\theta(x_t, \sigma_t) + \nabla_{x_t} \log p_\theta(y \mid x_t, \sigma_t)\big]\,dt + \sqrt{2 \dot{\sigma}_t \sigma_t} \, dW_t^{\mathcal{H}},
    \label{eq:guidance_sde_noise}
\end{equation}
where $W_t^{\mathcal{H}}$ is an $\mathcal{H}$-valued Wiener process with covariance $\mathcal{C}$, so that at $t=0$ samples follow $p_\theta(x \mid y) \propto p(y \mid x)\, p_\theta(x)$. The guidance term requires the intermediate likelihood $p_\theta(y \mid x_t, \sigma_t) = \int p(y \mid x_0)\, p_\theta(x_0 \mid x_t)\, dx_0$, which is intractable; following \citet{wu2023-TDS} we approximate it at the denoiser's reconstruction, $\tilde{p}_\theta(y \mid x_t, \sigma_t) := p(y \mid x_0 = \mathcal{D}_\theta(x_t, \sigma_t))$, which is exact at $t=0$. We use SMC to correct this approximation.

Discretising~\eqref{eq:guidance_sde_noise} with Euler--Maruyama yields the guided Euler--Maruyama (GEM) update,
\begin{equation}
    x_{k-1} = x_k - (\sigma_{k-1}^2 - \sigma_k^2)\frac{x_k - \mathcal{D}_\theta(x_k, \sigma_k)}{\sigma_k^2} - (\sigma_{k-1}^2 - \sigma_k^2)\, \nabla_{x_k} \log \tilde{p}_\theta(y \mid x_k, \sigma_k) + \sqrt{\sigma_{k-1}^2 - \sigma_k^2}\, z,
    \label{eq:EM_denoiser_guided_constant}
\end{equation}
with $z \sim \mathcal{N}(0, \mathcal{C})$. We write $q^{\textsc{em}}(x_{k-1} \mid x_k)$ for the unguided transition and $q_\theta^{\textsc{gem}}(x_{k-1} \mid x_k, y)$ for its guided counterpart. A complete explanation of GEM is given in Appendix~\ref{app:pseudocodes_gem}. 

\paragraph{Sequential Monte Carlo.}

SMC methods~\citep[e.g.,][]{chopin2020introduction} sample from a sequence of distributions using $N$ particles $\{x_k^{(i)}\}_{i=1}^N$ in $\mathcal{H}$, and have been used as conditional samplers for diffusion models \citep{wu2023-TDS, cardoso2023monte, dou2024diffusion}. Particles are propagated through Markov kernels $\{M_{k-1}(\cdot \mid x_k)\}_{k=1}^K$ (with $M_K = p_{\text{ref}}$), reweighted by $w_{k-1}^{(i)} \propto w_k^{(i)}\, G_{k-1}(x_k^{(i)}, x_{k-1}^{(i)})$ for potentials $G$, and resampled \citep{douc2005comparison} when the effective sample size falls below a threshold, so that the weighted population approximates a Feynman--Kac target $\nu_k$ (Appendix~\ref{app:smc_pseudocode}). We choose $\{M_k, G_k\}$ so that $\nu_0(x_0)$ targets the posterior; the estimator is then consistent as $N \to \infty$ regardless of the intermediate marginals. Taking $q_\theta^{\textsc{gem}}$ as the proposal together with the weight update
\begin{equation}
    G_{k-1}(x_k, x_{k-1}) = \frac{\tilde{p}_\theta(y \mid x_{k-1}, \sigma_{k-1})}{\tilde{p}_\theta(y \mid x_k, \sigma_k)}\, \frac{q^{\textsc{em}}(x_{k-1} \mid x_k)}{q_\theta^{\textsc{gem}}(x_{k-1} \mid x_k, y)},
    \label{eq:weight_update_full}
\end{equation}
recovers the Twisted Diffusion Sampler~\citep[TDS;][]{wu2023-TDS}, with $\nu_0(x_0) = q^{\textsc{em}}(x_0 \mid y)$ (Appendix~\ref{app:twisted_intermediate}).
\section{Methodology} \label{sec:sto_samplers}

\paragraph{Second-Order Integrators.} \citet{millard2026particleguideddiffusionmodelspartial} extend the EDM stochastic sampler of \citet{karras2022elucidatingdesignspacediffusionbased} with a guidance term. At each step, noise is injected to perturb the state from noise level $\sigma_k$ to $\hat\sigma_k = (1 + \chi_k)\sigma_k$, where $\chi_k \geq 0$ is a per-step churn parameter, and a Heun correction is applied for second-order accuracy. The full update is given in Appendix~\ref{app:pseudocodes_sosag}.

We instead adapt the multistep DPM++(2M) solver of \citet{Lu_2025} to the guidance setting. The method operates in the log signal-to-noise ratio domain $\lambda_k = -\log \sigma_k$, combining the current and previous denoised estimates for second-order accuracy. Letting $h_k = \lambda_{k-1} - \lambda_k$ denote the log-SNR step size and $\hat{x}_0 = \mathcal{D}_\theta(x_k, \sigma_k)$, the update is
\begin{align}
    x_{k-1} &= \underbrace{\frac{\sigma_{k-1}}{\sigma_k} e^{-h_k}\, x_k + \bigl(1 - e^{-2h_k}\bigr)\, \hat{x}_0}_{\text{1st-order DPM++ update}}
    \;+\; \underbrace{\frac{1 - e^{-2h_k}}{2\, r_k}\bigl(\hat{x}_0 - \hat{x}_0^{\text{prev}}\bigr)}_{\text{2nd-order correction}} \nonumber \\
    &\quad+\; \underbrace{\sigma_{k-1}\sqrt{1 - e^{-2h_k}}\;\epsilon_k}_{\text{stochastic term}}
    \;+\; \underbrace{\zeta_k\, \nabla_{x_k} \log \tilde{p}_\theta(y \mid x_k, \sigma_k)}_{\text{guidance term}},
    \label{eq:gdpmpp_update_app}
\end{align}
where $\epsilon_k \sim \mathcal{N}(0, \mathcal{C})$, $\hat{x}_0^{\text{prev}}$ is the denoised estimate from the previous step, $r_k = |\lambda_k^{\text{prev}} - \lambda_k| / h_k$, and $\zeta_k$ is a user-defined guidance schedule. The second-order correction is omitted at the first step, where no previous estimate is available, and at the final step for stability as $\sigma \to 0$. Because G-DPM++ reuses the cached estimate rather than evaluating the denoiser twice per step as in Heun's method, it is comparable in cost per step to first-order GEM while achieving second-order accuracy.\footnote{The deterministic form of DPM-Solver++(2M) has local error $O(h^3)$~\citep{Lu_2025}. We expect the stochastic variant to retain this order in the corresponding \emph{weak} sense; a formal weak-error analysis is left to future work.} Pseudocode and the VE specialisation of the DPM++ backbone are given in Appendix~\ref{app:pseudocodes_g-dpmpp}.

\paragraph{Pseudo-Bootstrap.} \label{sec:pBS}

The TDS weight update of \eqref{eq:weight_update_full} requires evaluating the proposal density in closed form, which fails for both samplers: SOSaG combines two denoiser evaluations nonlinearly via the Heun correction, while G-DPM++ depends on a cached previous estimate. \citet{millard2026particleguideddiffusionmodelspartial} proposed dropping the proposal-density ratio, yielding a bootstrap-style scheme \citep{gordon1993novel}:
\begin{equation*}
    \overline{M}_{k-1}(x_{k-1} \mid x_k) = \overline{p}_\theta(x_{k-1} \mid x_k, y), \qquad
    \overline{G}_{k-1}(x_k, x_{k-1}) = \frac{\tilde{p}_\theta(y \mid x_{k-1}, \sigma_{k-1})}{\tilde{p}_\theta(y \mid x_k, \sigma_k)},
\end{equation*}
where $\overline{p}_\theta(\cdot \mid x_k, y)$ is the intractable transition kernel induced by the chosen guided sampler. This admits the target $\overline{\nu}_0(x_0) \propto \overline{p}_\theta(x_0 \mid y)\, p_\theta(y \mid x_0)$, where $\overline{p}_\theta(x_0 \mid y)$ is the marginal at $k=0$ from simulating the guided sampler without SMC correction (Appendix~\ref{app:pBS_explained}).

This target appears at first glance to be biased relative to the true posterior, and \citet{millard2026particleguideddiffusionmodelspartial} justified the scheme empirically, on the grounds that the many-try feature of SMC combined with sophisticated proposals gave better results in practice. 

\section{Experimental Set Up}

\begin{table}[!t]
  \centering
  \tiny
  \setlength{\tabcolsep}{3pt}
  \caption{%
    Relative $L_2$ error ($\mathcal{E}$) and efficiency ($\mathcal{E}{\times}t$) across all PDE
    benchmarks, where $t$ is wall-clock time in seconds. \textsc{+pBS} columns give the error of
    each proposal wrapped in the pBS SMC framework. Best $\mathcal{E}$ and best $\mathcal{E}{\times}t$
    per row in \textbf{bold}.
    $^\dagger$Darcy Flow reports error rate rather than relative $L_2$ error.
  }
  \label{tab:combined}

  \newcommand{\pdeblock}[2]{%
    \begin{tabular}{l cc ccc ccc}
      \multicolumn{9}{c}{\textit{#1}} \\
      \toprule
      \multirow{2}{*}{$K$}
        & \multicolumn{2}{c}{DPS} & \multicolumn{3}{c}{SOSaG} & \multicolumn{3}{c}{G-DPM++} \\
      \cmidrule(lr){2-3} \cmidrule(lr){4-6} \cmidrule(lr){7-9}
      & $\mathcal{E}$ & $\mathcal{E}{\times}t$
      & $\mathcal{E}$ & $\mathcal{E}{\times}t$ & \textsc{+pBS}
      & $\mathcal{E}$ & $\mathcal{E}{\times}t$ & \textsc{+pBS} \\
      \midrule
      #2
      \bottomrule
    \end{tabular}%
  }

  \newcommand{\pdeblocknok}[2]{%
    \begin{tabular}{cc ccc ccc}
      \multicolumn{8}{c}{\textit{#1}} \\
      \toprule
      \multicolumn{2}{c}{DPS} & \multicolumn{3}{c}{SOSaG} & \multicolumn{3}{c}{G-DPM++} \\
      \cmidrule(lr){1-2} \cmidrule(lr){3-5} \cmidrule(lr){6-8}
      $\mathcal{E}$ & $\mathcal{E}{\times}t$
      & $\mathcal{E}$ & $\mathcal{E}{\times}t$ & \textsc{+pBS}
      & $\mathcal{E}$ & $\mathcal{E}{\times}t$ & \textsc{+pBS} \\
      \midrule
      #2
      \bottomrule
    \end{tabular}%
  }

  \pdeblock{Darcy Flow$^\dagger$}{
    25  & 0.1918 & 1.5881 & 0.1850 & 1.3234 & 0.1527 & 0.1588 & \textbf{0.6706} & \textbf{0.1203} \\
    50  & 0.1252 & 2.1180 & 0.1333 & 1.8811 & 0.1071 & 0.1196 & \textbf{1.0048} & \textbf{0.0937} \\
    100 & 0.0949 & 3.2017 & 0.1026 & 2.8351 & 0.0836 & 0.0926 & \textbf{1.5665} & \textbf{0.0770} \\
    150 & 0.0812 & 4.0898 & 0.0892 & 3.7497 & 0.0735 & 0.0807 & \textbf{2.0726} & \textbf{0.0716} \\
    200 & 0.0739 & 4.9831 & 0.0815 & 4.5865 & \textbf{0.0679} & 0.0736 & \textbf{2.4739} & 0.0721 \\
    500 & 0.0588 & 9.9581 & 0.0611 & 8.6949 & 0.0585 & \textbf{0.0570} & \textbf{4.8338} & \textbf{0.0570} \\
  }
  \hfill
  \pdeblocknok{Helmholtz}{
    0.4301 & 3.9446 & 0.5050 & 3.7137 & 0.4514 & 0.3878 & \textbf{1.8240} & \textbf{0.3646} \\
    0.3478 & 6.5391 & 0.3935 & 5.9539 & 0.3655 & 0.3306 & \textbf{3.1795} & \textbf{0.3209} \\
    0.2883 & 11.0421 & 0.3206 & 9.6462 & 0.3086 & 0.2823 & \textbf{5.4365} & \textbf{0.2747} \\
    0.2563 & 14.6694 & 0.2856 & 12.9472 & 0.2779 & 0.2573 & \textbf{7.5582} & \textbf{0.2544} \\
    0.2374 & 18.1976 & 0.2654 & 16.1422 & 0.2592 & 0.2420 & \textbf{9.2168} & \textbf{0.2363} \\
    0.2013 & 38.1182 & 0.2154 & 32.9309 & 0.2153 & \textbf{0.2012} & \textbf{19.4075} & 0.2039 \\
  }

  \pdeblock{NS (Bounded)}{
    25  & 0.2887 & 2.6848 & 0.3489 & 2.5966 & 0.1828 & 0.2765 & \textbf{1.3050} & \textbf{0.1498} \\
    50  & 0.1889 & 3.5777 & 0.2466 & 3.7219 & 0.1135 & 0.1813 & \textbf{1.7117} & \textbf{0.0935} \\
    100 & 0.1024 & 3.9109 & 0.1590 & 4.7486 & 0.0842 & 0.1010 & \textbf{1.9314} & \textbf{0.0498} \\
    150 & 0.0681 & 3.8924 & 0.1057 & 4.7945 & 0.0671 & 0.0651 & \textbf{1.8973} & \textbf{0.0379} \\
    200 & 0.0496 & 3.7933 & 0.0813 & 4.9390 & 0.0605 & 0.0474 & \textbf{1.8252} & \textbf{0.0336} \\
    500 & 0.0376 & 7.2684 & 0.0372 & 5.6697 & 0.0465 & 0.0312 & \textbf{2.9781} & \textbf{0.0286} \\
  }
  \hfill
  \pdeblocknok{NS (Unbounded)}{
    0.1500 & 1.3837 & 0.1744 & 1.3028 & 0.1584 & 0.1463 & \textbf{0.6907} & \textbf{0.1429} \\
    0.1295 & 2.4044 & 0.1427 & 2.1490 & 0.1365 & 0.1249 & \textbf{1.2047} & \textbf{0.1233} \\
    0.1157 & 4.3906 & 0.1256 & 3.8191 & 0.1215 & \textbf{0.1145} & \textbf{2.2238} & 0.1147 \\
    0.1090 & 6.2443 & 0.1170 & 5.3048 & 0.1141 & 0.1097 & \textbf{3.1428} & \textbf{0.1089} \\
    \textbf{0.1045} & 7.9849 & 0.1115 & 6.7496 & 0.1097 & 0.1062 & \textbf{4.0939} & 0.1056 \\
    \textbf{0.0900} & 17.1192 & 0.0999 & 15.2329 & 0.0993 & 0.0961 & \textbf{9.2735} & 0.0963 \\
  }

  \pdeblock{Poisson}{
    25  & 0.4175 & 3.4347 & 0.4694 & 3.2026 & 0.3927 & 0.4192 & \textbf{1.7916} & \textbf{0.3788} \\
    50  & 0.3572 & 6.0461 & 0.4175 & 5.7597 & 0.3541 & 0.3717 & \textbf{3.1136} & \textbf{0.3394} \\
    100 & 0.3084 & 10.4387 & 0.3746 & 10.4184 & 0.3277 & 0.3226 & \textbf{5.5649} & \textbf{0.2996} \\
    150 & \textbf{0.2858} & 14.2541 & 0.3448 & 14.4339 & 0.3131 & 0.2996 & \textbf{7.6062} & 0.2859 \\
    200 & 0.2712 & 18.0289 & 0.3275 & 18.4473 & 0.2993 & 0.2855 & \textbf{9.5098} & \textbf{0.2711} \\
    500 & \textbf{0.2365} & 40.3794 & 0.2792 & 39.2457 & 0.2618 & 0.2452 & \textbf{20.7547} & 0.2375 \\
  }
\end{table}

We test our methods on five benchmark PDEs for inverse problems: Darcy flow (DF), the inhomogeneous Helmholtz equation (IHE), Navier--Stokes in bounded and unbounded domains (NS-Bounded, NS-Unbounded), and the Poisson equation (Appendix~\ref{app:pde_details}). Our implementation builds on the open-source code of \citet{yao2026guided}, using the pretrained models they provide, and we compare against their DPS baseline, which substantially outperformed PINN, DeepONet, PINO and FNO baselines \citep[Table 1]{yao2026guided}. 

\paragraph{Metrics.}
We report the relative $L_2$ error $\mathcal{E}$ between predicted and ground-truth fields, except for Darcy flow, which uses an error rate (Appendix~\ref{app:pde_details}). To summarise the accuracy--compute trade-off in a single number we also report $\mathcal{E} \times t$, where $t$ is wall-clock inference time in seconds; lower is better. All experiments were run on a single NVIDIA A100 GPU, and results are averaged over 100 test instances, each with an independently sampled observation mask.

\paragraph{Observation setup.} Sparse observations consist of 500 randomly selected pixels of the coefficient field $a$, from which the method must reconstruct $u$. Snapshots are on a $128 \times 128$ grid, so observations cover approximately $3\%$ of each field (Appendix~\ref{app:pde_details}, Figure~\ref{fig:known_indices}).

\paragraph{Sampler and SMC settings.}
DPS and the stochastic proposals produce a single sample per run, while the SMC methods use $N = 8$ particles with systematic resampling triggered when the effective sample size drops below $N/2$. We vary the number of discretisation steps $K \in \{25, 50, 100, 150, 200, 500\}$ to assess performance under different compute budgets. Likelihood weights $\beta, \gamma, \omega$ in~\eqref{eq:likelihoods_expanded} are tuned per PDE and inherited from \citet[Appendix G, Table 4]{yao2026guided}.

\section{Discussion and Conclusions}

\begin{figure*}[!t]
    \centering
    \includegraphics[width=\linewidth]{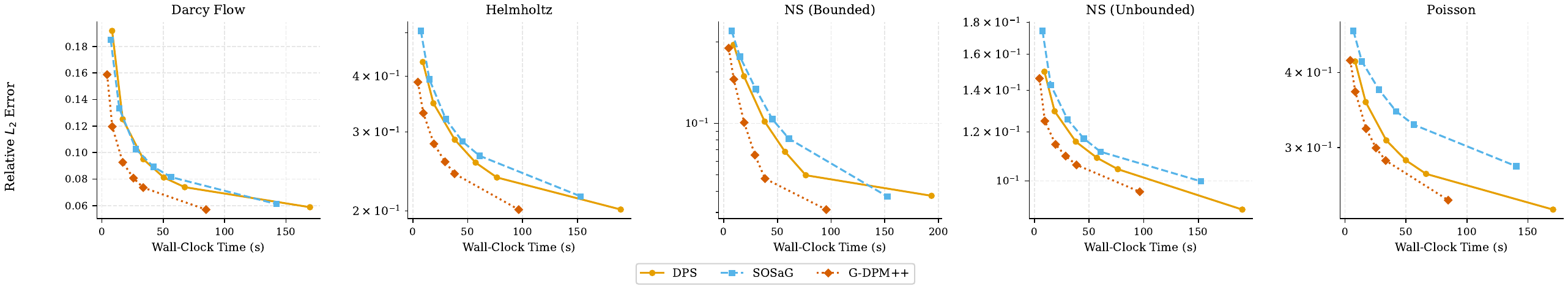}
    \caption{Sampler error vs.\ wall-clock time (efficiency) across the five PDE benchmarks. Each curve traces the trade-off between accuracy and compute as the number of sampling steps is varied; lower-and-leftward is better. 
    }
    \label{fig:sampling_time}
\end{figure*}

\paragraph{Proposal comparison.}
Table~\ref{tab:combined} shows that G-DPM++ achieves the best time-adjusted error $\mathcal{E} \times t$ on every benchmark and step count, and the lowest relative error $\mathcal{E}$ on the majority: as a multistep second-order method that recycles previous score estimates, it incurs negligible additional cost per step relative to DPS while producing a more accurate trajectory. The advantage is most pronounced at low step counts ($K \leq 50$), where the second-order correction compensates for coarse discretisation, and narrows as $K$ increases and all proposals converge to the same continuous-time limit; at $K \geq 150$ DPS attains the lowest raw error on several benchmarks, though at substantially greater cost. Figure~\ref{fig:sampling_time} shows the resulting error--time curves.

\paragraph{Effect of pBS.}
Wrapping each proposal in pBS improves accuracy across most benchmarks and step counts, with resampling and reweighting correcting for approximation error accumulated during sampling. G-DPM++ with pBS is the strongest method in 23 of the 30 inverse-problem settings (Appendix~\ref{app:tables}, Table~\ref{tab:wins}). The gain is largest where the proposal is under-resolved: for NS-Bounded at $K=25$, pBS reduces relative error from $0.2765$ to $0.1498$, a $46\%$ reduction through SMC alone. It diminishes as $K$ grows, and in a small number of settings (Helmholtz at $K=500$, NS-Unbounded at $K \geq 100$) the SMC-augmented sampler is essentially tied with or marginally worse than its proposal-only counterpart. Example reconstructions are shown in Appendix~\ref{app:graphs}.

\paragraph{Practical considerations.}
The pBS framework is well suited to settings where the target solution is deterministic. Although diffusion models are typically used to sample from distributions, the motivation here is computational: in many complex PDE scenarios a traditional solver can take days or weeks, and must be rerun for each new parameter set. A diffusion prior generalises across inputs, so approximate solutions are fast and accurate enough to let a user explore parameters quickly, reserving the traditional solver for the cases that warrant it. In this context the uncertainty quantification aspect of diffusion models is less physically meaningful, and treating the true target as a deterministic function is better suited to the PDE setting. Traditional solvers are also best suited to scenarios where the full initial conditions are known, whereas diffusion priors with guidance handle sparse observations naturally. Finally, pBS is agnostic to the choice of stochastic sampler: we use two common second-order methods, and alternatives such as the Stochastic Adams Solver~\citep[SA-Solver;][]{NEURIPS2023_f4a68064} could be substituted.

\paragraph{Limitations.}
Results on the \emph{forward} problem (Appendix~\ref{app:tables}, Table~\ref{tab:forward}) are more nuanced. SMC G-DPM++ wins every benchmark at $K=25$ and matches or beats DPS at $K=50$, but at $K \geq 100$ DPS frequently surpasses the SMC-augmented methods, taking 17 of 30 forward settings overall. We attribute this to the differing role of SMC across regimes: its advantage is largest when the proposal is under-resolved, either because $K$ is small or because the problem is ill-conditioned. Forward problems at high $K$ satisfy neither, so additional particle diversity offers little benefit and may introduce variance through resampling. Inverse problems remain ill-conditioned regardless of $K$, since the map from sparse observations to coefficient fields is underdetermined, and we hypothesise this yields a likelihood landscape on which a pointwise gradient is unreliable. A tempering ablation (Appendix~\ref{app:tempering}) shows that stronger tempering generally lowers error, most visibly for inverse problems at large $K$, but that the effect is not monotone and tuning appears problem-dependent. A fuller characterisation of the pBS target, and of when it coincides with the diffusion posterior, is left to future work.

Our residual likelihood~\eqref{eq:likelihoods_expanded} makes the reliance on the governing equation explicit: $\omega$ weights the PDE residual against the observation terms $\beta, \gamma$. Where the simulator is misspecified, this term would propagate equation error into the reconstruction, shifting the appropriate balance toward observation guidance. Characterising this trade-off and whether particle diversity under pBS offers any robustness to it, is a natural direction for future work.


\section*{Acknowledgements}

This work was partially supported by the Swedish Research Council and the Wallenberg AI, Autonomous Systems and Software Program (WASP) funded by the Knut and Alice Wallenberg (KAW) Foundation. Computations were enabled by the supercomputing resource Berzelius provided by National Supercomputer Centre at Link\"{o}ping University and the KAW foundation.


\newpage
{\small
\bibliographystyle{plainnat}
\bibliography{references}
}

\newpage
\appendix
\section{Dataset Details} \label{app:pde_details}

\begin{figure*}[!tb]
    \centering
    \includegraphics[width=.95\linewidth]{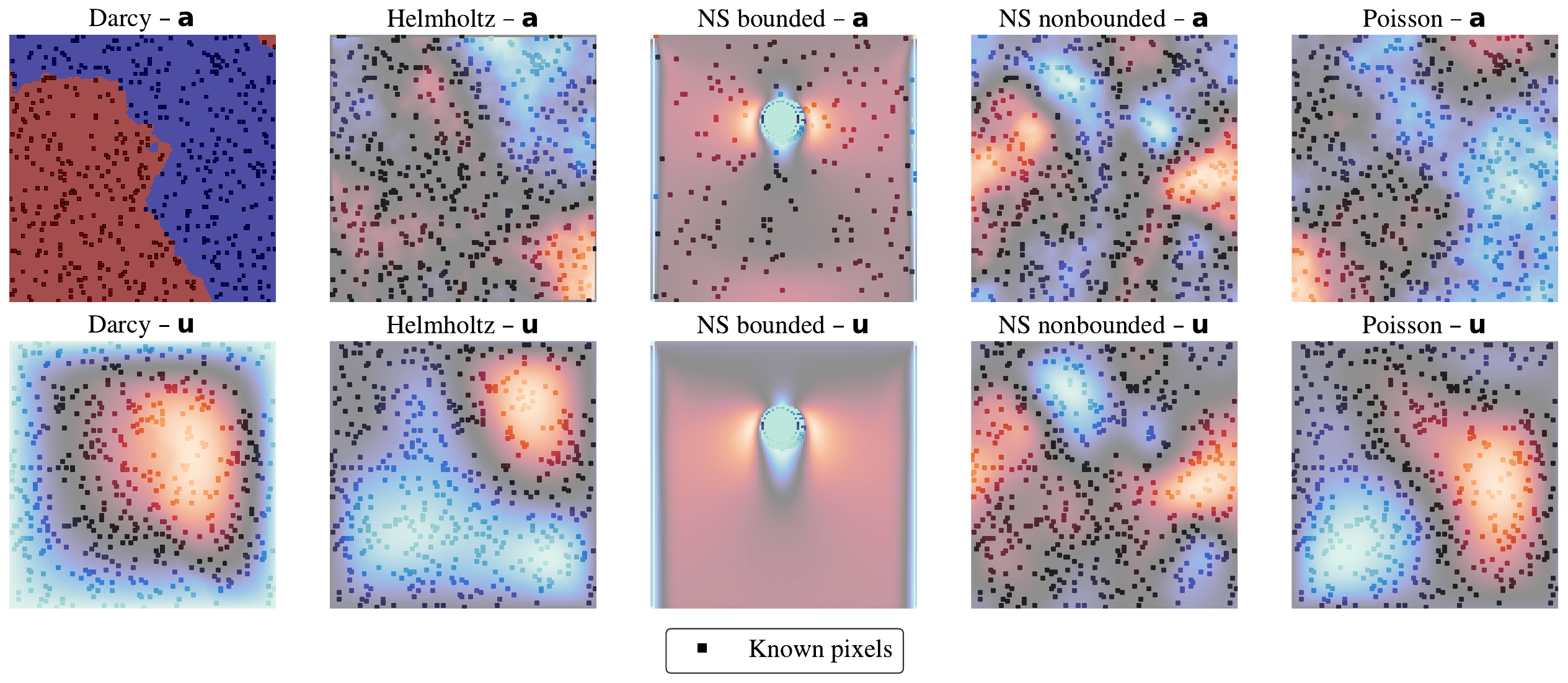}
    \caption{Contour plot of the coefficient and solution fields overlayed with the observed values (highlighted pixels) for each PDE. The top row shows the coefficient field and the bottom row shows the solution field.}
    \label{fig:known_indices}
\end{figure*}

\subsection*{Time-Dependent PDE Formulation}

For a time-dependent PDE defined over a time horizon $[0, \mathcal{T}]$, the solution field $u(\cdot, \tau)$ takes values in $\mathcal{U}$ for each $\tau$, and we write $u \in L^2([0,\mathcal{T}];\, \mathcal{U})$ to denote the full space-time solution. The PDE is written as
\begin{equation}
    \begin{split}
        f(c, \tau, a, a_0, u) = 0&, \quad c \in \Omega,\ \tau \in (0, \mathcal{T}], \\
        u(c, \tau) = g(c, \tau)&, \quad c \in \partial\Omega,\ \tau \in [0, \mathcal{T}], \\
        u(c, 0) = a_0(c)&,
    \end{split}
\end{equation}
where $a_0 \in \mathcal{A}_0 \subset L^2(\Omega)$ is the initial condition, and $a$ collects any fixed PDE coefficients (e.g.\ viscosity, forcing). The solution operator is
\begin{equation}
    \mathcal{G} : \mathcal{A}_0 \to L^2([0,\mathcal{T}];\,\mathcal{U}), \quad a_0 \mapsto u(\cdot,\cdot),
\end{equation}
and the inverse problem is to recover $a_0$ and/or the solution at a later time $u_{\mathcal{T}} = u(\cdot, \mathcal{T})$ from sparse observations.

\paragraph{Darcy Flow.}
In Darcy flow, we solve for the pressure field $u(c)$ given the permeability field $a(c)$ of a specific medium:
\begin{equation}
\begin{aligned}
 -\nabla \cdot (a(c)\, \nabla u(c)) &= s(c), && c \in \Omega, \\
 u(c) &= 0, && c \in \partial \Omega,
\end{aligned}
\label{eq:darcy}
\end{equation}
where $s(c)$ is a source term such as a forcing function. For these experiments, we set $s(c) = 1$.

\paragraph{Inhomogeneous Helmholtz and Poisson Equations.}
In the inhomogeneous Helmholtz equation we solve for the wave field $u(c)$ given a forcing term $a(c)$:
\begin{equation}
\begin{aligned}
\nabla^{2} u(c) + k^{2} u(c) &= a(c), && c \in \Omega, \\
u(c) &= 0, && c \in \partial \Omega,
\end{aligned}
\label{eq:helmholtz}
\end{equation}
where $k$ is the wavenumber and $a(c)$ is a piecewise constant function. When $k = 0$, this reduces to the Poisson equation,
\begin{equation}
\begin{aligned}
\nabla^{2} u(c) &= a(c), && c \in \Omega, \\
u(c) &= 0, && c \in \partial \Omega,
\end{aligned}
\label{eq:poisson}
\end{equation}
which we treat as a separate benchmark in our experiments.

\paragraph{Unbounded Incompressible Navier--Stokes.}
We use the vorticity formulation of the unbounded Navier--Stokes equation:
\begin{equation}
\begin{aligned}
\frac{\partial w(c,\tau)}{\partial \tau} + v(c,\tau) \cdot \nabla w(c,\tau) &= \nu\, \nabla^2 w(c,\tau) + q(c), && c \in \Omega,\ \tau \in [0,\mathcal{T}], \\
\nabla \cdot v(c,\tau) &= 0, && c \in \Omega,\ \tau \in [0,\mathcal{T}],
\end{aligned}
\label{eq:nb_navier_stokes}
\end{equation}
where $w = \nabla \times v$ is the vorticity, $v(c,\tau)$ is the velocity at $c$ at time $\tau$, and $q(c)$ is a forcing field. We follow the same setup as \citet{huang2024diffusionpdegenerativepdesolvingpartial}, setting $\nu = 1 \times 10^{-3}$ and learning the joint distribution of $w_0$ and $w_{\mathcal{T}}$. The dynamics are simulated over $\mathcal{T} = 10$ time steps, equivalent to one second. Because our model outputs are vorticity at $\tau = 0$ and $\tau = \mathcal{T}$ only --- with no intermediate timesteps --- we cannot evaluate the time derivative in~\eqref{eq:nb_navier_stokes}, so we use the simplified guidance term $f = \nabla \cdot w(c, \tau)$.

\paragraph{Bounded Incompressible Navier--Stokes.}
We consider the bounded 2D Navier--Stokes equation:
\begin{equation}
\begin{aligned}
\frac{\partial v(c,\tau)}{\partial \tau} + v(c,\tau) \cdot \nabla v(c,\tau) + \frac{1}{\rho} \nabla p &= \nu\, \nabla^2 v(c,\tau), && c \in \Omega,\ \tau \in [0,\mathcal{T}], \\
\nabla \cdot v(c,\tau) &= 0, && c \in \Omega,\ \tau \in [0,\mathcal{T}],
\end{aligned}
\label{eq:bounded_navier_stokes}
\end{equation}
where $\rho = 1$ and $\nu = 1 \times 10^{-3}$. Following the data-generation process of \citet{huang2024diffusionpdegenerativepdesolvingpartial}, 2D cylinders of random radius and position are generated inside the grid. We learn the joint distribution of $v_0$ and $v_{\mathcal{T}}$ with $\mathcal{T} = 4$, simulating $0.4$ seconds. As in the unbounded case, we use the simplified guidance term $f = \nabla \cdot v(c, \tau)$.

\section{Function-Space Diffusion: Technical Remarks} \label{app:fn_space_remarks}

\paragraph{Reference measure.} In infinite dimensions, the reference distribution cannot be taken as $\mathcal{N}(0, \sigma_{\max}^2 I)$, as is done in finite-dimensional diffusion literature, since the identity operator is not trace-class on $\mathcal{H}$. We instead use $\mathcal{N}(0, \sigma_{\max}^2 \mathcal{C})$ with $\mathcal{C}$ trace-class, ensuring the measure is well-defined.

\paragraph{Score and Cameron--Martin space.} The score 

\begin{equation}
    \nabla_{x_t} \log p(x_t, \sigma_t) = \frac{x_t - \mathcal{D}_\theta(x_t, \sigma_t)}{\sigma_t^2},
    \label{eq:score_function}
\end{equation}
 is the Fr\'echet derivative of $\log p$ in the Cameron--Martin space of the Gaussian measure; the forward noising process is a $\mathcal{H}$-valued Wiener process as in \citet{karras2022elucidatingdesignspacediffusionbased}. The noise-prediction parametrisation of~\eqref{eq:score_function} is consistent with the finite-dimensional EDM formulation; we refer to \citet{karras2022elucidatingdesignspacediffusionbased} for the equivalence in the function-space setting.

\paragraph{Probability flow ODE.} An alternative to the reverse SDE is the probability flow ODE,
\begin{equation*}
    dx_t = -\dot{\sigma}_t \sigma_t \nabla_{x_t} \log p(x_t, \sigma_t)\, dt,
\end{equation*}
which yields the same marginals \citep{song2021scorebasedgenerativemodelingstochastic, karras2022elucidatingdesignspacediffusionbased}. We do not use this formulation in our experiments.

\section{Sequential Monte Carlo Derivations}

\subsection{Sequential Monte Carlo Pseudocode} \label{app:smc_pseudocode}

\begin{algorithm}[H]
\caption{Sequential Monte Carlo (SMC)}
\label{alg:smc}
\begin{algorithmic}[1]
\REQUIRE Number of particles $N$, time steps $K$; proposals $M_K(x_K), M_k(x_{k-1} \mid x_{k})$; incremental weight functions $G_K(x_K), G_k(x_{k-1},x_k)$; resampling threshold $N_{\text{eff}} \in [1, N]$.
\STATE \textbf{Initialisation:}
\FOR{$i = 1 \rightarrow N$}
    \STATE $x_K^{(i)} \sim M_K(x^{(i)}_K)$
    \STATE $w_K^{(i)} = G_K(x^{(i)}_K)$
\ENDFOR
\FOR{$k = K \rightarrow 1$}
    \FOR{$i = 1 \rightarrow N$}
        \STATE \textbf{Propagate:} $x_{k-1}^{(i)} \sim M_k(x^{(i)}_{k-1} \mid x^{(i)}_{k})$ 
        \STATE \textbf{Weight update:} $\tilde{w}_{k-1}^{(i)} = w^{(i)}_{k}\, G_k(x_{k-1}^{(i)}, x_k^{(i)})$
    \ENDFOR
    \STATE \textbf{Normalise:} $w_{k-1}^{(i)} = \tilde{w}_{k-1}^{(i)} \Big/ \sum_{j=1}^N \tilde{w}_{k-1}^{(j)}$, for all $i$
    \STATE \textbf{Compute effective sample size:} $\mathrm{ESS}_k = 1 \big/ \sum_{i=1}^N (w_{k-1}^{(i)})^2$
    \IF{$\mathrm{ESS}_k \leq N_{\text{eff}}$}
        \STATE \textbf{Resampling (Multinomial):}
        \FOR{$i = 1 \rightarrow N$}
            \STATE $a^{(i)} \sim \mathrm{Cat}(w_{k-1}^{(1)}, \dots, w_{k-1}^{(N)})$
            \STATE $x_{k-1}^{(i)} = x_{k-1}^{(a^{(i)})}$
            \STATE $w_{k-1}^{(i)} = 1/N$
        \ENDFOR
    \ENDIF
\ENDFOR
\STATE \textbf{Return:} $\{(x_0^{(i)}, w_0^{(i)})\}_{i=1}^N$
\end{algorithmic}
\end{algorithm}

\subsection{Sampler Update Equations and Pseudocode} \label{app:pseudocodes}

\subsubsection{Guided Euler--Maruyama} \label{app:pseudocodes_gem}

To generate samples we draw $x_T \sim p_{\text{ref}}$ and denoise via the reverse-time SDE,
\begin{equation}
    dx_t = -2\dot{\sigma}_t \sigma_t \nabla_{x_t} \log p(x_t, \sigma_t)\, dt + \sqrt{2\dot{\sigma}_t \sigma_t}\, dW_t^{\mathcal{H}}.
    \label{eq:reverse_sde_noise_form}
\end{equation}

The simplest discretisation of the reverse-time SDE~\eqref{eq:reverse_sde_noise_form} is the Euler--Maruyama scheme:
\begin{equation}
    x_{k-1} = x_{k} - (\sigma_{k-1}^2 - \sigma_{k}^2)\, \nabla_{x_{k}} \log p(x_{k}, \sigma_{k}) + \sqrt{\sigma_{k-1}^2 - \sigma_{k}^2}\, z,
    \label{eq:EM_score_app}
\end{equation}
where $z \sim \mathcal{N}(0, \mathcal{C})$, and the discretisation interval is implicitly accounted for in the noise schedule $\{\sigma_k\}$. We denote this transition as $q^{\textsc{em}}(x_{k-1} \mid x_k) = \mathcal{N}(x_{k-1};\, \mu_{\textsc{em}},\, (\sigma_{k-1}^2 - \sigma_k^2)\, \mathcal{C})$, where $\mu_{\textsc{em}} = x_k - (\sigma_{k-1}^2 - \sigma_k^2)\, \nabla_{x_k} \log p(x_k, \sigma_k)$. Simulating from $k = K$ to $k = 0$ yields a trajectory distributed according to $q^{\textsc{em}}(x_{0:K})$.

Substituting the guided score from~\eqref{eq:guidance_sde_noise} for the unconditional score yields the GEM update:
\begin{equation}
    x_{k-1} = x_{k} - (\sigma_{k-1}^2 - \sigma_{k}^2)\frac{x_{k} - \mathcal{D}_{\theta}(x_{k}, \sigma_{k})}{\sigma_{k}^2} - (\sigma_{k-1}^2 - \sigma_{k}^2)\, \nabla_{x_{k}} \log \tilde{p}_\theta(y \mid x_k, \sigma_k) + \sqrt{\sigma_{k-1}^2 - \sigma_{k}^2}\, z.
    \label{eq:EM_denoiser_guided_constant_app}
\end{equation}

\begin{algorithm}[h!]
\caption{GEM Proposal}
\label{alg:gem_proposal}
\begin{algorithmic}[1]
\REQUIRE{$\mathcal{D}_{\theta}(x; \sigma),\, \sigma_k,\, \sigma_{k-1},\, x_{k}$}
    \STATE Sample $\xi_{k} \sim \mathcal{N}(0, \mathcal{C})$
    \STATE $d_{k} \leftarrow \dfrac{x_{k} - \mathcal{D}_{\theta}(x_{k}, \sigma_{k})}{\sigma_{k}}$
    \STATE $x_{k-1} \leftarrow x_{k} + (\sigma_{k-1}^2 - \sigma_{k}^2)\, d_{k} + \sqrt{\sigma_{k-1}^2 - \sigma_{k}^2}\, \xi_{k}$
    \STATE $x_{k-1} \leftarrow x_{k-1} - (\sigma_{k-1}^2 - \sigma_{k}^2)\, \nabla_{x_{k}}\log \tilde{p}_\theta(y \mid x_k, \sigma_k)$
    \STATE \textbf{return} $x_{k-1}$
\end{algorithmic}
\end{algorithm}

\subsubsection{SOSaG} \label{app:pseudocodes_sosag}

The second-order stochastic sampler with guidance (SOSaG) is based on the EDM stochastic sampler of \citet{karras2022elucidatingdesignspacediffusionbased}. At each step, additional noise is injected to perturb the state from noise level $\sigma_k$ to $\hat{\sigma}_k = (1 + \chi_k)\sigma_k$, yielding $\hat{x}_k = x_k + \sqrt{\hat{\sigma}_k^2 - \sigma_k^2}\,\epsilon_k$ with $\epsilon_k \sim \mathcal{N}(0, \mathcal{C})$, where $\chi_k \geq 0$ is the per-step churn parameter. The update then proceeds via a second-order Heun correction with guidance:
\begin{align}
    d_k &= \frac{\hat{x}_k - \mathcal{D}_\theta(\hat{x}_k,\, \hat{\sigma}_k)}{\hat{\sigma}_k}, \nonumber \\
    x_{k-1}^{(1)} &= \hat{x}_k + (\sigma_{k-1} - \hat{\sigma}_k)\, d_k, \nonumber \\[4pt]
    x_{k-1} &= \hat{x}_k + (\sigma_{k-1} - \hat{\sigma}_k)\!\left(\tfrac{1}{2} d_k + \tfrac{1}{2}\, \frac{x_{k-1}^{(1)} - \mathcal{D}_\theta(x_{k-1}^{(1)},\, \sigma_{k-1})}{\sigma_{k-1}}\right) \nonumber \\
    &\quad - (\sigma_{k-1}^2 - \hat\sigma_k^2)\, \nabla_{\hat{x}_k} \log \tilde{p}_\theta(y \mid \hat{x}_k, \hat\sigma_k),
    \label{eq:sosag_update_app}
\end{align}
where the second-order correction is omitted at the final step ($\sigma_{k-1} = 0$), reducing to a single Euler step. The churn step reintroduces stochasticity at each iteration, preventing the sampler from collapsing onto a deterministic trajectory, while the Heun correction reduces discretisation error relative to the first-order scheme.

\begin{algorithm}[h!]
\caption{SOSaG Proposal}
\label{alg:sosag_proposal}
\begin{algorithmic}[1]
\REQUIRE{$\mathcal{D}_{\theta}(x; \sigma),\, \sigma_{k},\, \sigma_{k-1},\, x_{k},\, \chi_{k}$}
    \STATE Sample $\epsilon_{k} \sim \mathcal{N}(0, \mathcal{C})$
    \STATE $\hat{\sigma}_{k} \leftarrow (1 + \chi_{k})\, \sigma_{k}$
    \STATE $\hat{x}_{k} \leftarrow x_{k} + \sqrt{\hat{\sigma}^{2}_{k} - \sigma^{2}_{k}}\, \epsilon_{k}$
    \STATE $d_{k} \leftarrow \dfrac{\hat{x}_{k} - \mathcal{D}_{\theta}(\hat{x}_{k},\hat{\sigma}_{k})}{\hat{\sigma}_{k}}$
    \STATE $x_{k-1} \leftarrow \hat{x}_{k} + (\sigma_{k-1} - \hat{\sigma}_{k})\, d_{k}$
    \IF{$\sigma_{k-1} \neq 0$}
        \STATE $d_{k-1} \leftarrow \dfrac{x_{k-1} - \mathcal{D}_{\theta}(x_{k-1}, \sigma_{k-1})}{\sigma_{k-1}}$
        \STATE $x_{k-1} \leftarrow \hat{x}_{k} + (\sigma_{k-1} - \hat{\sigma}_{k})\!\left(\tfrac{1}{2} d_{k} + \tfrac{1}{2} d_{k-1}\right)$
    \ENDIF
    \STATE $x_{k-1} \leftarrow x_{k-1} - (\sigma_{k-1}^2 - \hat{\sigma}_{k}^2)\, \nabla_{\hat{x}_{k}} \log \tilde{p}_\theta(y \mid \hat{x}_{k}, \hat{\sigma}_k)$
    \STATE \textbf{return} $x_{k-1}$
\end{algorithmic}
\end{algorithm}

\subsubsection{G-DPM++} \label{app:pseudocodes_g-dpmpp}

We record here how the stochastic DPM-Solver++(2M) update of \citet{Lu_2025},
stated for a general noise schedule, specializes to the variance-exploding (VE)
schedule used in our sampler. This documents that the backbone of G-DPM++
(Eq.~\ref{eq:gdpmpp_update_app}, without the guidance term) is a faithful instance
of the published method rather than a new scheme.

\paragraph{The general update.}
Work in the log-SNR variable $\lambda=\log(\alpha_\lambda/\sigma_\lambda)$, and let
$\lambda_r<\lambda_s<\lambda_t$ be three consecutive nodes with step sizes
$h:=\lambda_t-\lambda_s$ and $h':=\lambda_s-\lambda_r$, and ratio $r_1:=h'/h$. Writing
$P_s:=x_\theta(x_{\lambda_s},\lambda_s)$ and $P_r:=x_\theta(x_{\lambda_r},\lambda_r)$ for
the data-prediction outputs at the two most recent nodes, and approximating
$x_\theta(x_\lambda,\lambda)$ by the linear extrapolant through $P_s,P_r$, the
SDE-DPM-Solver++(2M) update reads \citep{Lu_2025}
\begin{equation}
  x_{\lambda_t}=\frac{\sigma_t}{\sigma_s}e^{-h}\,x_{\lambda_s}
   +\alpha_t\bigl(1-e^{-2h}\bigr)\,P_s
   +\alpha_t\,\frac{1-e^{-2h}}{2}\,\frac{P_s-P_r}{r_1}
   +\sigma_t\sqrt{1-e^{-2h}}\;z_s,
  \qquad z_s\sim\mathcal N(0,\mathcal C),
  \label{eq:app-general}
\end{equation}
where $\sigma_s:=\sigma_{\lambda_s}$, $\sigma_t:=\sigma_{\lambda_t}$, and $\mathcal C$ is
the noise covariance. The schedule enters \eqref{eq:app-general} only through the pair
$(\alpha_t,\sigma_t)$ and through $\lambda$, which determines $h$.

\paragraph{The VE schedule.}
The VE schedule is the choice
\[
  \alpha_\lambda\equiv 1,\qquad \sigma_\lambda=e^{-\lambda},
\]
so that the signal is never attenuated and all noising is carried by $\sigma_\lambda$.
Under $\alpha_\lambda\equiv1$ the log-SNR reduces to
\begin{equation}
  \lambda=\log\frac{\alpha_\lambda}{\sigma_\lambda}=-\log\sigma_\lambda,
  \label{eq:app-ve-time}
\end{equation}
which is exactly the mapping $\lambda=-\log\sigma$ used in the sampler, and inverts to
$\sigma_\lambda=e^{-\lambda}$. Consequently the log-SNR step size is
\[
  h=\lambda_t-\lambda_s=-\log\sigma_t+\log\sigma_s=\log\frac{\sigma_s}{\sigma_t},
\]
so that
\begin{equation}
  e^{-h}=\frac{\sigma_t}{\sigma_s},\qquad
  e^{-2h}=\Bigl(\frac{\sigma_t}{\sigma_s}\Bigr)^{2}.
  \label{eq:app-ve-h}
\end{equation}
Since $\lambda$ increases during sampling while $\sigma$ decreases, $\sigma_t<\sigma_s$
and $h>0$.

\paragraph{Substituting $\alpha_t=1$.}
Setting $\alpha_t=1$ removes every signal-scale prefactor. The leading (decay) coefficient
combines the explicit ratio with $e^{-h}$,
\[
  \frac{\sigma_t}{\sigma_s}\,e^{-h}
   =\frac{\sigma_t}{\sigma_s}\cdot\frac{\sigma_t}{\sigma_s}
   =\Bigl(\frac{\sigma_t}{\sigma_s}\Bigr)^{2},
\]
the first- and second-order coefficients lose their $\alpha_t$, and the diffusion
amplitude retains its explicit $\sigma_t$. Collecting terms gives the VE update
\begin{equation}
  x_{\lambda_t}=\Bigl(\frac{\sigma_t}{\sigma_s}\Bigr)^{2}x_{\lambda_s}
   +\bigl(1-e^{-2h}\bigr)\,P_s
   +\frac{1-e^{-2h}}{2}\,\frac{P_s-P_r}{r_1}
   +\sigma_t\sqrt{1-e^{-2h}}\;z_s,
  \label{eq:app-ve-update}
\end{equation}
with $e^{-2h}=(\sigma_t/\sigma_s)^2$ throughout. Note the decay coefficient
$(\sigma_t/\sigma_s)^2$ equals $\frac{\sigma_t}{\sigma_s}e^{-h}$; our implementation
evaluates it as the latter product, which accounts for the apparent extra factor relative
to a first-order VE step.

\paragraph{Correspondence with G-DPM++.}
Identifying the sampling indices $s\!\to\!k$, $t\!\to\!k-1$, $r\!\to\!\text{prev}$ (so
$P_s=\hat x_0$, $P_r=\hat x_0^{\text{prev}}$, $r_1=r_k$, $z_s=\epsilon_k$),
Eq.~\eqref{eq:app-ve-update} is exactly the backbone of the G-DPM++ update
\eqref{eq:gdpmpp_update_app}: the decay term $(\sigma_{k-1}/\sigma_k)^2 x_k$, the
first-order weight $1-e^{-2h_k}$ on $\hat x_0$, the second-order correction
$\tfrac{1-e^{-2h_k}}{2r_k}(\hat x_0-\hat x_0^{\text{prev}})$, and the noise amplitude
$\sigma_{k-1}\sqrt{1-e^{-2h_k}}$. G-DPM++ adds the guidance term of
Eq.~\eqref{eq:gdpmpp_update_app} to this backbone; the reparametrization above concerns
the backbone only.

\begin{algorithm}[h!]
\caption{G-DPM++ Proposal}
\label{alg:gdpmpp}
\begin{algorithmic}[1]
\REQUIRE{$\mathcal{D}_{\theta}(x; \sigma),\, \sigma_{k},\, \sigma_{k-1},\, x_{k},\, \hat{x}_0^{\text{prev}},\, \sigma_{k}^{\text{prev}}$}
    \STATE $\lambda_k \leftarrow -\log \sigma_k, \qquad \lambda_{k-1} \leftarrow -\log \sigma_{k-1}$
    \STATE $h_k \leftarrow \lambda_{k-1} - \lambda_k$
    \STATE $\hat{x}_0 \leftarrow \mathcal{D}_{\theta}(x_k,\, \sigma_k)$
    \STATE $x_{k-1} \leftarrow \dfrac{\sigma_{k-1}}{\sigma_k}\, e^{-h_k}\, x_k + \bigl(1 - e^{-2h_k}\bigr)\, \hat{x}_0$
    \IF{$\hat{x}_0^{\text{prev}}$ is available \AND $\sigma_{k-1} \neq 0$}
        \STATE $\lambda_k^{\text{prev}} \leftarrow -\log \sigma_k^{\text{prev}}$
        \STATE $r_k \leftarrow |\lambda_k^{\text{prev}} - \lambda_k| \,/\, h_k$
        \STATE $x_{k-1} \leftarrow x_{k-1} + \bigl(1 - e^{-2h_k}\bigr)\, \dfrac{\hat{x}_0 - \hat{x}_0^{\text{prev}}}{2\, r_k}$
    \ENDIF
    \STATE Sample $\epsilon_k \sim \mathcal{N}(0, \mathcal{C})$
    \STATE $x_{k-1} \leftarrow x_{k-1} + \sigma_{k-1}\, \sqrt{1 - e^{-2h_k}}\, \epsilon_k$
    \STATE $x_{k-1} \leftarrow x_{k-1} - (\sigma_{k-1}^2 - \sigma_k^2)\, \nabla_{x_k} \log \tilde{p}_\theta(y \mid x_k, \sigma_k)$
    \STATE Cache $\hat{x}_0^{\text{prev}} \leftarrow \hat{x}_0$, \quad $\sigma_k^{\text{prev}} \leftarrow \sigma_k$
    \STATE \textbf{return} $x_{k-1}$
\end{algorithmic}
\end{algorithm}

\subsection{GEM Proposal Derivation} \label{app:twisted_intermediate}

An SMC sampler which evolves a population of $N$ particles for $K$ iterations with transitions $M$ and potential functions $G$ ca ne be describes by the followign Feynmann--Kac formula. 
\begin{align}
    \nu_k(x_{k:K}) &\propto G_K(x_K)\, M_K(x_K) \prod_{j=k+1}^K G_{j-1}(x_j, x_{j-1})\, M_{j-1}(x_{j-1} \mid x_j).
    \label{eq:FK_formula}
\end{align}

The FK formulation used to derive the targets is given in equation~\eqref{eq:FK_formula}; we restate it here for convenience:
\begin{align}
    \nu(x_{0:K}) = \frac{1}{\mathcal{L}}\, G_K(x_K)\, M_K(x_K) \prod_{j=1}^{K} G_{j-1}(x_j, x_{j-1})\, M_{j-1}(x_{j-1} \mid x_{j}),
    \label{eq:FK_formula_again}
\end{align}
where $\mathcal{L}$ is the normalisation constant. Using
\begin{align}
    &M_K(x_K) = p_K(x_K), && M_{j-1}(x_{j-1} \mid x_{j}) = q_\theta^{\textsc{gem}}(x_{j-1} \mid x_{j}, y), \nonumber \\
    &G_K(x_K) = \tilde{p}_{\theta}(y \mid x_{K}, \sigma_K), && G_{j-1}(x_j, x_{j-1}) = \frac{\tilde{p}_{\theta}(y \mid x_{j-1}, \sigma_{j-1})}{\tilde{p}_{\theta}(y \mid x_{j}, \sigma_j)}\, \frac{q^{\textsc{em}}(x_{j-1} \mid x_{j})}{q_\theta^{\textsc{gem}}(x_{j-1} \mid x_{j}, y)}. \nonumber
\end{align}
For $s \in \{1, \ldots, K\}$, substituting these into~\eqref{eq:FK_formula_again} yields the marginal at iteration $s$:
\begin{align}
    \nu(x_{s-1:K}) &\propto \tilde{p}_{\theta}(y \mid x_{K}, \sigma_K)\, p_K(x_K) \prod_{j=s}^{K} \frac{\tilde{p}_{\theta}(y \mid x_{j-1}, \sigma_{j-1})}{\tilde{p}_{\theta}(y \mid x_{j}, \sigma_j)}\, \frac{q^{\textsc{em}}(x_{j-1} \mid x_{j})}{q_\theta^{\textsc{gem}}(x_{j-1} \mid x_{j}, y)}\, q_\theta^{\textsc{gem}}(x_{j-1} \mid x_{j}, y) \\
    &= \tilde{p}_{\theta}(y \mid x_{s-1}, \sigma_{s-1})\, p_K(x_K) \prod_{j=s}^{K} q^{\textsc{em}}(x_{j-1} \mid x_{j}) \\
    &= \tilde{p}_{\theta}(y \mid x_{s-1}, \sigma_{s-1})\, q^{\textsc{em}}(x_{s-1:K}).
\end{align}
Marginalising over $x_{s:K}$ recovers $\nu(x_{s-1}) \propto \tilde{p}_\theta(y \mid x_{s-1}, \sigma_{s-1})\, q^{\textsc{em}}(x_{s-1})$, and at $s=1$ the final marginal is
\begin{align}
    \nu(x_0) \propto \tilde{p}_\theta(y \mid x_0, 0)\, q^{\textsc{em}}(x_0) = p(y \mid x_0)\, q^{\textsc{em}}(x_0) \propto q^{\textsc{em}}(x_0 \mid y),
\end{align}
where the second equality uses $\tilde{p}_\theta(y \mid x_0, 0) = p(y \mid x_0)$ (the approximate likelihood is exact at $t=0$).

\subsection{Pseudo-Bootstrap Derivation} \label{app:pBS_explained}

For the pBS formulation of the FK process, we use:
\begin{align}
    &M_K(x_K) = p_K(x_K), && \overline{M}_{j-1}(x_{j-1} \mid x_{j}) = \overline{p}_\theta(x_{j-1} \mid x_{j}, y), \nonumber \\
    &G_K(x_K) = \tilde{p}_{\theta}(y \mid x_{K}, \sigma_K), && \overline{G}_{j-1}(x_j, x_{j-1}) = \frac{\tilde{p}_{\theta}(y \mid x_{j-1}, \sigma_{j-1})}{\tilde{p}_{\theta}(y \mid x_{j}, \sigma_j)}, \nonumber
\end{align}
where $\overline{p}_\theta(\cdot \mid x_j, y)$ is the (intractable) transition kernel induced by the chosen guided sampler. Substituting into~\eqref{eq:FK_formula_again}:
\begin{align}
    \overline{\nu}(x_{s-1:K}) &\propto \tilde{p}_{\theta}(y \mid x_{K}, \sigma_K)\, p_K(x_K) \prod_{j=s}^{K} \frac{\tilde{p}_{\theta}(y \mid x_{j-1}, \sigma_{j-1})}{\tilde{p}_{\theta}(y \mid x_{j}, \sigma_j)}\, \overline{p}_\theta(x_{j-1} \mid x_{j}, y) \\
    &= \tilde{p}_{\theta}(y \mid x_{s-1}, \sigma_{s-1})\, p_K(x_K) \prod_{j=s}^{K} \overline{p}_\theta(x_{j-1} \mid x_{j}, y) \\
    &= \tilde{p}_{\theta}(y \mid x_{s-1}, \sigma_{s-1})\, \overline{p}_\theta(x_{s-1:K}).
\end{align}
At $s=1$, marginalising over $x_{1:K}$ gives the final target
\begin{align}
    \overline{\nu}_0(x_0) \propto p(y \mid x_0)\, \overline{p}_\theta(x_0 \mid y),
\end{align}
which is the body claim. Unlike the GEM derivation, the proposal-density ratio is absent from $\overline{G}_{j-1}$, so the result is a likelihood-reweighted version of the posterior induced by the guided sampler rather than the exact posterior. 

\newpage
\section{Supplementary Experimental Results}

\subsection{Tables} \label{app:tables}

\begin{table*}[htbp]
\centering
\tiny
\caption{Forward problems: relative $L_2$ error. Bold indicates best performance per column within each PDE.}
\label{tab:forward}
\renewcommand{\arraystretch}{1.15}
\begin{tabular}{llcccccc}
\toprule
\textbf{PDE} & \textbf{Solver} & \textbf{25 steps} & \textbf{50 steps} & \textbf{100 steps} & \textbf{150 steps} & \textbf{200 steps} & \textbf{500 steps} \\
\midrule
\multirow{7}{*}{Darcy}
  & DPM++           & 0.0703 & 0.0409 & 0.0355 & 0.0347 & 0.0343 & 0.0330 \\
  & DPS             & 0.1011 & 0.0457 & \textbf{0.0346} & 0.0340 & \textbf{0.0324} & \textbf{0.0316} \\
  & G-DPM++ pBS     & \textbf{0.0425} & \textbf{0.0368} & 0.0353 & \textbf{0.0339} & 0.0344 & 0.0339 \\
  & SOSaG pBS       & 0.0665 & 0.0434 & 0.0381 & 0.0374 & 0.0354 & 0.0339 \\
  & SOSaG           & 0.0864 & 0.0452 & 0.0377 & 0.0354 & 0.0362 & 0.0343 \\
\midrule
\multirow{7}{*}{Helmholtz}
  & DPM++           & 0.0953 & 0.0605 & 0.0432 & 0.0390 & 0.0343 & \textbf{0.0300} \\
  & DPS             & 0.1265 & \textbf{0.0486} & \textbf{0.0344} & \textbf{0.0325} & \textbf{0.0311} & 0.0320 \\
  & G-DPM++ pBS     & \textbf{0.0814} & 0.0611 & 0.0437 & 0.0394 & 0.0363 & 0.0318 \\
  & SOSaG pBS       & 0.0966 & 0.0672 & 0.0523 & 0.0469 & 0.0428 & 0.0367 \\
  & SOSaG           & 0.1736 & 0.0909 & 0.0651 & 0.0524 & 0.0468 & 0.0323 \\
\midrule
\multirow{7}{*}{NS-Bounded}
  & DPM++           & 0.2433 & 0.1548 & 0.1150 & \textbf{0.0938} & \textbf{0.0906} & 0.0958 \\
  & DPS             & 0.2643 & 0.1853 & 0.1265 & 0.1162 & 0.1047 & \textbf{0.0735} \\
  & G-DPM++ pBS     & \textbf{0.1260} & \textbf{0.0998} & 0.1211 & 0.1315 & 0.1285 & 0.1237 \\
  & SOSaG pBS       & 0.1504 & 0.1134 & \textbf{0.1081} & 0.1027 & 0.1098 & 0.1411 \\
  & SOSaG           & 0.2783 & 0.1979 & 0.1434 & 0.1151 & 0.1486 & 0.1137 \\
\midrule
\multirow{7}{*}{NS-Unbounded}
  & DPM++           & 0.1484 & 0.1067 & 0.0778 & 0.0667 & 0.0601 & 0.0462 \\
  & DPS             & 0.1713 & \textbf{0.0873} & \textbf{0.0619} & \textbf{0.0544} & \textbf{0.0491} & \textbf{0.0414} \\
  & G-DPM++ pBS     & \textbf{0.1245} & 0.0943 & 0.0720 & 0.0622 & 0.0578 & 0.0465 \\
  & SOSaG pBS       & 0.1612 & 0.1036 & 0.0814 & 0.0723 & 0.0652 & 0.0527 \\
  & SOSaG           & 0.2660 & 0.1544 & 0.1145 & 0.0935 & 0.0792 & 0.0534 \\
\midrule
\multirow{7}{*}{Poisson}
  & DPM++           & 0.0654 & 0.0460 & 0.0353 & 0.0292 & 0.0283 & \textbf{0.0246} \\
  & DPS             & 0.0816 & \textbf{0.0328} & \textbf{0.0263} & \textbf{0.0246} & \textbf{0.0246} & 0.0253 \\
  & G-DPM++ pBS     & \textbf{0.0542} & 0.0402 & 0.0317 & 0.0316 & 0.0284 & 0.0273 \\
  & SOSaG pBS       & 0.0682 & 0.0462 & 0.0373 & 0.0332 & 0.0306 & 0.0299 \\
  & SOSaG           & 0.1239 & 0.0676 & 0.0519 & 0.0384 & 0.0363 & 0.0271 \\
\bottomrule
\end{tabular}
\end{table*}

\begin{table*}[htbp]
\centering
\scriptsize
\caption{Number of step-count slots (out of 6 per PDE) where each solver achieves the lowest error. Forward: 5 PDEs $\times$ 6 steps $=$ 30 slots. Inverse: 5 PDEs $\times$ 6 steps $=$ 30 slots. Combined total: 60 slots.}
\label{tab:wins}
\renewcommand{\arraystretch}{1.2}
\begin{tabular}{lcccc}
\toprule
\textbf{Solver} & \textbf{Forward wins} & \textbf{Inverse wins} & \textbf{Combined wins} & \textbf{Combined (\%)} \\
\midrule
DPS           & 17 & 4  & 21 & 35.0\% \\
DPM++         & 4  & 2  & 6  & 10.0\% \\
G-DPM++ pBS   & 8  & 23 & 31 & 51.7\% \\
SOSaG         & 0  & 0  & 0  &  0.0\% \\
SOSaG pBS     & 1  & 1  & 2  &  3.3\% \\
\midrule
\textit{Total} & \textit{30} & \textit{30} & \textit{60} & 100\% \\
\bottomrule
\end{tabular}
\end{table*}

\subsection{Tempering Results} \label{app:tempering}

\begin{table*}[htbp]
\centering
\tiny
\caption{Temper constant and steps ablation for pBS SMC methods (relative $L_2$ error), including DPS at temperature 1 (DPS has no defined temper constant, so it is reported only in the TC~1 column; `--' marks cells without data). \textbf{Bold} indicates the best result among DPS, SOSaG-pBS, and G-DPM++-pBS across all temper constants, for each Steps block, reported separately for the forward and inverse problems.}
\label{tab:temper_steps_ablation_combined}
\renewcommand{\arraystretch}{1.15}
\setlength{\tabcolsep}{3pt}
\resizebox{0.7\textwidth}{!}{%
\begin{tabular}{lllccccc||ccccc}
\toprule
 & & & \multicolumn{5}{c||}{\textbf{Forward Problem}} & \multicolumn{5}{c}{\textbf{Inverse Problem}} \\
\cmidrule(lr){4-8} \cmidrule(lr){9-13}
\textbf{PDE} & \textbf{Steps} & \textbf{Solver} & \textbf{TC 1} & \textbf{TC 50} & \textbf{TC 100} & \textbf{TC 500} & \textbf{TC 1000} & \textbf{TC 1} & \textbf{TC 50} & \textbf{TC 100} & \textbf{TC 500} & \textbf{TC 1000} \\
\midrule
  \multirow{15}{*}{Darcy Flow} & \multirow{3}{*}{25 steps} & DPS & 0.1011 & -- & -- & -- & -- & 0.1918 & -- & -- & -- & -- \\
   &  & SOSaG-pBS & 0.0665 & 0.0495 & 0.0513 & 0.0501 & 0.0466 & 0.1527 & 0.1236 & 0.1223 & 0.1208 & 0.1195 \\
   &  & G-DPM++-pBS & 0.0425 & 0.0424 & 0.0429 & \textbf{0.0408} & 0.0423 & 0.1203 & 0.1130 & \textbf{0.1121} & 0.1124 & 0.1122 \\
  \cmidrule(lr){2-13}
   & \multirow{3}{*}{50 steps} & DPS & 0.0457 & -- & -- & -- & -- & 0.1252 & -- & -- & -- & -- \\
   &  & SOSaG-pBS & 0.0434 & 0.0384 & 0.0380 & 0.0367 & 0.0385 & 0.1071 & 0.0955 & 0.0891 & 0.0919 & 0.0899 \\
   &  & G-DPM++-pBS & 0.0368 & 0.0370 & 0.0372 & 0.0372 & \textbf{0.0364} & 0.0937 & 0.0906 & \textbf{0.0888} & 0.0893 & 0.0912 \\
  \cmidrule(lr){2-13}
   & \multirow{3}{*}{100 steps} & DPS & 0.0346 & -- & -- & -- & -- & 0.0949 & -- & -- & -- & -- \\
   &  & SOSaG-pBS & 0.0381 & 0.0370 & 0.0361 & 0.0356 & 0.0371 & 0.0836 & 0.0732 & 0.0725 & 0.0699 & \textbf{0.0676} \\
   &  & G-DPM++-pBS & 0.0353 & 0.0350 & 0.0351 & \textbf{0.0344} & 0.0346 & 0.0770 & 0.0727 & 0.0732 & 0.0719 & 0.0719 \\
  \cmidrule(lr){2-13}
   & \multirow{3}{*}{150 steps} & DPS & 0.0340 & -- & -- & -- & -- & 0.0812 & -- & -- & -- & -- \\
   &  & SOSaG-pBS & 0.0374 & 0.0346 & 0.0351 & 0.0352 & 0.0352 & 0.0735 & 0.0680 & 0.0651 & 0.0607 & \textbf{0.0604} \\
   &  & G-DPM++-pBS & \textbf{0.0339} & 0.0339 & 0.0341 & 0.0350 & 0.0342 & 0.0716 & 0.0666 & 0.0671 & 0.0670 & 0.0654 \\
  \cmidrule(lr){2-13}
   & \multirow{3}{*}{200 steps} & DPS & \textbf{0.0324} & -- & -- & -- & -- & 0.0739 & -- & -- & -- & -- \\
   &  & SOSaG-pBS & 0.0354 & 0.0349 & 0.0347 & 0.0351 & 0.0352 & 0.0679 & 0.0640 & 0.0623 & 0.0578 & \textbf{0.0566} \\
   &  & G-DPM++-pBS & 0.0344 & 0.0329 & 0.0334 & 0.0332 & 0.0343 & 0.0721 & 0.0630 & 0.0614 & 0.0619 & 0.0603 \\
\midrule
  \multirow{15}{*}{Helmholtz} & \multirow{3}{*}{25 steps} & DPS & 0.1265 & -- & -- & -- & -- & 0.4301 & -- & -- & -- & -- \\
   &  & SOSaG-pBS & 0.0966 & 0.1142 & 0.1124 & 0.1195 & 0.1152 & 0.4514 & 0.4298 & 0.4291 & 0.4337 & 0.4304 \\
   &  & G-DPM++-pBS & \textbf{0.0814} & 0.0844 & 0.0826 & 0.0848 & 0.0872 & 0.3646 & 0.3626 & \textbf{0.3604} & 0.3689 & 0.3683 \\
  \cmidrule(lr){2-13}
   & \multirow{3}{*}{50 steps} & DPS & \textbf{0.0486} & -- & -- & -- & -- & 0.3478 & -- & -- & -- & -- \\
   &  & SOSaG-pBS & 0.0672 & 0.0631 & 0.0628 & 0.0643 & 0.0654 & 0.3655 & 0.3660 & 0.3638 & 0.3657 & 0.3648 \\
   &  & G-DPM++-pBS & 0.0611 & 0.0580 & 0.0572 & 0.0581 & 0.0567 & 0.3209 & 0.3205 & 0.3164 & 0.3190 & \textbf{0.3160} \\
  \cmidrule(lr){2-13}
   & \multirow{3}{*}{100 steps} & DPS & \textbf{0.0344} & -- & -- & -- & -- & 0.2883 & -- & -- & -- & -- \\
   &  & SOSaG-pBS & 0.0523 & 0.0493 & 0.0496 & 0.0505 & 0.0481 & 0.3086 & 0.3157 & 0.3136 & 0.3138 & 0.3109 \\
   &  & G-DPM++-pBS & 0.0437 & 0.0448 & 0.0441 & 0.0424 & 0.0444 & 0.2747 & 0.2756 & 0.2749 & \textbf{0.2746} & 0.2761 \\
  \cmidrule(lr){2-13}
   & \multirow{3}{*}{150 steps} & DPS & \textbf{0.0325} & -- & -- & -- & -- & 0.2563 & -- & -- & -- & -- \\
   &  & SOSaG-pBS & 0.0469 & 0.0420 & 0.0428 & 0.0436 & 0.0449 & 0.2779 & 0.2818 & 0.2810 & 0.2801 & 0.2788 \\
   &  & G-DPM++-pBS & 0.0394 & 0.0402 & 0.0407 & 0.0385 & 0.0398 & 0.2544 & 0.2507 & 0.2516 & \textbf{0.2506} & 0.2506 \\
  \cmidrule(lr){2-13}
   & \multirow{3}{*}{200 steps} & DPS & \textbf{0.0311} & -- & -- & -- & -- & 0.2374 & -- & -- & -- & -- \\
   &  & SOSaG-pBS & 0.0428 & 0.0406 & 0.0406 & 0.0424 & 0.0411 & 0.2592 & 0.2641 & 0.2619 & 0.2605 & 0.2597 \\
   &  & G-DPM++-pBS & 0.0363 & 0.0390 & 0.0377 & 0.0365 & 0.0363 & \textbf{0.2363} & 0.2382 & 0.2388 & 0.2363 & 0.2394 \\
\midrule
  \multirow{15}{*}{NS (Bounded)} & \multirow{3}{*}{25 steps} & DPS & 0.2643 & -- & -- & -- & -- & 0.2887 & -- & -- & -- & -- \\
   &  & SOSaG-pBS & 0.1504 & 0.1122 & \textbf{0.1103} & 0.1145 & 0.1229 & 0.1828 & 0.0921 & 0.0891 & 0.0833 & 0.0844 \\
   &  & G-DPM++-pBS & 0.1260 & 0.1303 & 0.1284 & 0.1359 & 0.1429 & 0.1498 & 0.0990 & 0.0847 & \textbf{0.0805} & 0.0835 \\
  \cmidrule(lr){2-13}
   & \multirow{3}{*}{50 steps} & DPS & 0.1853 & -- & -- & -- & -- & 0.1889 & -- & -- & -- & -- \\
   &  & SOSaG-pBS & 0.1134 & 0.1073 & 0.0921 & 0.0999 & 0.1000 & 0.1135 & 0.0628 & 0.0620 & 0.0606 & 0.0604 \\
   &  & G-DPM++-pBS & 0.0998 & 0.0951 & \textbf{0.0915} & 0.1252 & 0.1096 & 0.0935 & 0.0560 & 0.0505 & \textbf{0.0479} & 0.0479 \\
  \cmidrule(lr){2-13}
   & \multirow{3}{*}{100 steps} & DPS & 0.1265 & -- & -- & -- & -- & 0.1024 & -- & -- & -- & -- \\
   &  & SOSaG-pBS & 0.1081 & 0.0981 & 0.0982 & 0.0980 & 0.0980 & 0.0842 & 0.0623 & 0.0627 & 0.0620 & 0.0623 \\
   &  & G-DPM++-pBS & 0.1211 & \textbf{0.0830} & 0.0890 & 0.0879 & 0.1041 & 0.0498 & 0.0357 & 0.0349 & 0.0339 & \textbf{0.0334} \\
  \cmidrule(lr){2-13}
   & \multirow{3}{*}{150 steps} & DPS & 0.1162 & -- & -- & -- & -- & 0.0681 & -- & -- & -- & -- \\
   &  & SOSaG-pBS & 0.1027 & 0.1018 & 0.0947 & 0.0947 & 0.0947 & 0.0671 & 0.0537 & 0.0536 & 0.0536 & 0.0536 \\
   &  & G-DPM++-pBS & 0.1315 & 0.0790 & 0.0860 & \textbf{0.0788} & 0.0789 & 0.0379 & 0.0316 & 0.0313 & \textbf{0.0308} & 0.0311 \\
  \cmidrule(lr){2-13}
   & \multirow{3}{*}{200 steps} & DPS & 0.1047 & -- & -- & -- & -- & 0.0496 & -- & -- & -- & -- \\
   &  & SOSaG-pBS & 0.1098 & 0.1006 & 0.0930 & 0.0930 & 0.0930 & 0.0605 & 0.0495 & 0.0488 & 0.0488 & 0.0488 \\
   &  & G-DPM++-pBS & 0.1285 & \textbf{0.0775} & 0.0911 & 0.0775 & 0.0836 & 0.0336 & 0.0303 & 0.0303 & \textbf{0.0301} & 0.0302 \\
\midrule
  \multirow{15}{*}{NS (Unbounded)} & \multirow{3}{*}{25 steps} & DPS & 0.1713 & -- & -- & -- & -- & 0.1500 & -- & -- & -- & -- \\
   &  & SOSaG-pBS & 0.1612 & 0.1353 & 0.1388 & 0.1386 & 0.1368 & 0.1584 & 0.1594 & 0.1591 & 0.1593 & 0.1600 \\
   &  & G-DPM++-pBS & 0.1245 & \textbf{0.1194} & 0.1196 & 0.1204 & 0.1232 & 0.1429 & \textbf{0.1415} & 0.1421 & 0.1427 & 0.1428 \\
  \cmidrule(lr){2-13}
   & \multirow{3}{*}{50 steps} & DPS & 0.0873 & -- & -- & -- & -- & 0.1295 & -- & -- & -- & -- \\
   &  & SOSaG-pBS & 0.1036 & 0.0951 & 0.0941 & 0.0946 & 0.0946 & 0.1365 & 0.1367 & 0.1370 & 0.1369 & 0.1369 \\
   &  & G-DPM++-pBS & 0.0943 & \textbf{0.0870} & 0.0877 & 0.0880 & 0.0883 & \textbf{0.1233} & 0.1233 & 0.1236 & 0.1242 & 0.1239 \\
  \cmidrule(lr){2-13}
   & \multirow{3}{*}{100 steps} & DPS & \textbf{0.0619} & -- & -- & -- & -- & 0.1157 & -- & -- & -- & -- \\
   &  & SOSaG-pBS & 0.0814 & 0.0779 & 0.0784 & 0.0775 & 0.0778 & 0.1215 & 0.1214 & 0.1210 & 0.1210 & 0.1211 \\
   &  & G-DPM++-pBS & 0.0720 & 0.0670 & 0.0662 & 0.0671 & 0.0666 & 0.1147 & 0.1145 & \textbf{0.1139} & 0.1142 & 0.1140 \\
  \cmidrule(lr){2-13}
   & \multirow{3}{*}{150 steps} & DPS & \textbf{0.0544} & -- & -- & -- & -- & 0.1090 & -- & -- & -- & -- \\
   &  & SOSaG-pBS & 0.0723 & 0.0676 & 0.0687 & 0.0693 & 0.0683 & 0.1141 & 0.1146 & 0.1141 & 0.1143 & 0.1141 \\
   &  & G-DPM++-pBS & 0.0622 & 0.0578 & 0.0582 & 0.0593 & 0.0583 & \textbf{0.1089} & 0.1094 & 0.1094 & 0.1095 & 0.1091 \\
  \cmidrule(lr){2-13}
   & \multirow{3}{*}{200 steps} & DPS & \textbf{0.0491} & -- & -- & -- & -- & \textbf{0.1045} & -- & -- & -- & -- \\
   &  & SOSaG-pBS & 0.0652 & 0.0619 & 0.0621 & 0.0612 & 0.0616 & 0.1097 & 0.1099 & 0.1100 & 0.1097 & 0.1097 \\
   &  & G-DPM++-pBS & 0.0578 & 0.0538 & 0.0533 & 0.0528 & 0.0528 & 0.1056 & 0.1056 & 0.1052 & 0.1054 & 0.1055 \\
\midrule
  \multirow{15}{*}{Poisson} & \multirow{3}{*}{25 steps} & DPS & 0.0816 & -- & -- & -- & -- & 0.4175 & -- & -- & -- & -- \\
   &  & SOSaG-pBS & 0.0682 & 0.0546 & 0.0544 & 0.0528 & 0.0577 & 0.3927 & 0.3718 & \textbf{0.3662} & 0.3756 & 0.3746 \\
   &  & G-DPM++-pBS & 0.0542 & \textbf{0.0512} & 0.0526 & 0.0515 & 0.0530 & 0.3788 & 0.3671 & 0.3673 & 0.3697 & 0.3705 \\
  \cmidrule(lr){2-13}
   & \multirow{3}{*}{50 steps} & DPS & \textbf{0.0328} & -- & -- & -- & -- & 0.3572 & -- & -- & -- & -- \\
   &  & SOSaG-pBS & 0.0462 & 0.0411 & 0.0403 & 0.0391 & 0.0398 & 0.3541 & 0.3203 & 0.3163 & 0.3159 & \textbf{0.3158} \\
   &  & G-DPM++-pBS & 0.0402 & 0.0382 & 0.0387 & 0.0380 & 0.0377 & 0.3394 & 0.3267 & 0.3209 & 0.3169 & 0.3209 \\
  \cmidrule(lr){2-13}
   & \multirow{3}{*}{100 steps} & DPS & \textbf{0.0263} & -- & -- & -- & -- & 0.3084 & -- & -- & -- & -- \\
   &  & SOSaG-pBS & 0.0373 & 0.0359 & 0.0356 & 0.0362 & 0.0354 & 0.3277 & 0.2884 & 0.2803 & 0.2808 & 0.2815 \\
   &  & G-DPM++-pBS & 0.0317 & 0.0316 & 0.0301 & 0.0299 & 0.0311 & 0.2996 & 0.2881 & 0.2852 & 0.2786 & \textbf{0.2760} \\
  \cmidrule(lr){2-13}
   & \multirow{3}{*}{150 steps} & DPS & \textbf{0.0246} & -- & -- & -- & -- & 0.2858 & -- & -- & -- & -- \\
   &  & SOSaG-pBS & 0.0332 & 0.0317 & 0.0308 & 0.0318 & 0.0322 & 0.3131 & 0.2666 & 0.2647 & 0.2602 & 0.2620 \\
   &  & G-DPM++-pBS & 0.0316 & 0.0298 & 0.0291 & 0.0299 & 0.0297 & 0.2859 & 0.2694 & 0.2662 & \textbf{0.2534} & 0.2560 \\
  \cmidrule(lr){2-13}
   & \multirow{3}{*}{200 steps} & DPS & \textbf{0.0246} & -- & -- & -- & -- & 0.2712 & -- & -- & -- & -- \\
   &  & SOSaG-pBS & 0.0306 & 0.0309 & 0.0311 & 0.0309 & 0.0302 & 0.2993 & 0.2615 & 0.2539 & 0.2497 & 0.2490 \\
   &  & G-DPM++-pBS & 0.0284 & 0.0283 & 0.0274 & 0.0270 & 0.0269 & 0.2711 & 0.2599 & 0.2547 & 0.2435 & \textbf{0.2428} \\
\bottomrule
\end{tabular}
}
\end{table*}

\begin{table*}[htbp]
\centering
\scriptsize
\caption{Number of (PDE, Steps) slots (out of 25 per problem type) where each solver achieves the lowest error, best result taken across all temper constants for that PDE and step count. Forward: 5 PDEs $\times$ 5 steps $=$ 25 slots. Inverse: 5 PDEs $\times$ 5 steps $=$ 25 slots. Combined total: 50 slots.}
\label{tab:wins_tc_ablation}
\renewcommand{\arraystretch}{1.2}
\begin{tabular}{lcccc}
\toprule
\textbf{Solver} & \textbf{Forward wins} & \textbf{Inverse wins} & \textbf{Combined wins} & \textbf{Combined (\%)} \\
\midrule
DPS           & 12 & 1  & 13 & 26.0\% \\
SOSaG pBS     & 1  & 5  & 6  & 12.0\% \\
G-DPM++ pBS   & 12 & 19 & 31 & 62.0\% \\
\midrule
\textit{Total} & \textit{25} & \textit{25} & \textit{50} & 100\% \\
\bottomrule
\end{tabular}
\end{table*}

\begin{figure}[ht]
    \centering
    \begin{subfigure}[t]{\linewidth}
        \centering
        \includegraphics[width=\linewidth]{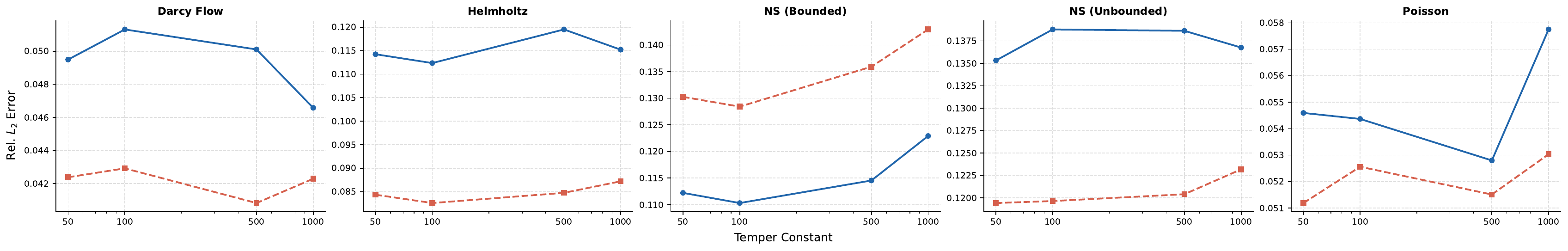}
        \caption{25 steps}
    \end{subfigure}
    \par\vspace{0.8em}
    \begin{subfigure}[t]{\linewidth}
        \centering
        \includegraphics[width=\linewidth]{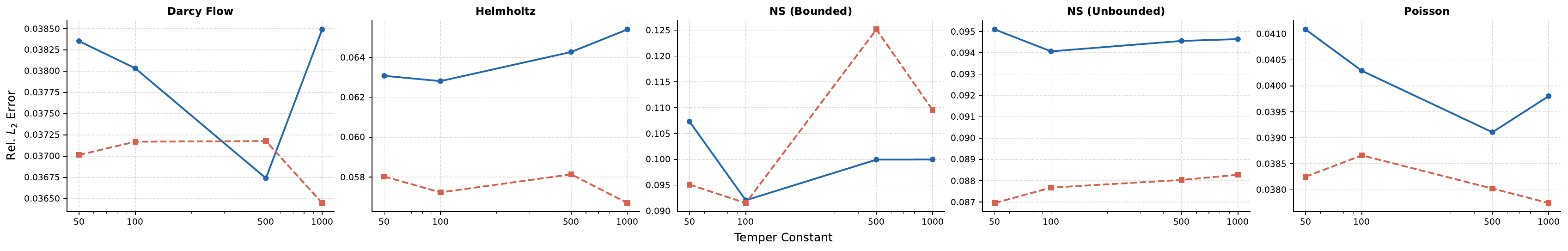}
        \caption{50 steps}
    \end{subfigure}
    \par\vspace{0.8em}
    \begin{subfigure}[t]{\linewidth}
        \centering
        \includegraphics[width=\linewidth]{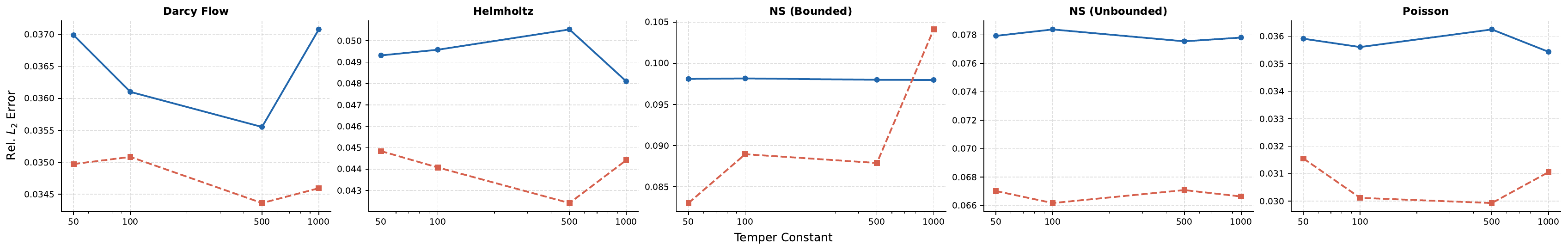}
        \caption{100 steps}
    \end{subfigure}
    \par\vspace{0.8em}
    \begin{subfigure}[t]{\linewidth}
        \centering
        \includegraphics[width=\linewidth]{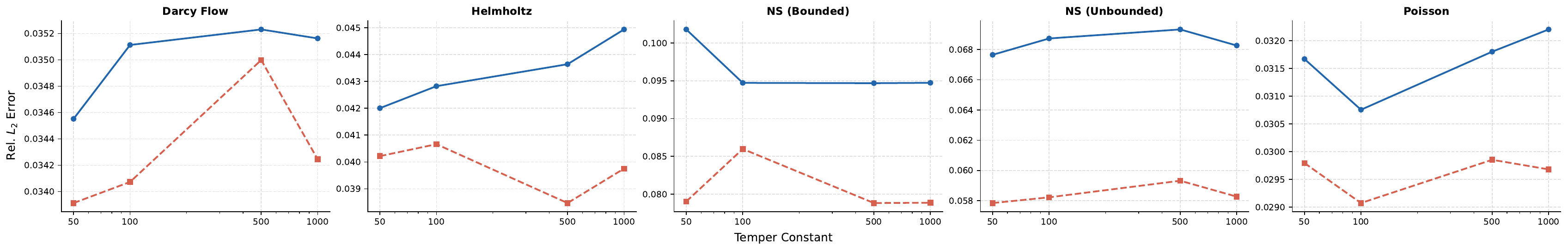}
        \caption{150 steps}
    \end{subfigure}
    \par\vspace{0.8em}
    \begin{subfigure}[t]{\linewidth}
        \centering
        \includegraphics[width=\linewidth]{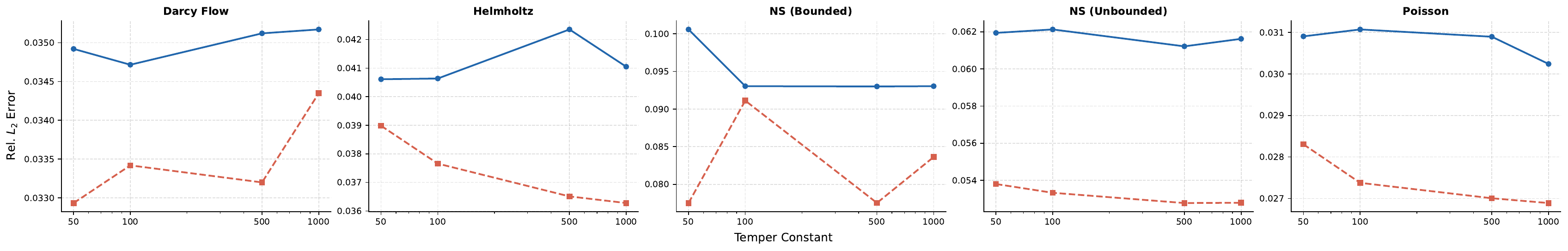}
        \caption{200 steps}
    \end{subfigure}
    \par\vspace{1em}
    \includegraphics[width=0.4\linewidth]{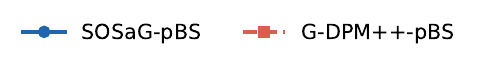}
    \caption{Temper constant ablation across step counts for the forward PDE problems. Generally we find that the tempering parameter value is not correlated with performance.}
    \label{fig:temper_ablation_forward}
\end{figure}

\begin{figure}[ht]
    \centering
    \begin{subfigure}[t]{\linewidth}
        \centering
        \includegraphics[width=\linewidth]{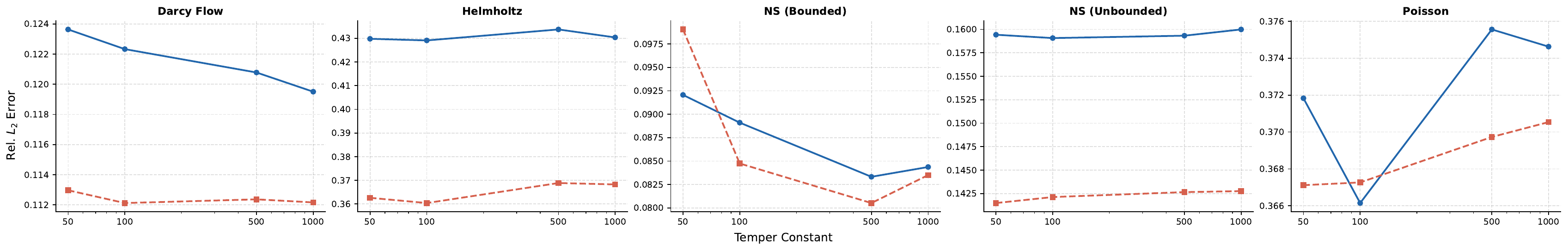}
        \caption{25 steps}
    \end{subfigure}
    \par\vspace{0.8em}
    \begin{subfigure}[t]{\linewidth}
        \centering
        \includegraphics[width=\linewidth]{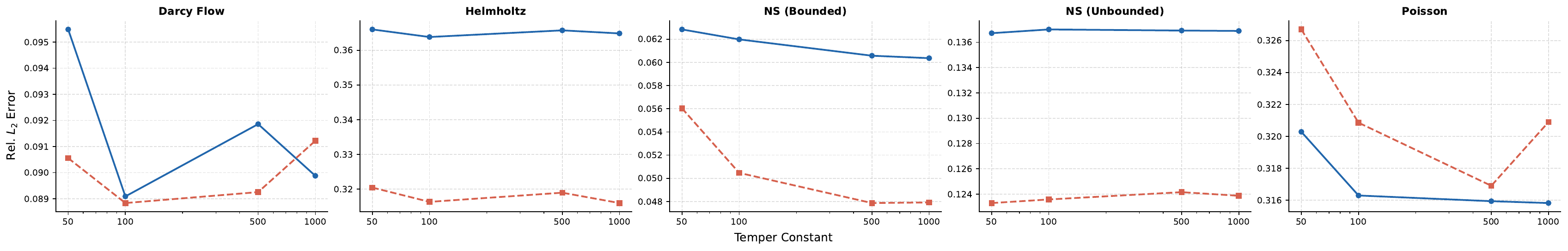}
        \caption{50 steps}
    \end{subfigure}
    \par\vspace{0.8em}
    \begin{subfigure}[t]{\linewidth}
        \centering
        \includegraphics[width=\linewidth]{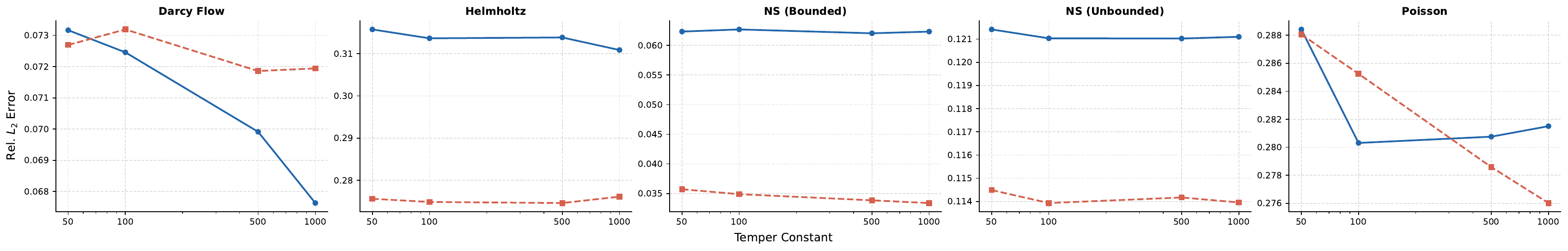}
        \caption{100 steps}
    \end{subfigure}
    \par\vspace{0.8em}
    \begin{subfigure}[t]{\linewidth}
        \centering
        \includegraphics[width=\linewidth]{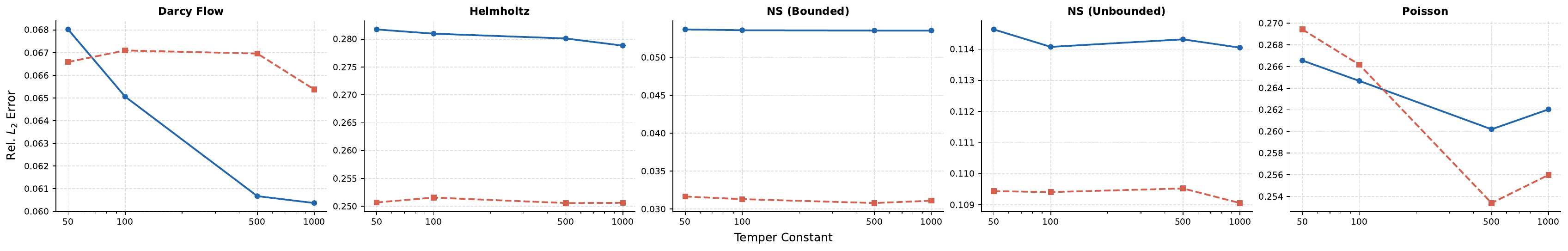}
        \caption{150 steps}
    \end{subfigure}
    \par\vspace{0.8em}
    \begin{subfigure}[t]{\linewidth}
        \centering
        \includegraphics[width=\linewidth]{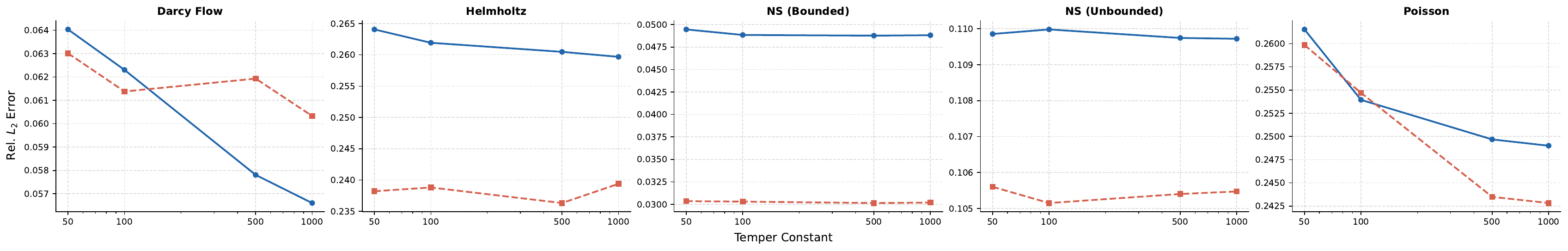}
        \caption{200 steps}
    \end{subfigure}
    \par\vspace{1em}
    \includegraphics[width=0.4\linewidth]{figures/tempering_figs/plot_temper_ablation_legend.pdf}
    \caption{Temper constant ablation across step counts for the inverse PDE problems. We find in the inverse problems that the correlation between the tempering parameter value and relative error is a lot stronger than the forward problems. Darcy Flow and Poisson in particular show strong correlations between better relative error and a higher tempering parameter. We also notice that the increased performance is more pronounced at higher step counts.}
    \label{fig:temper_ablation_inverse}
\end{figure}

\subsection{Reconstruction Examples} \label{app:graphs}

\begin{figure*}[t]
    \centering
    \begin{subfigure}{0.8\textwidth}
        \includegraphics[width=\textwidth]{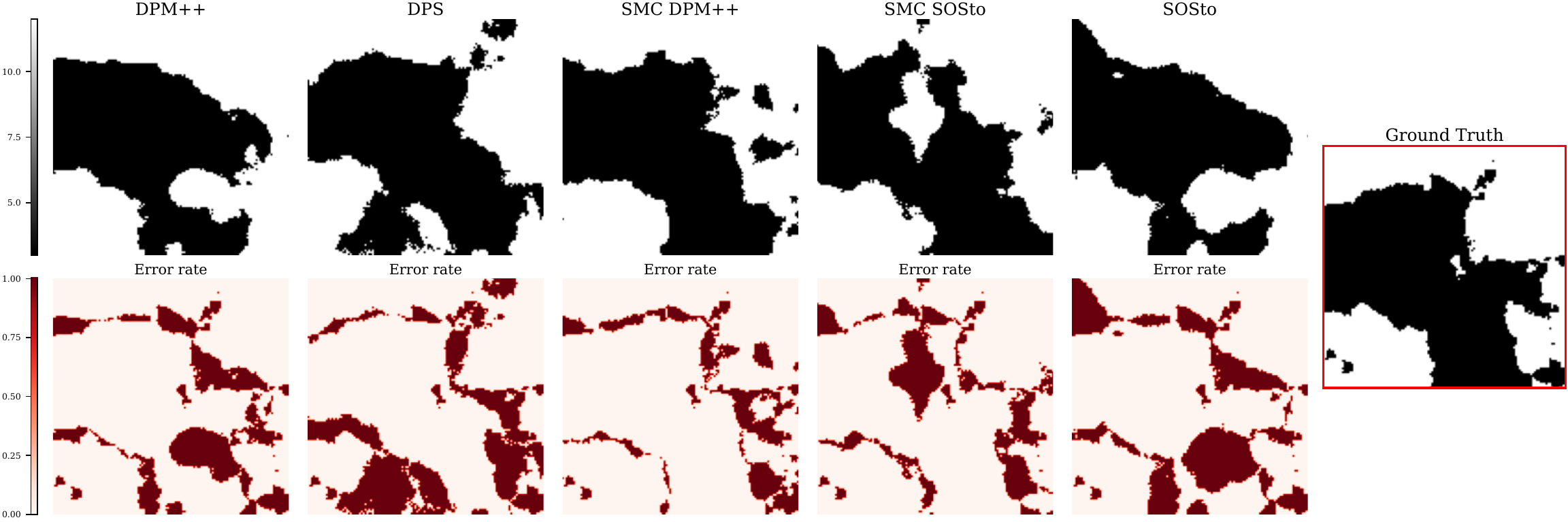}
        \caption{Darcy flow}
    \end{subfigure}
    \vspace{0.5em}
    \begin{subfigure}{0.8\textwidth}
        \includegraphics[width=\textwidth]{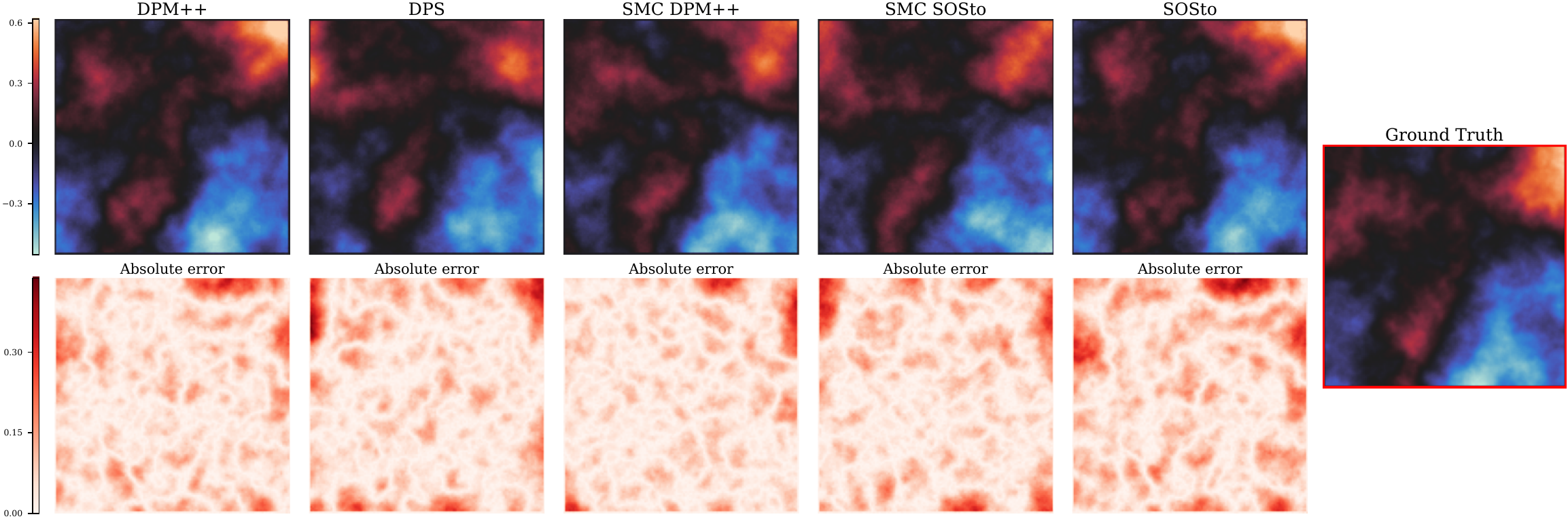}
        \caption{Helmholtz}
    \end{subfigure}
    \vspace{0.5em}
    \begin{subfigure}{0.8\textwidth}
        \includegraphics[width=\textwidth]{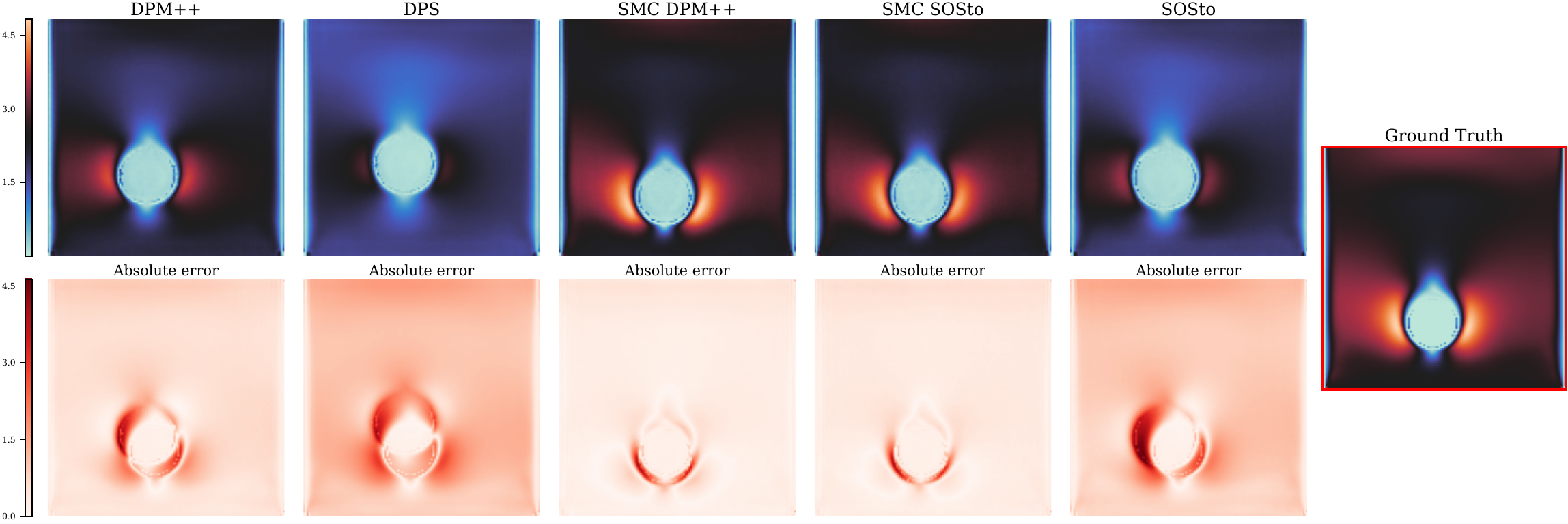}
        \caption{NS-Bounded}
    \end{subfigure}
    \vspace{0.5em}
    \begin{subfigure}{0.8\textwidth}
        \includegraphics[width=\textwidth]{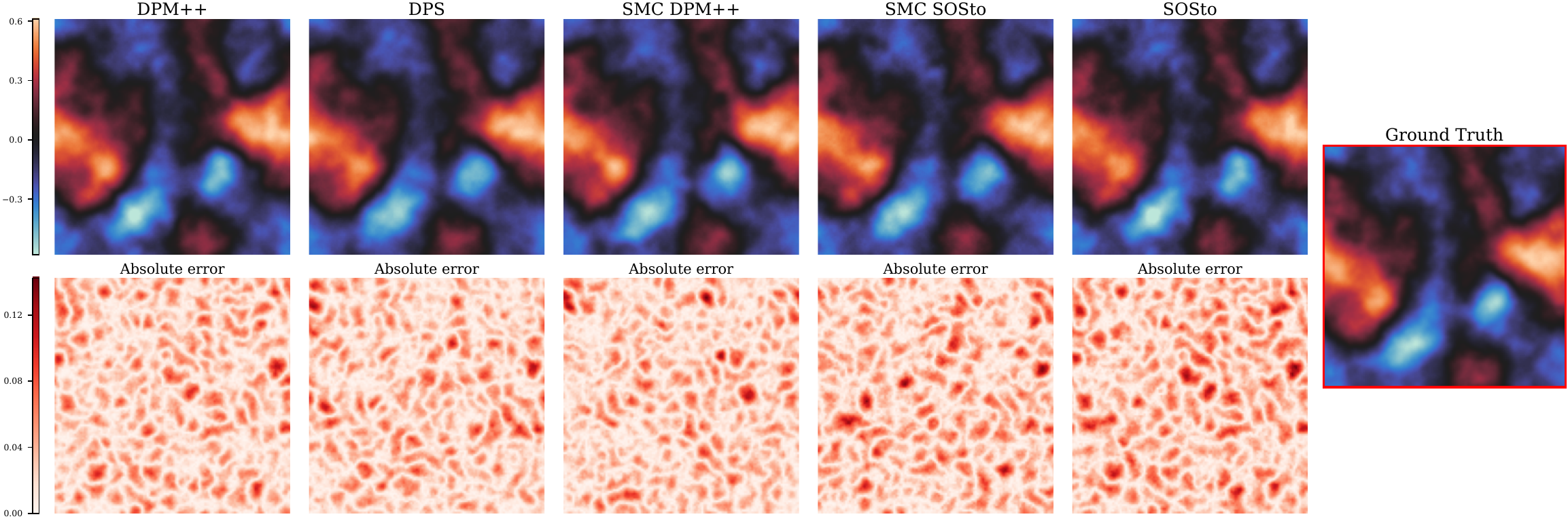}
        \caption{NS-Unbounded}
    \end{subfigure}
    \vspace{0.5em}
    \begin{subfigure}{0.8\textwidth}
        \includegraphics[width=\textwidth]{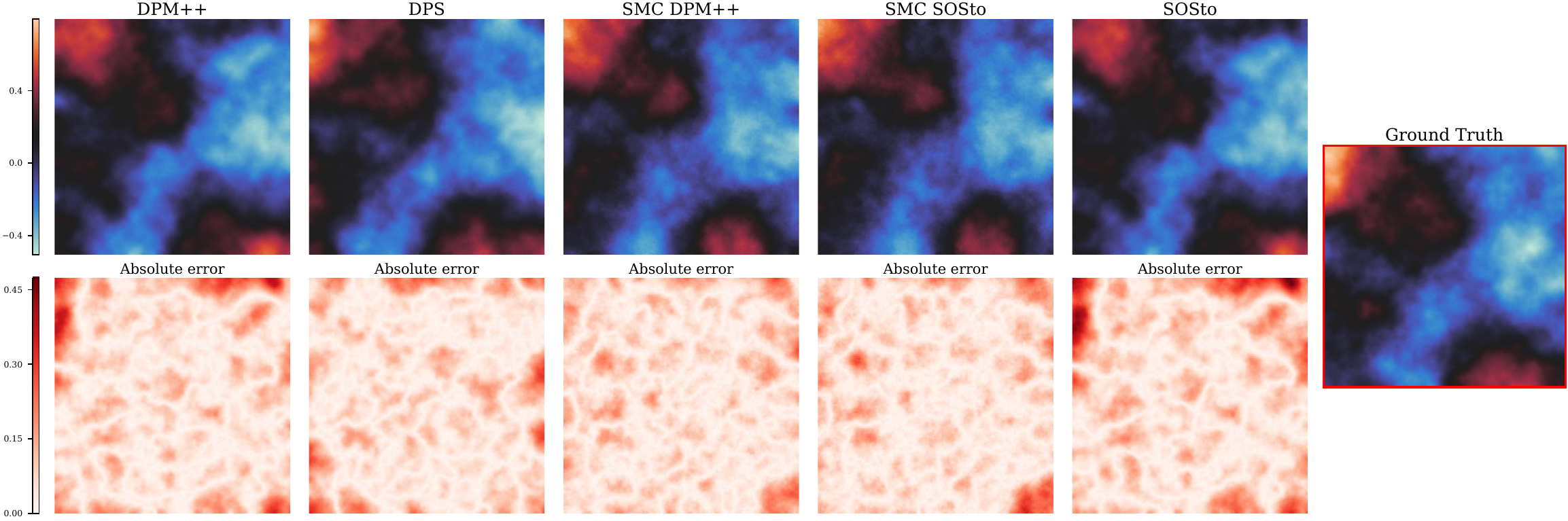}
        \caption{Poisson}
    \end{subfigure}
    \caption{Inverse problem coefficient reconstruction $a$. 25 sampling steps.}
    \label{fig:recon_25_ch0}
\end{figure*}

\begin{figure*}[t]
    \centering
    \begin{subfigure}{0.8\textwidth}
        \includegraphics[width=\textwidth]{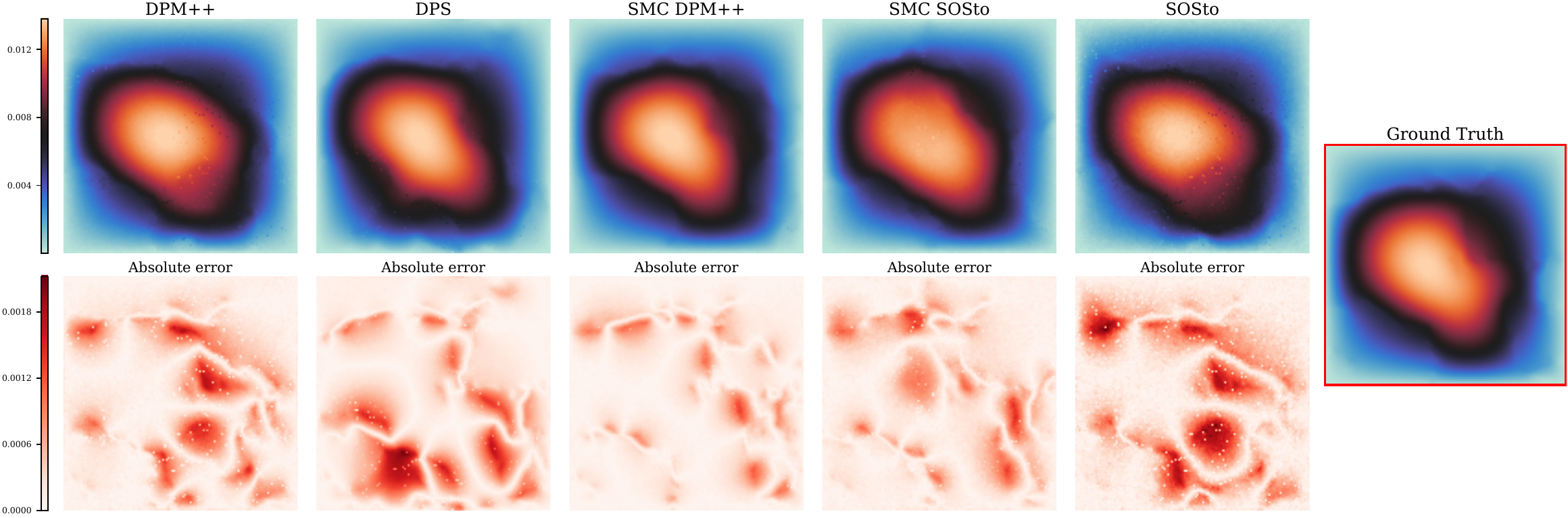}
        \caption{Darcy flow}
    \end{subfigure}
    \vspace{0.5em}
    \begin{subfigure}{0.8\textwidth}
        \includegraphics[width=\textwidth]{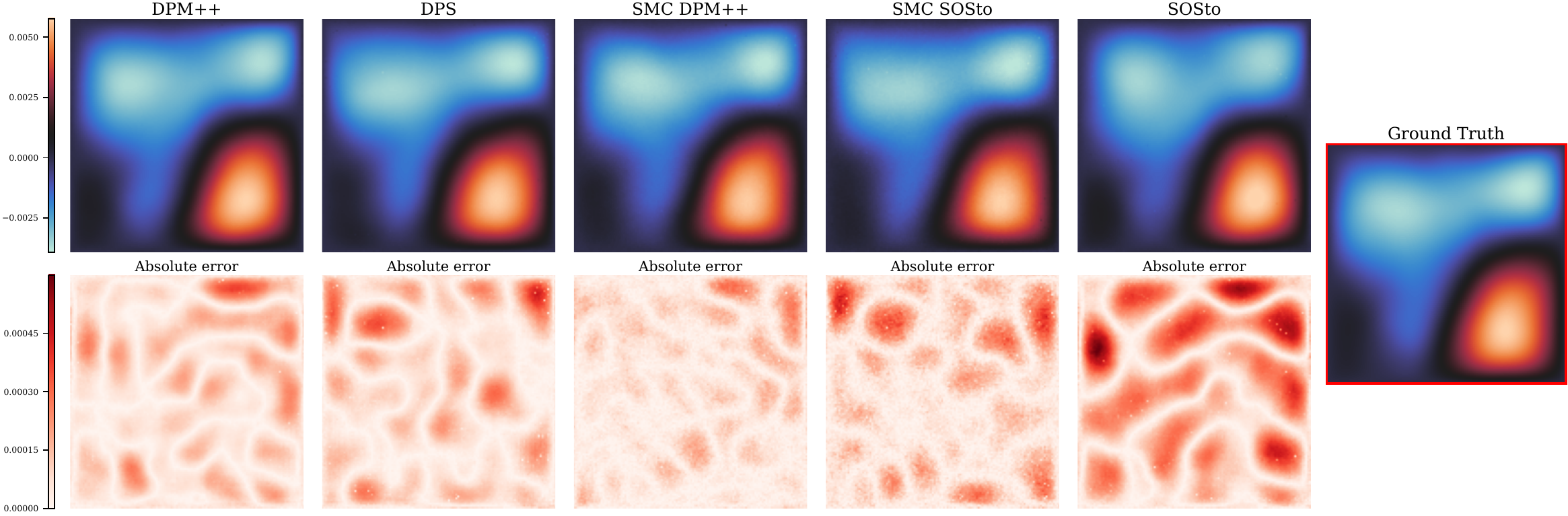}
        \caption{Helmholtz}
    \end{subfigure}
    \vspace{0.5em}
    \begin{subfigure}{0.8\textwidth}
        \includegraphics[width=\textwidth]{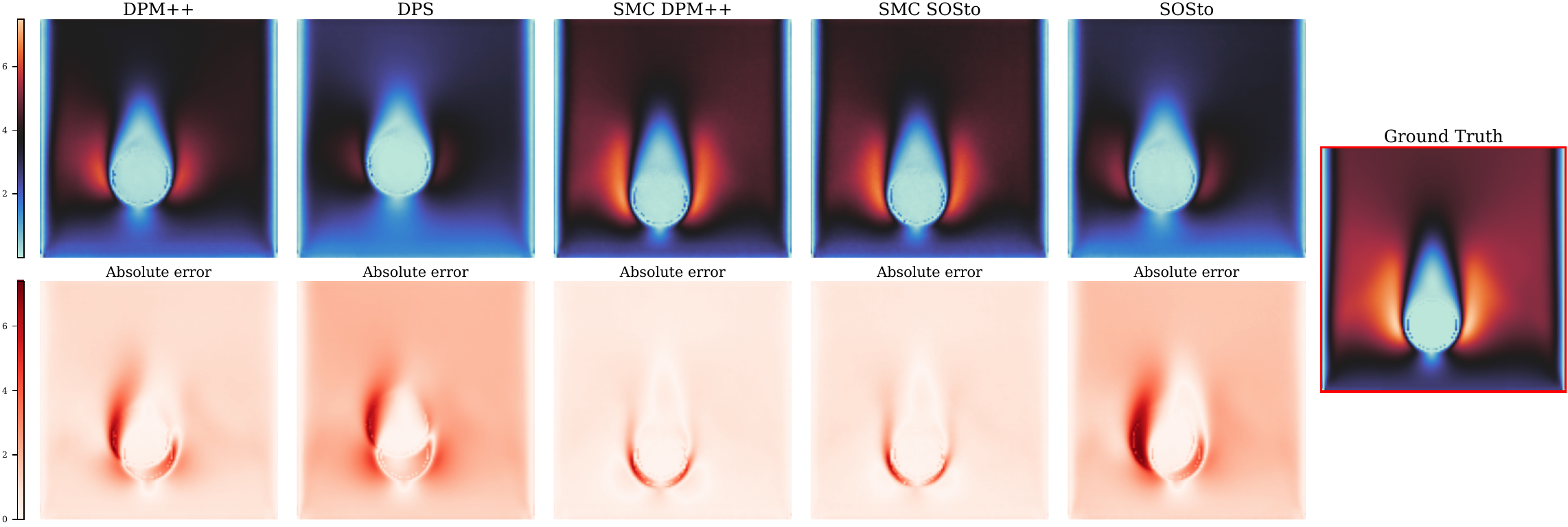}
        \caption{NS-Bounded}
    \end{subfigure}
    \vspace{0.5em}
    \begin{subfigure}{0.8\textwidth}
        \includegraphics[width=\textwidth]{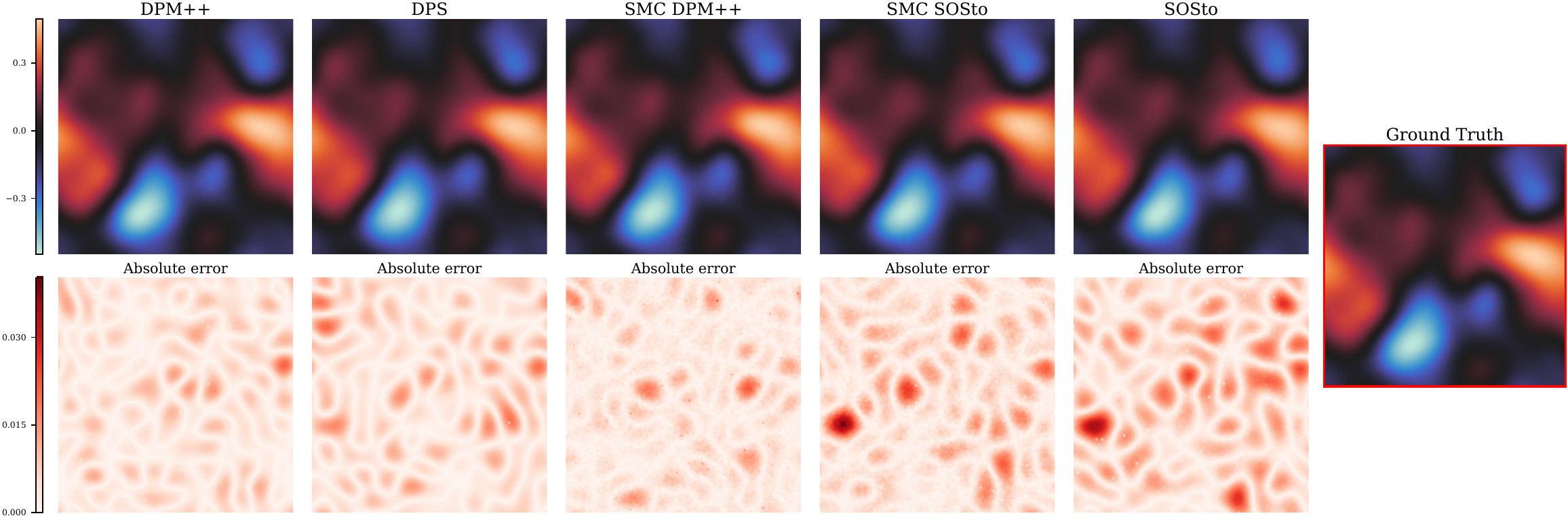}
        \caption{NS-Unbounded}
    \end{subfigure}
    \vspace{0.5em}
    \begin{subfigure}{0.8\textwidth}
        \includegraphics[width=\textwidth]{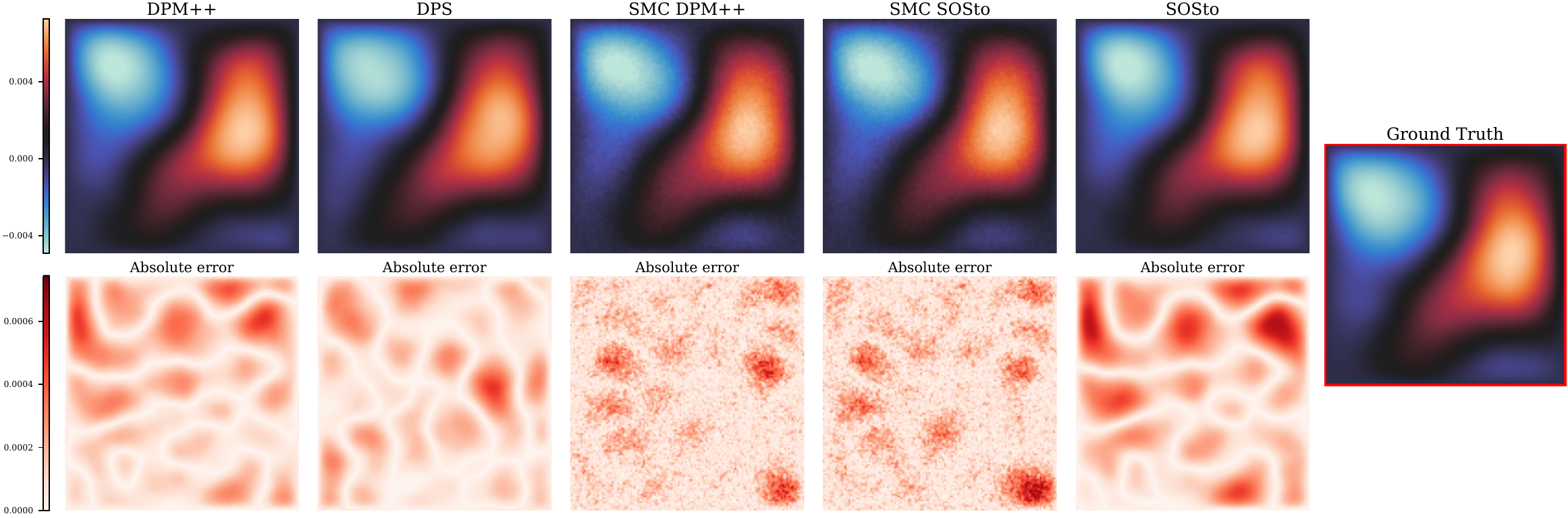}
        \caption{Poisson}
    \end{subfigure}
    \caption{Inverse problem solution field reconstruction $u$. 25 sampling steps.}
    \label{fig:recon_25_ch1}
\end{figure*}

\begin{figure*}[t]
    \centering
    \begin{subfigure}{0.8\textwidth}
        \includegraphics[width=\textwidth]{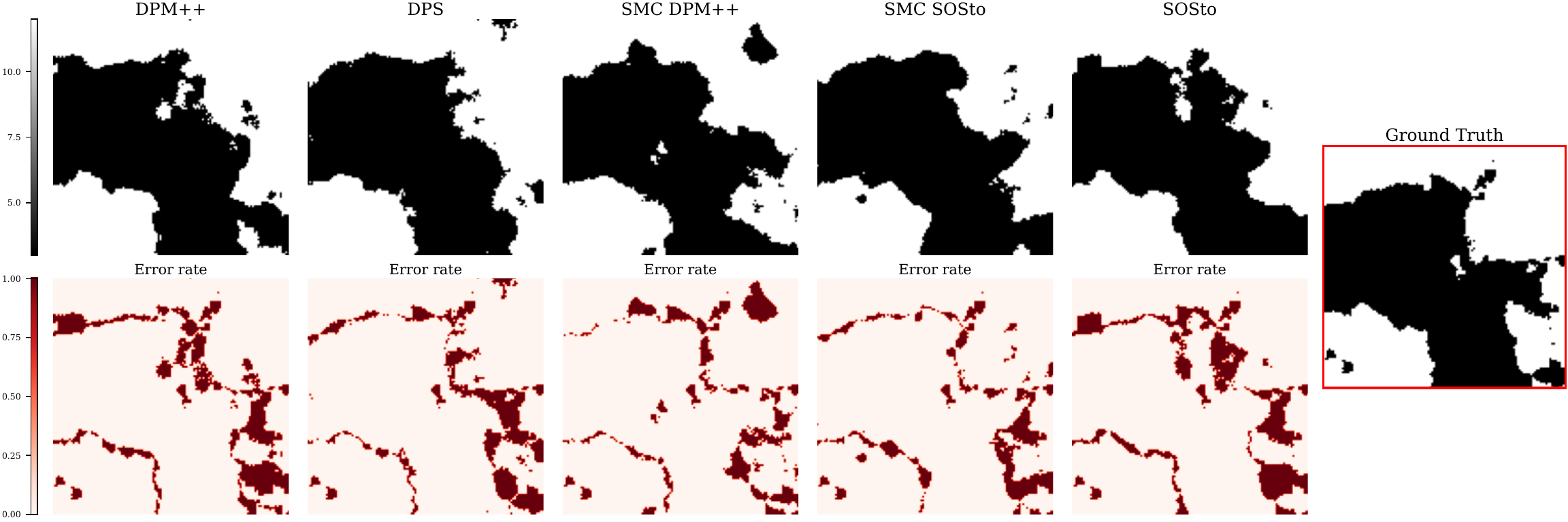}
        \caption{Darcy flow}
    \end{subfigure}
    \vspace{0.5em}
    \begin{subfigure}{0.8\textwidth}
        \includegraphics[width=\textwidth]{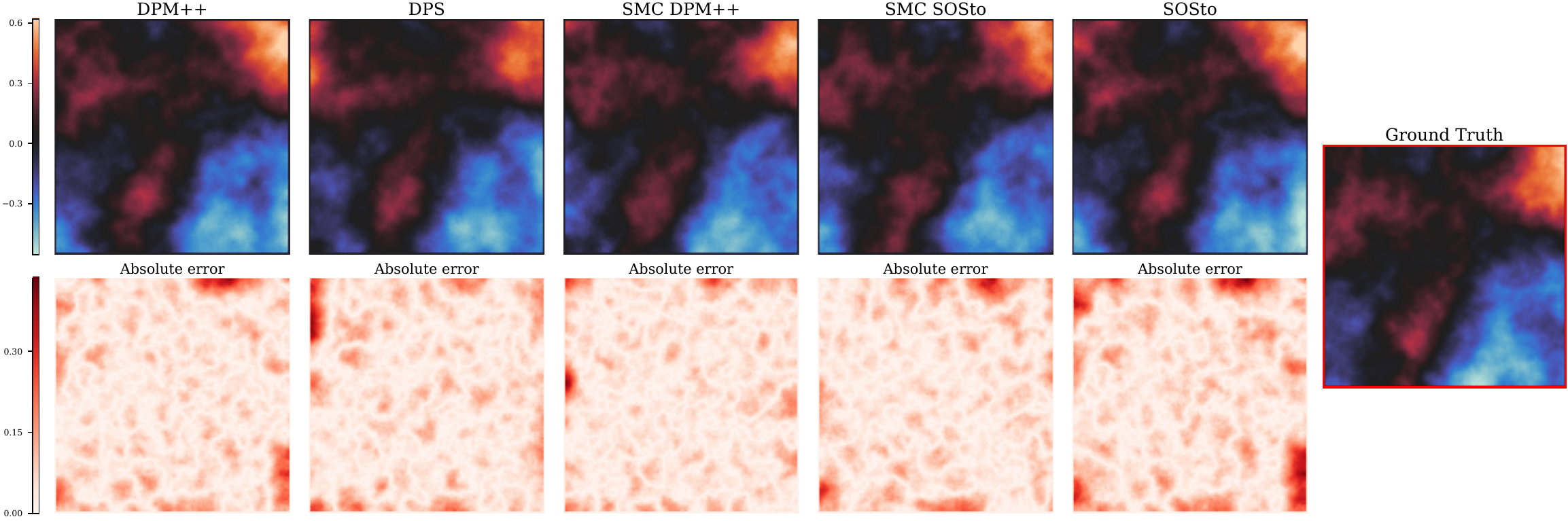}
        \caption{Helmholtz}
    \end{subfigure}
    \vspace{0.5em}
    \begin{subfigure}{0.8\textwidth}
        \includegraphics[width=\textwidth]{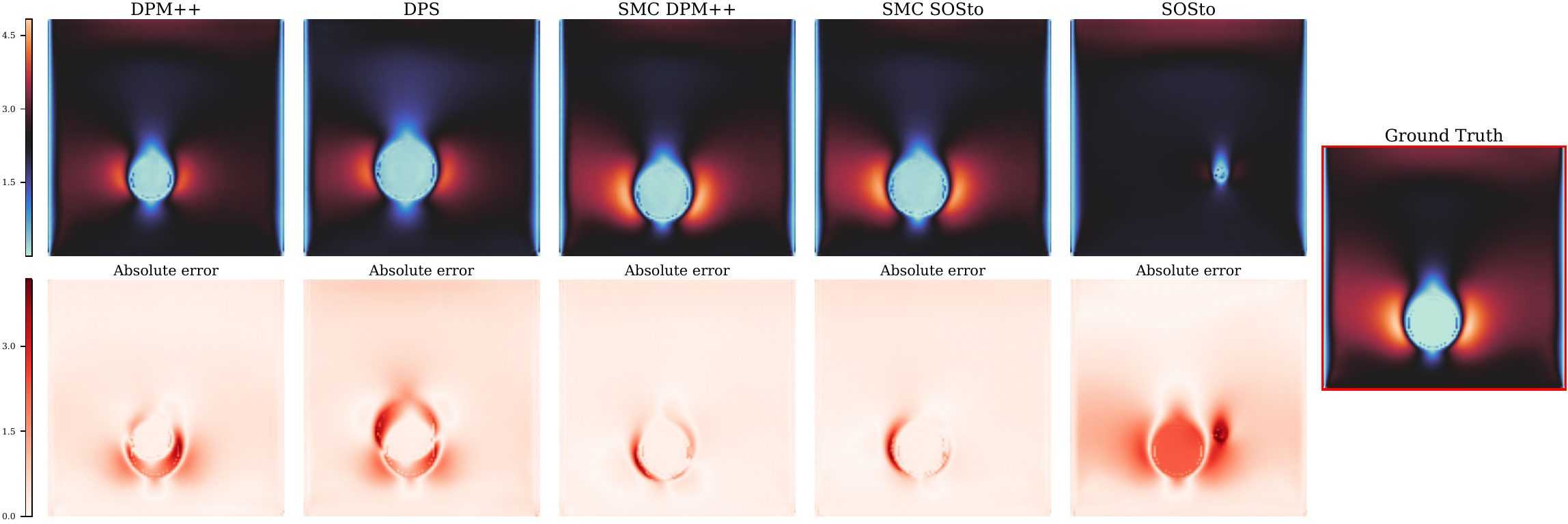}
        \caption{NS-Bounded}
    \end{subfigure}
    \vspace{0.5em}
    \begin{subfigure}{0.8\textwidth}
        \includegraphics[width=\textwidth]{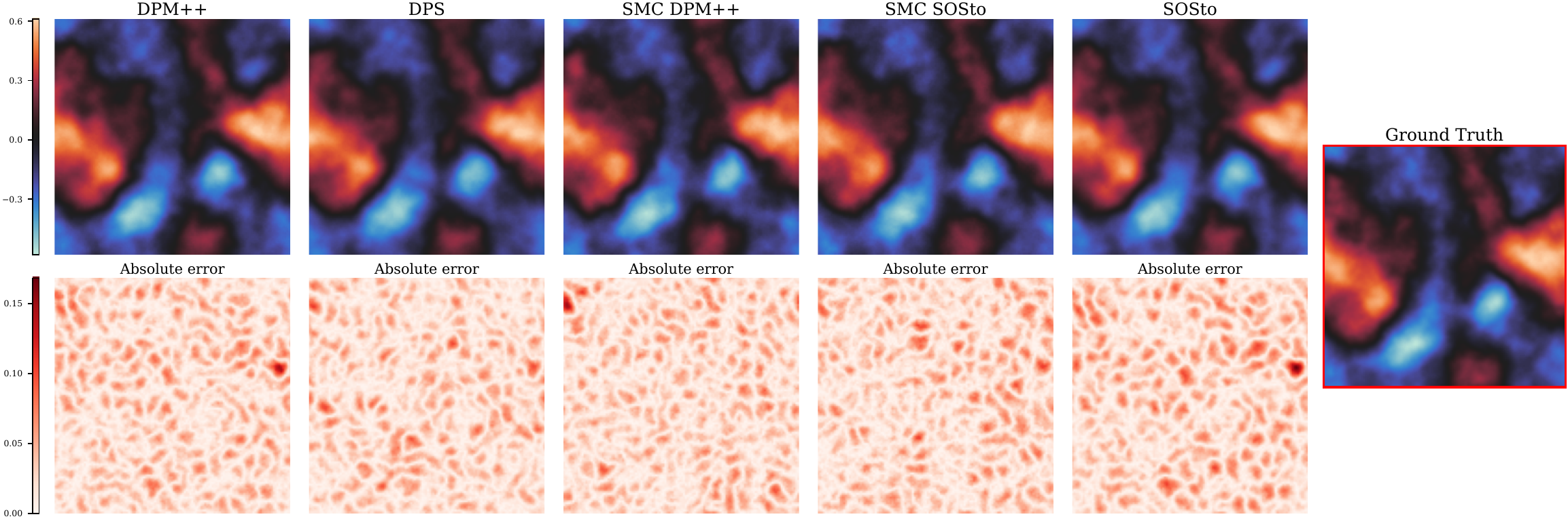}
        \caption{NS-Unbounded}
    \end{subfigure}
    \vspace{0.5em}
    \begin{subfigure}{0.8\textwidth}
        \includegraphics[width=\textwidth]{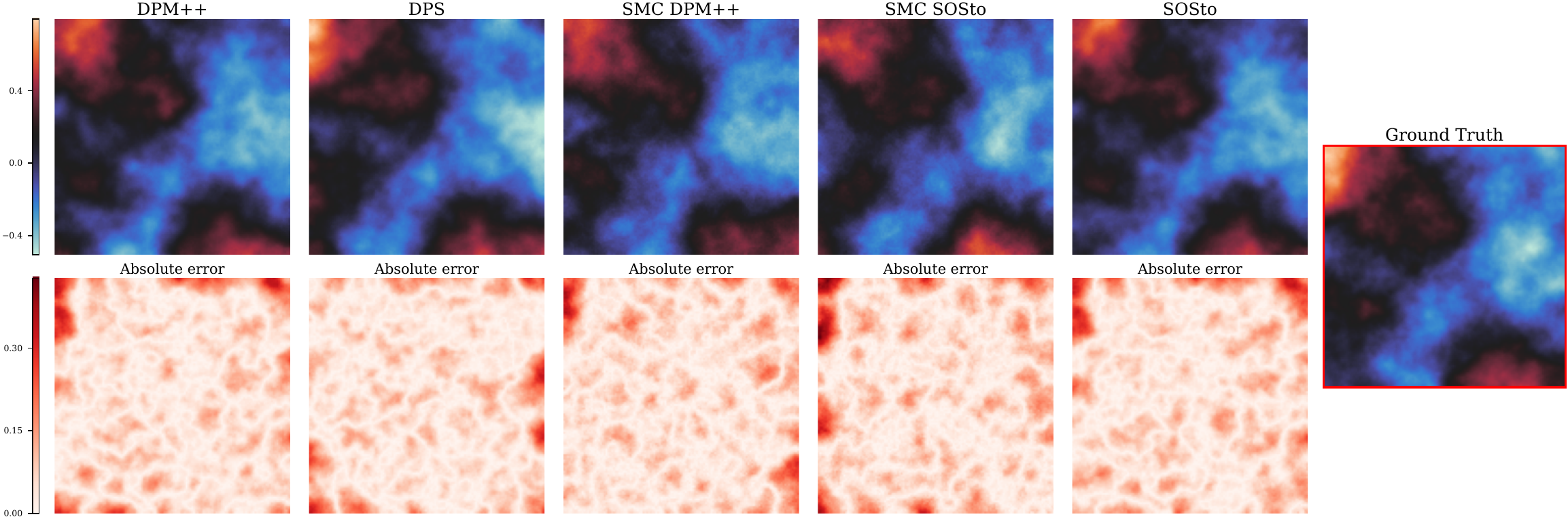}
        \caption{Poisson}
    \end{subfigure}
    \caption{Inverse problem coefficient reconstruction $a$. 50 sampling steps.}
    \label{fig:recon_50_ch0}
\end{figure*}

\begin{figure*}[t]
    \centering
    \begin{subfigure}{0.8\textwidth}
        \includegraphics[width=\textwidth]{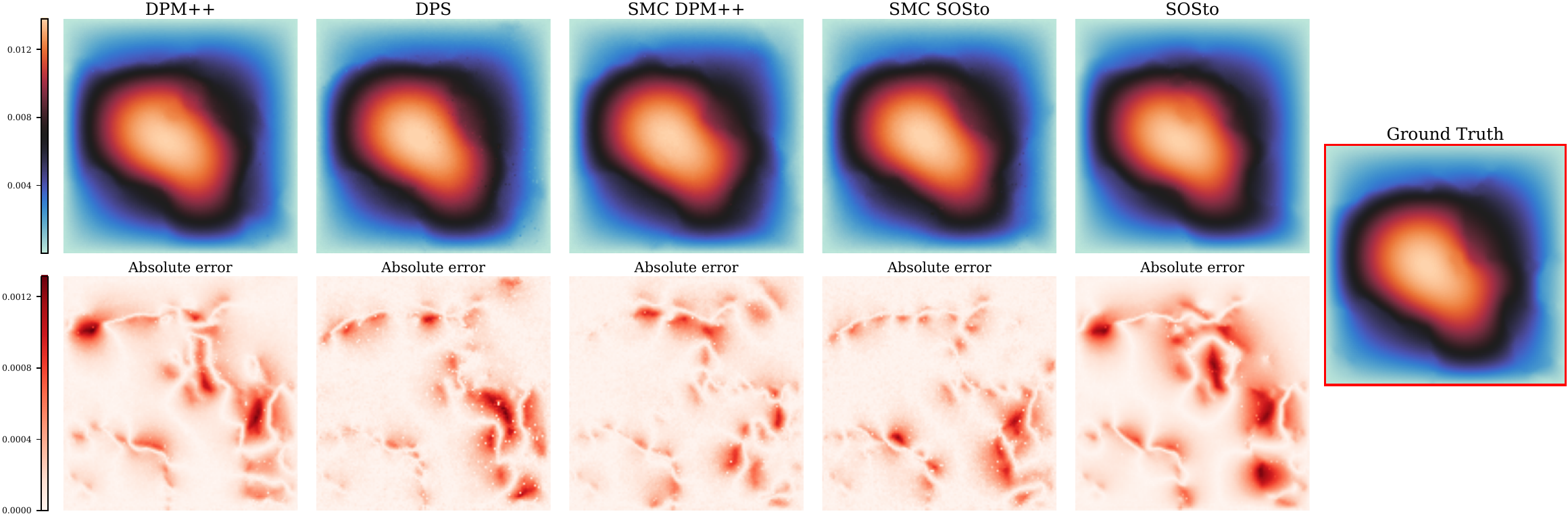}
        \caption{Darcy flow}
    \end{subfigure}
    \vspace{0.5em}
    \begin{subfigure}{0.8\textwidth}
        \includegraphics[width=\textwidth]{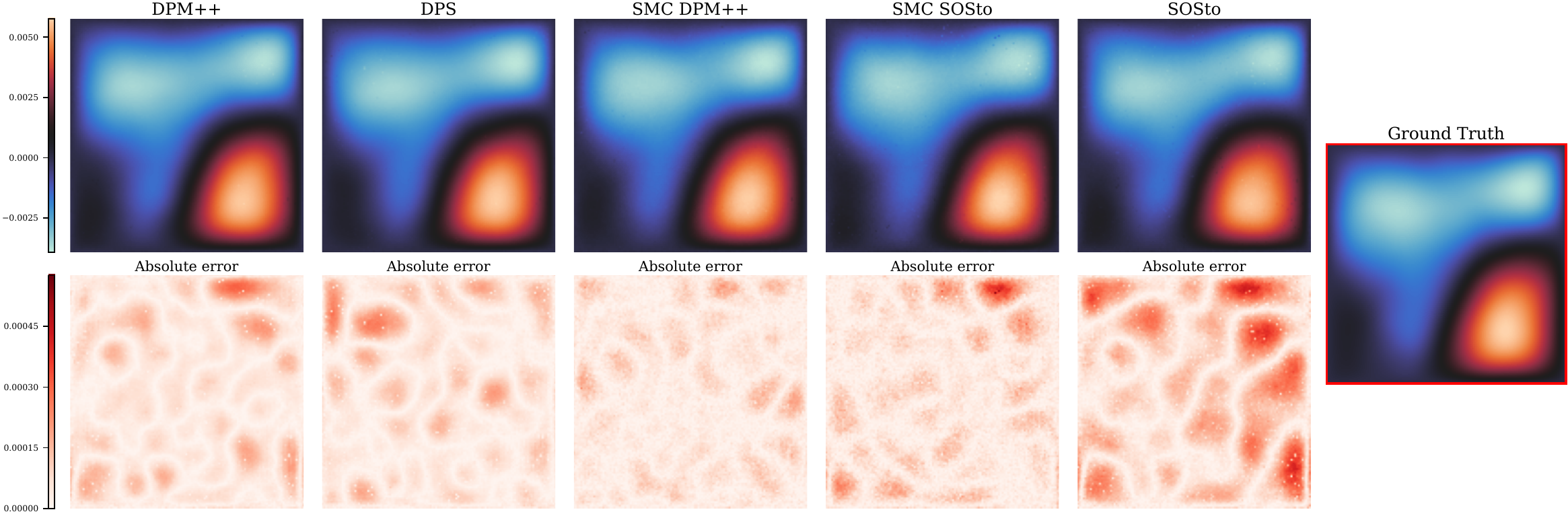}
        \caption{Helmholtz}
    \end{subfigure}
    \vspace{0.5em}
    \begin{subfigure}{0.8\textwidth}
        \includegraphics[width=\textwidth]{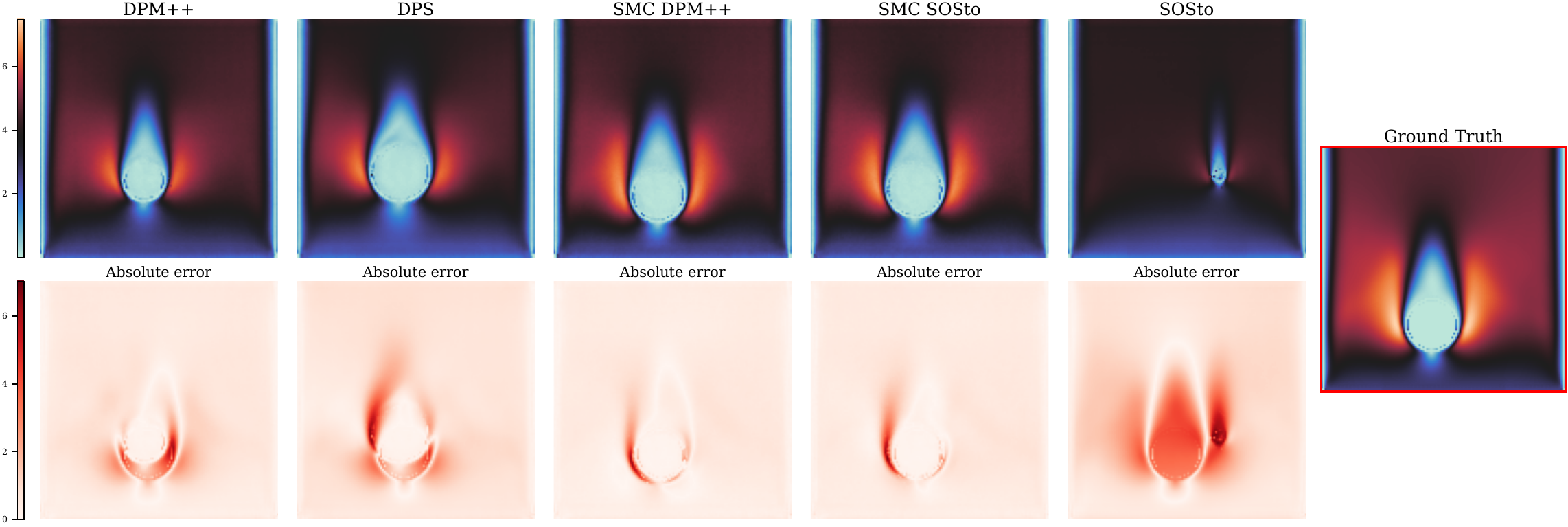}
        \caption{NS-Bounded}
    \end{subfigure}
    \vspace{0.5em}
    \begin{subfigure}{0.8\textwidth}
        \includegraphics[width=\textwidth]{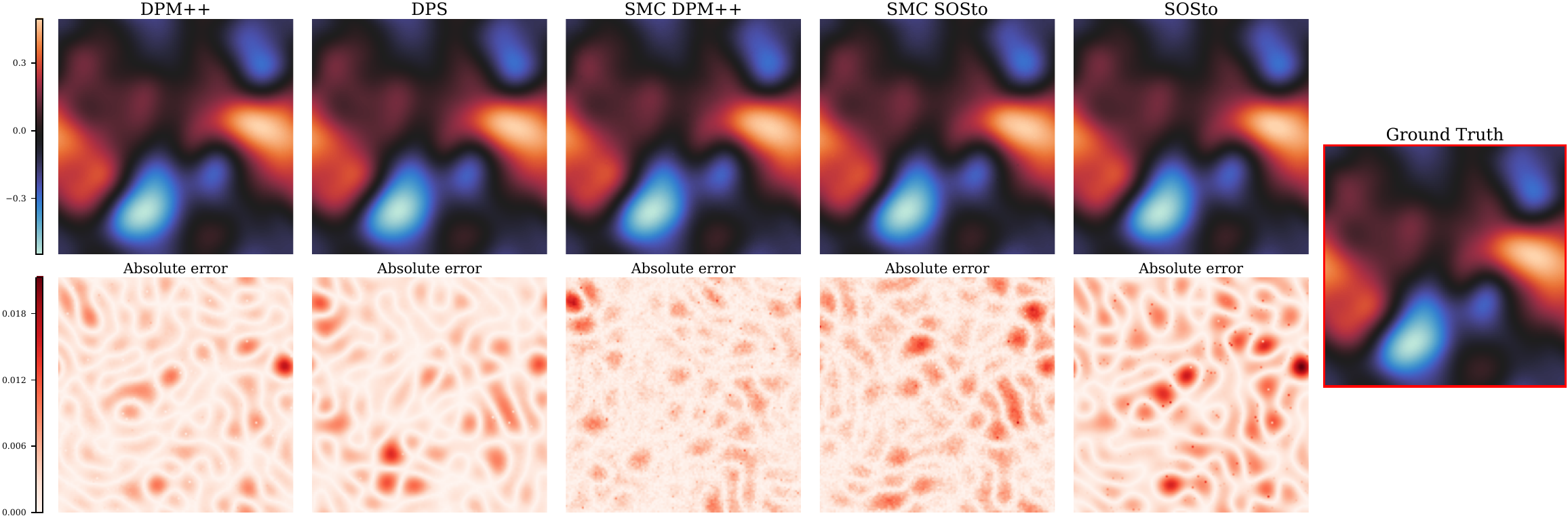}
        \caption{NS-Unbounded}
    \end{subfigure}
    \vspace{0.5em}
    \begin{subfigure}{0.8\textwidth}
        \includegraphics[width=\textwidth]{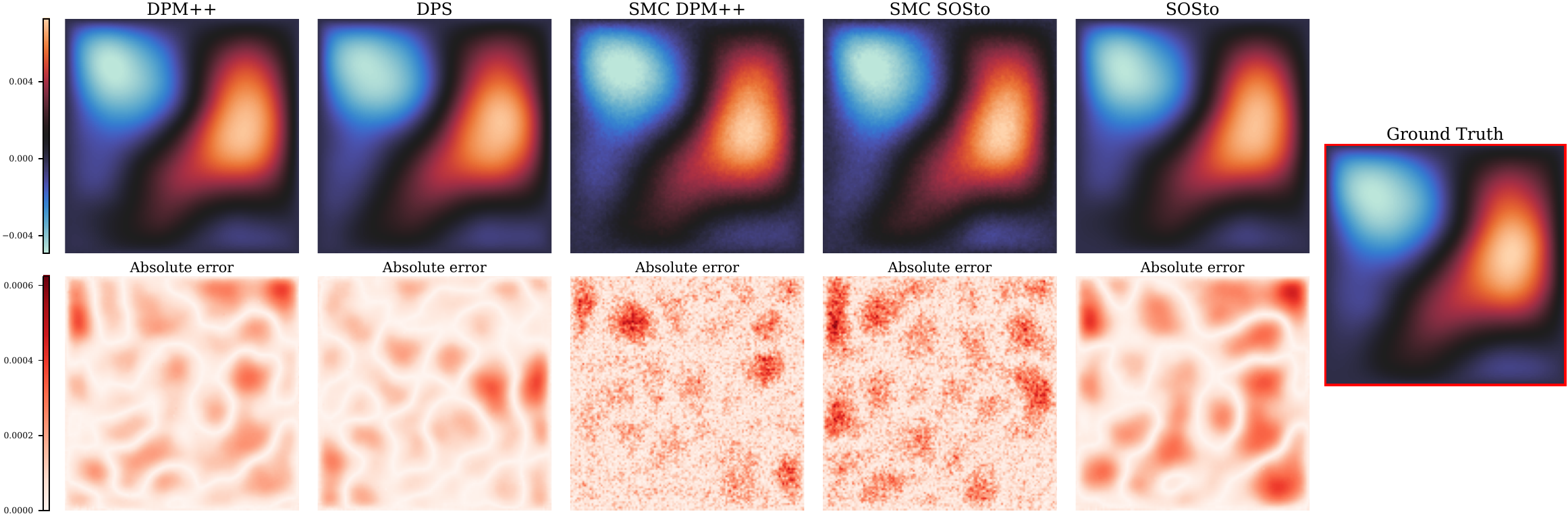}
        \caption{Poisson}
    \end{subfigure}
    \caption{Inverse problem solution field reconstruction $u$. 50 sampling steps.}
    \label{fig:recon_50_ch1}
\end{figure*}

\begin{figure*}[t]
    \centering
    \begin{subfigure}{0.8\textwidth}
        \includegraphics[width=\textwidth]{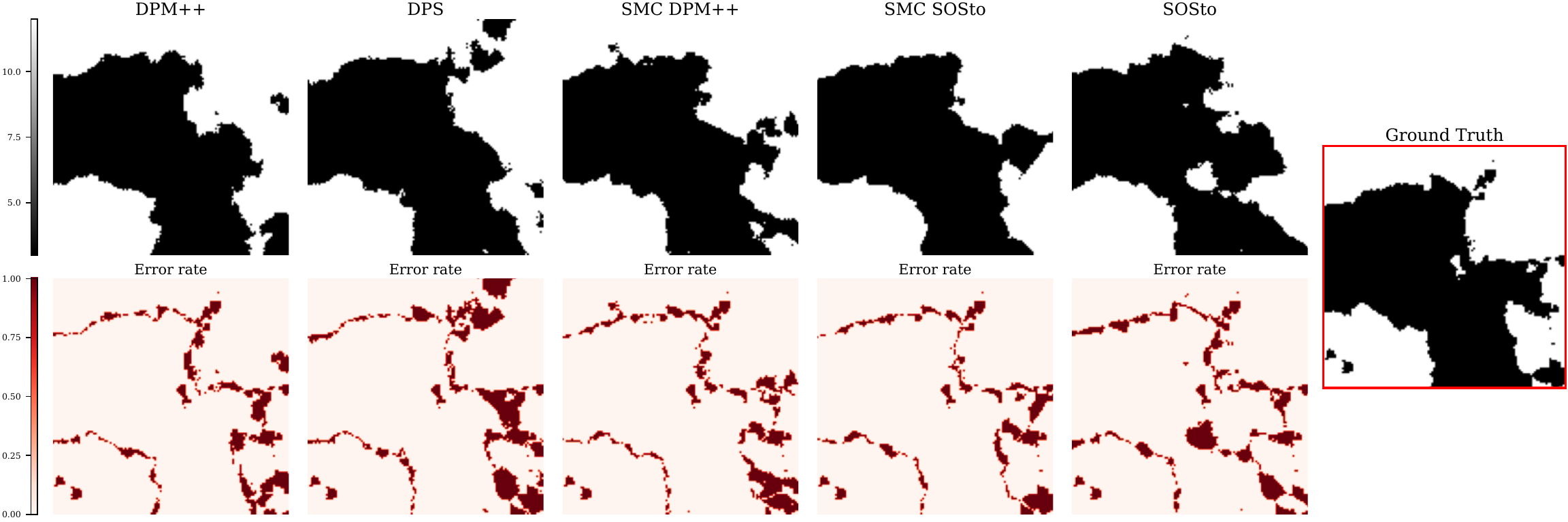}
        \caption{Darcy flow}
    \end{subfigure}
    \vspace{0.5em}
    \begin{subfigure}{0.8\textwidth}
        \includegraphics[width=\textwidth]{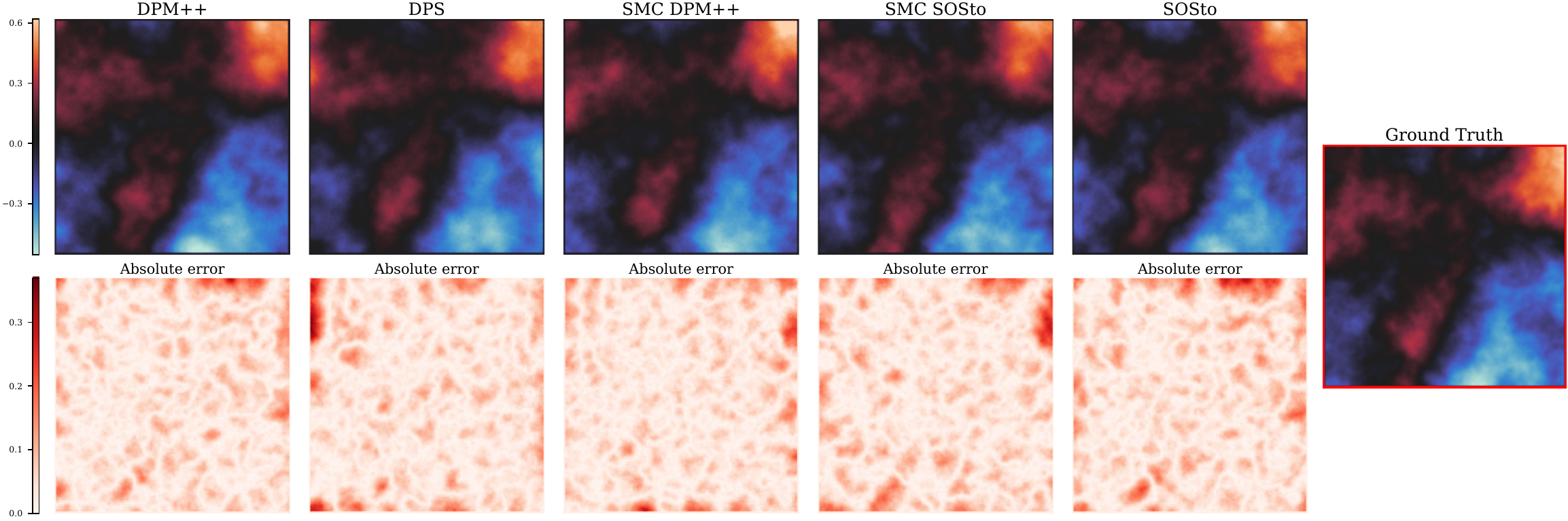}
        \caption{Helmholtz}
    \end{subfigure}
    \vspace{0.5em}
    \begin{subfigure}{0.8\textwidth}
        \includegraphics[width=\textwidth]{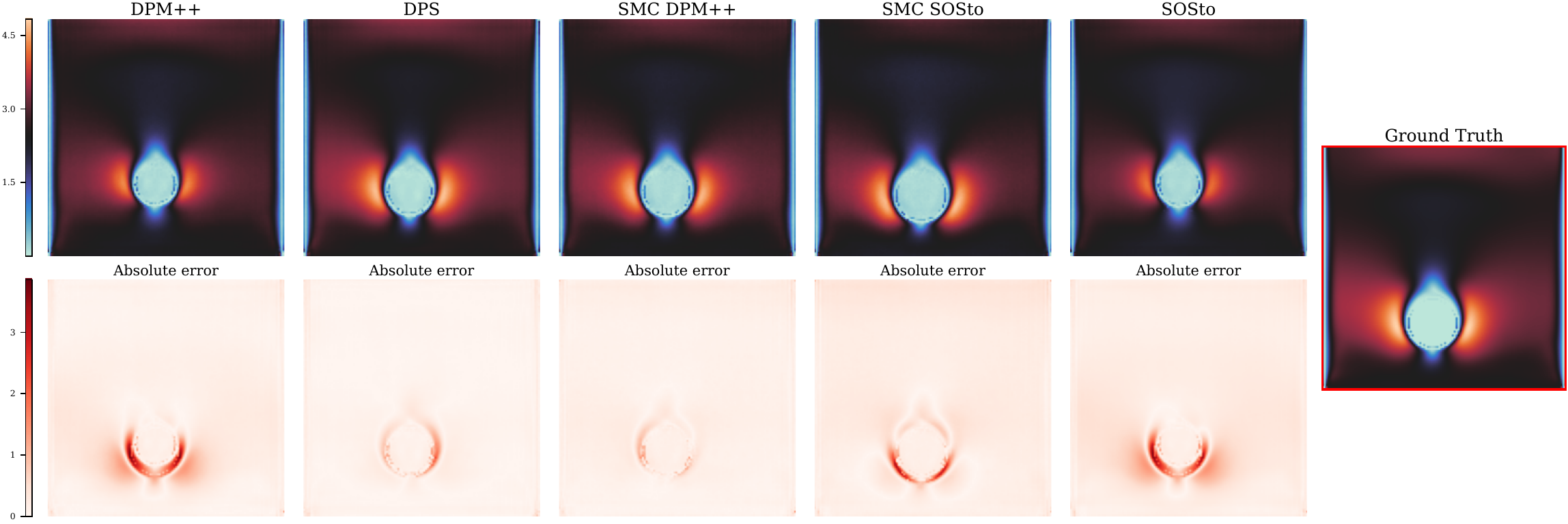}
        \caption{NS-Bounded}
    \end{subfigure}
    \vspace{0.5em}
    \begin{subfigure}{0.8\textwidth}
        \includegraphics[width=\textwidth]{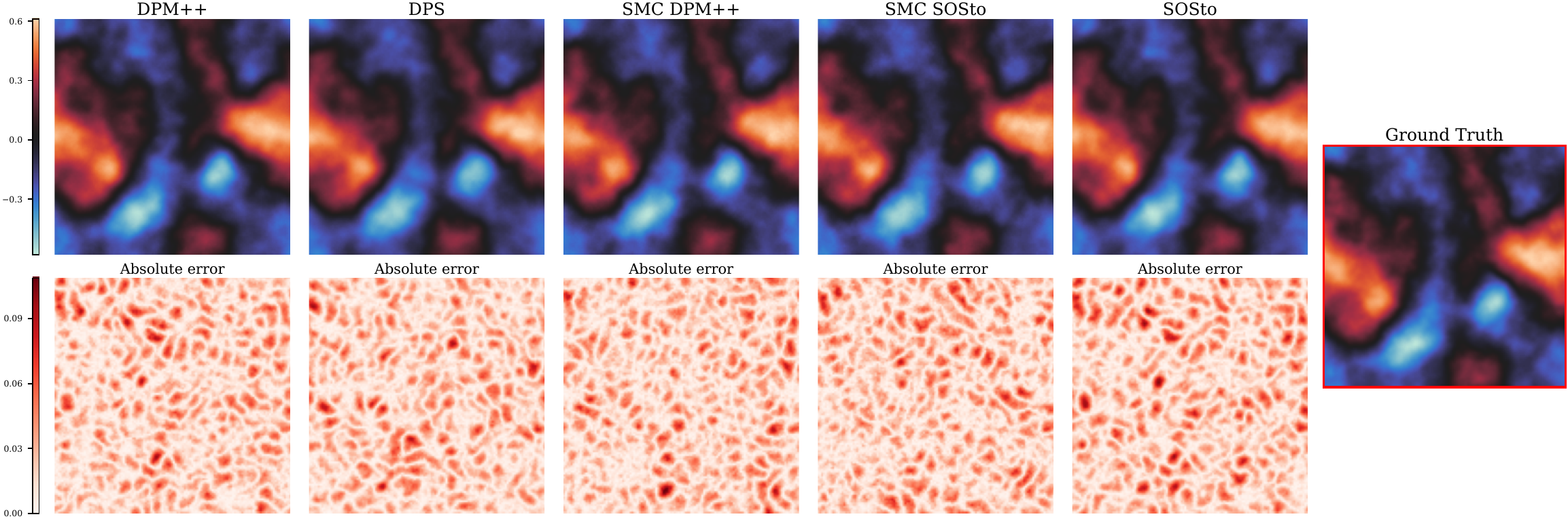}
        \caption{NS-Unbounded}
    \end{subfigure}
    \vspace{0.5em}
    \begin{subfigure}{0.8\textwidth}
        \includegraphics[width=\textwidth]{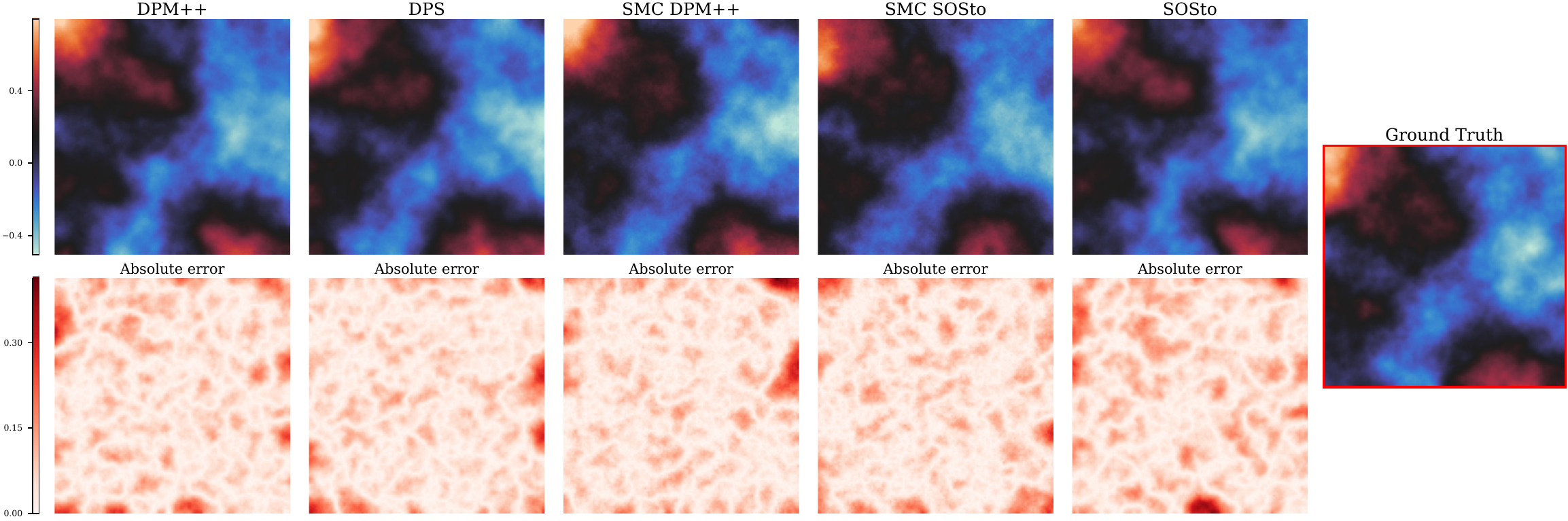}
        \caption{Poisson}
    \end{subfigure}
    \caption{Inverse problem coefficient reconstruction $a$. 100 sampling steps.}
    \label{fig:recon_100_ch0}
\end{figure*}

\begin{figure*}[t]
    \centering
    \begin{subfigure}{0.8\textwidth}
        \includegraphics[width=\textwidth]{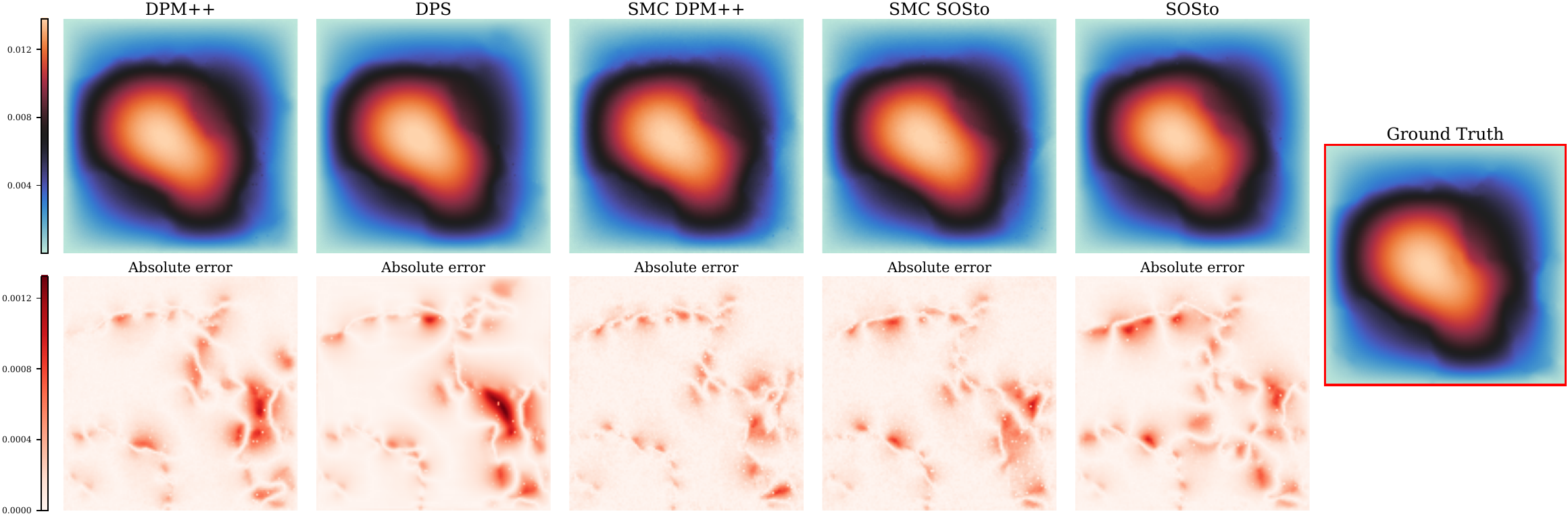}
        \caption{Darcy flow}
    \end{subfigure}
    \vspace{0.5em}
    \begin{subfigure}{0.8\textwidth}
        \includegraphics[width=\textwidth]{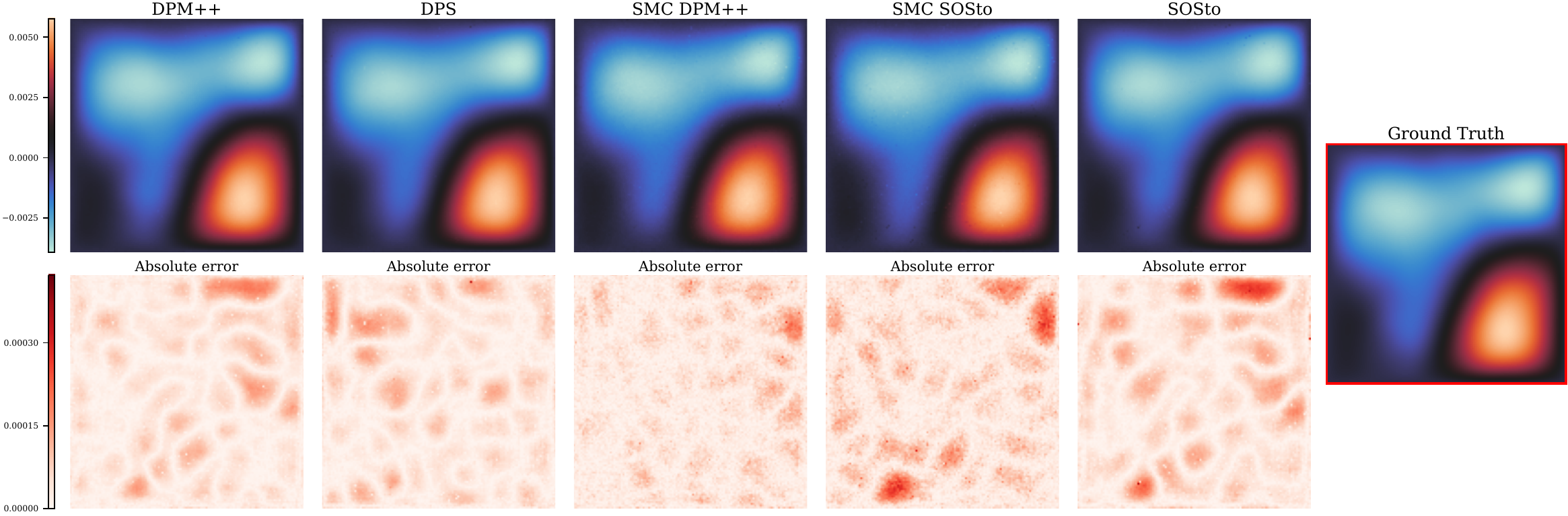}
        \caption{Helmholtz}
    \end{subfigure}
    \vspace{0.5em}
    \begin{subfigure}{0.8\textwidth}
        \includegraphics[width=\textwidth]{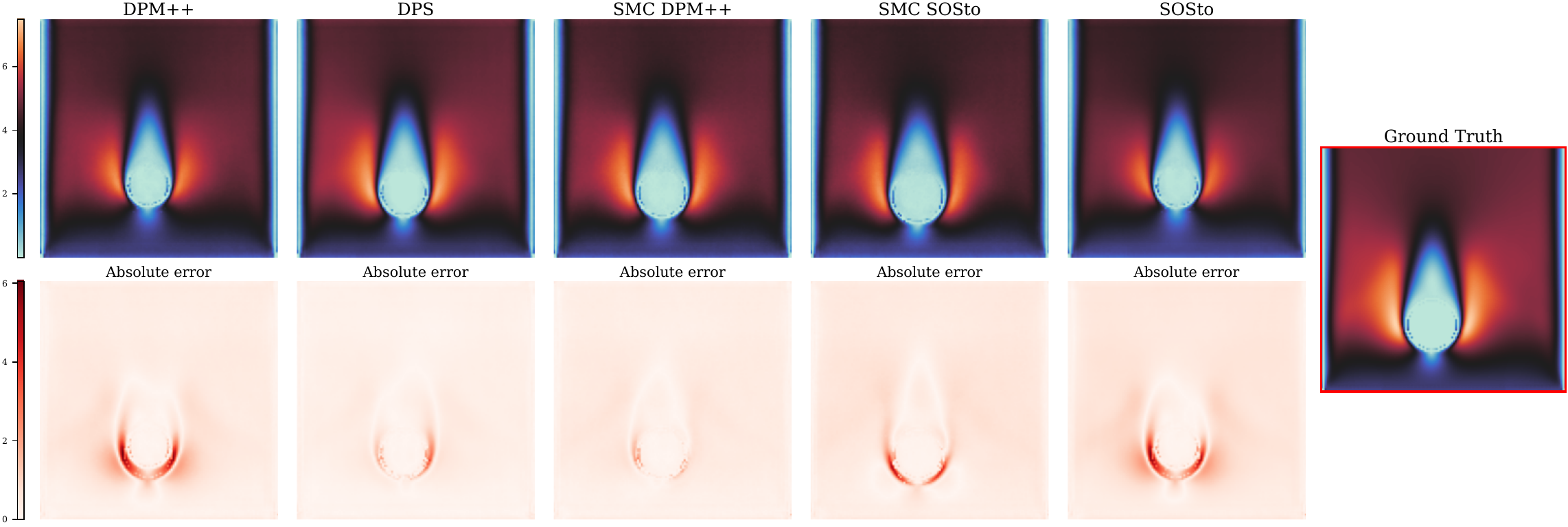}
        \caption{NS-Bounded}
    \end{subfigure}
    \vspace{0.5em}
    \begin{subfigure}{0.8\textwidth}
        \includegraphics[width=\textwidth]{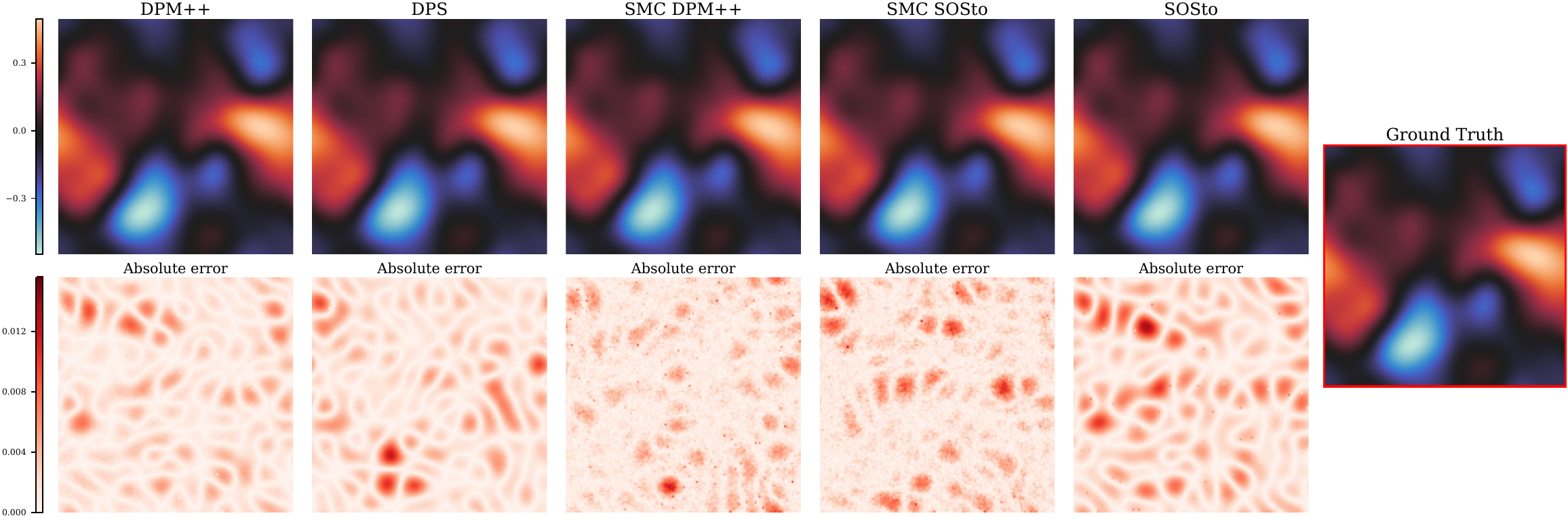}
        \caption{NS-Unbounded}
    \end{subfigure}
    \vspace{0.5em}
    \begin{subfigure}{0.8\textwidth}
        \includegraphics[width=\textwidth]{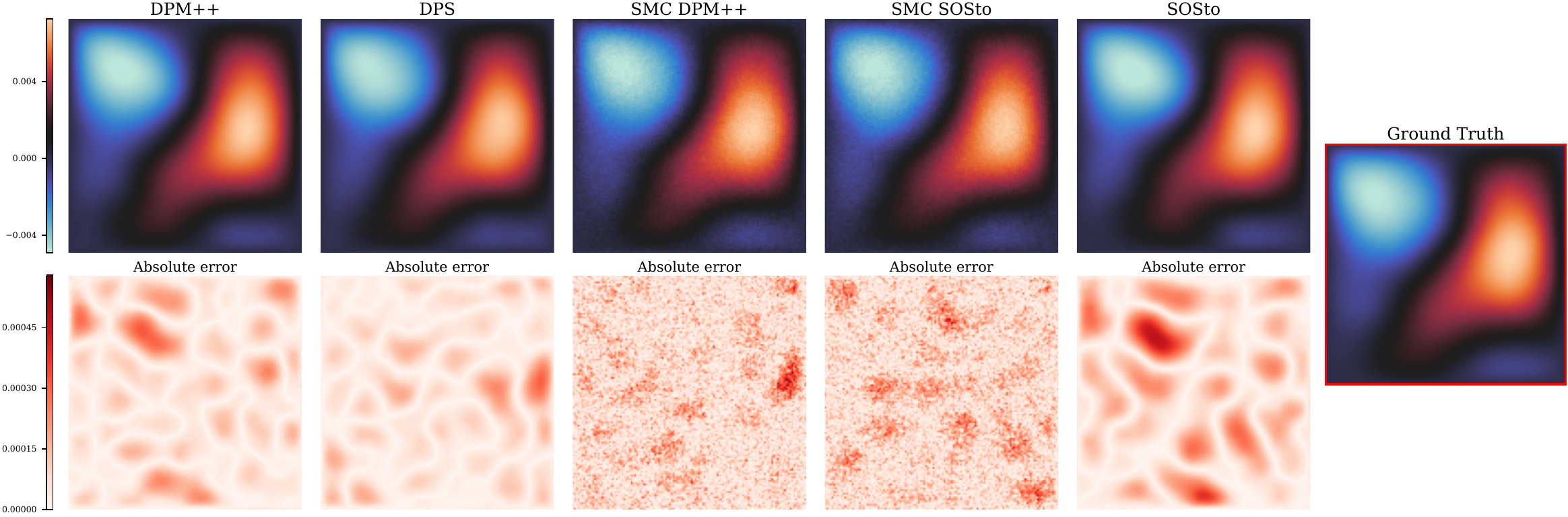}
        \caption{Poisson}
    \end{subfigure}
    \caption{Inverse problem solution field reconstruction $u$. 100 sampling steps.}
    \label{fig:recon_100_ch1}
\end{figure*}

\begin{figure*}[t]
    \centering
    \begin{subfigure}{0.8\textwidth}
        \includegraphics[width=\textwidth]{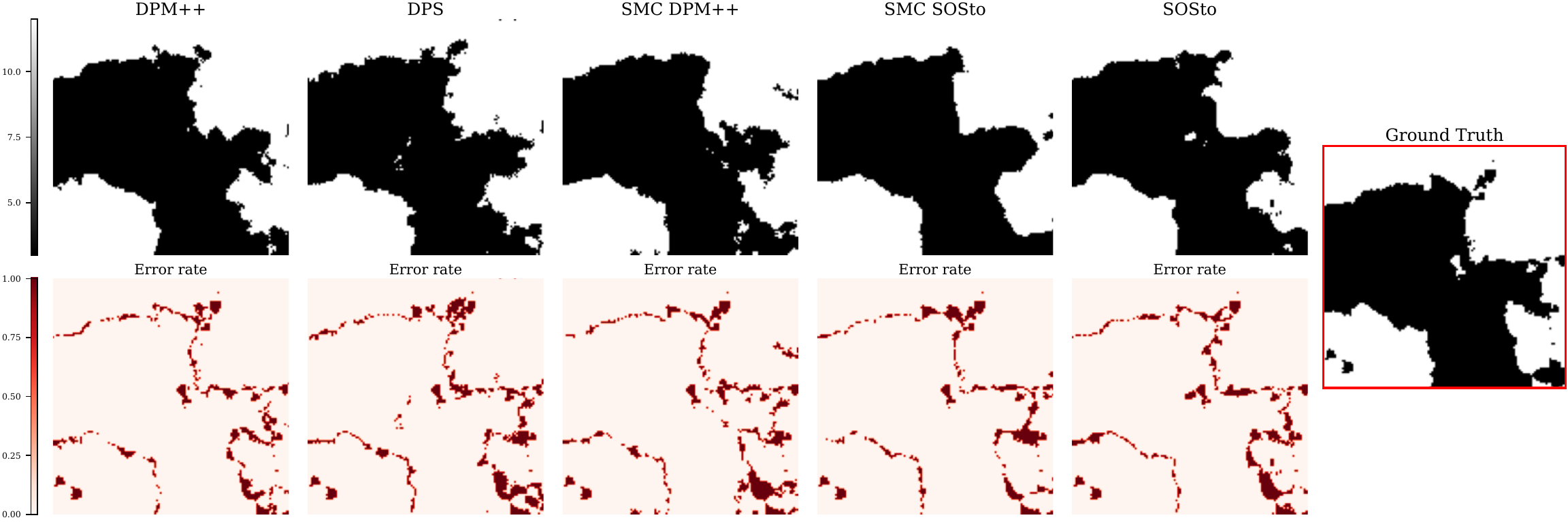}
        \caption{Darcy flow}
    \end{subfigure}
    \vspace{0.5em}
    \begin{subfigure}{0.8\textwidth}
        \includegraphics[width=\textwidth]{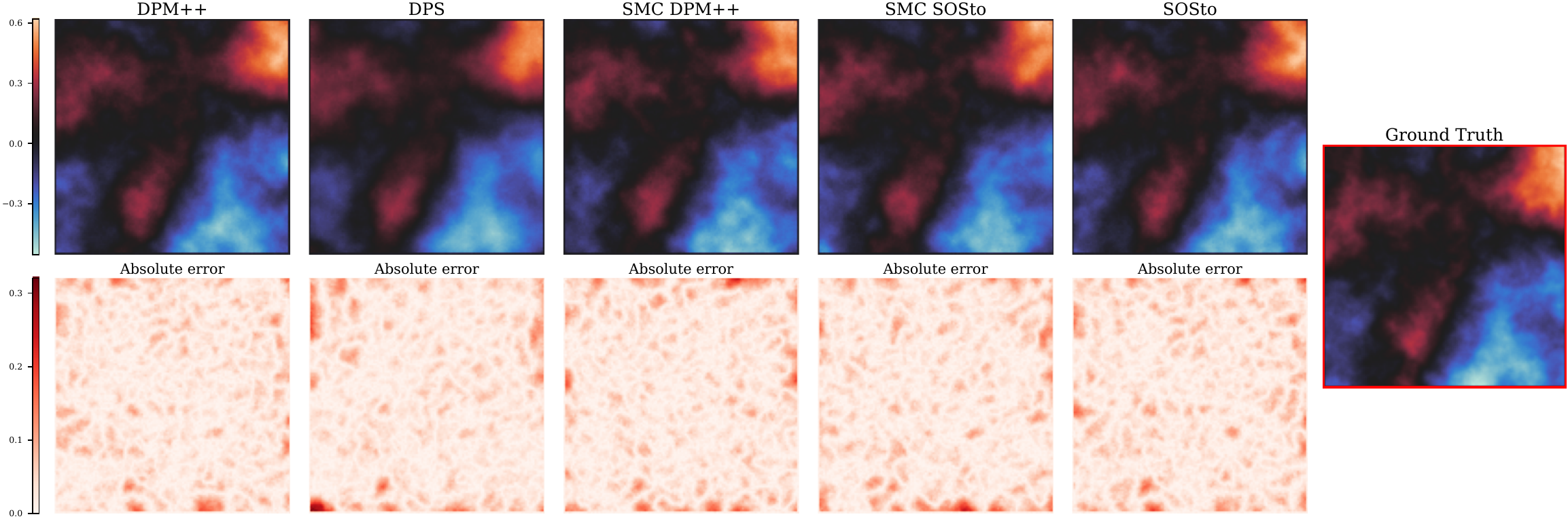}
        \caption{Helmholtz}
    \end{subfigure}
    \vspace{0.5em}
    \begin{subfigure}{0.8\textwidth}
        \includegraphics[width=\textwidth]{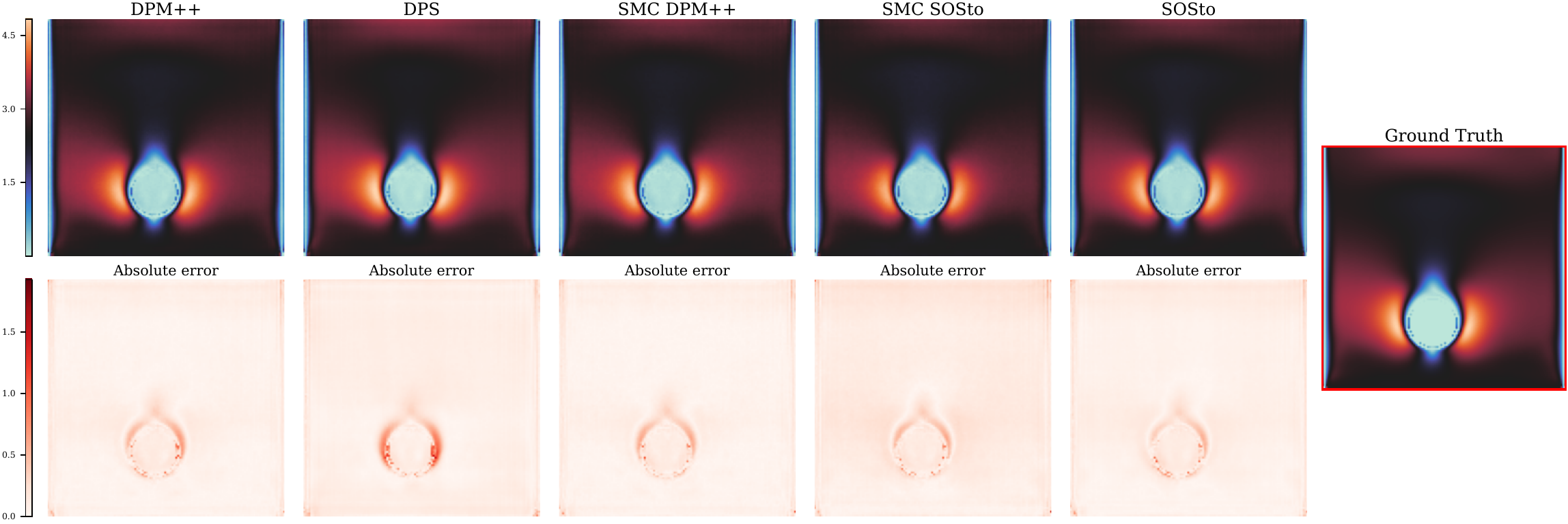}
        \caption{NS-Bounded}
    \end{subfigure}
    \vspace{0.5em}
    \begin{subfigure}{0.8\textwidth}
        \includegraphics[width=\textwidth]{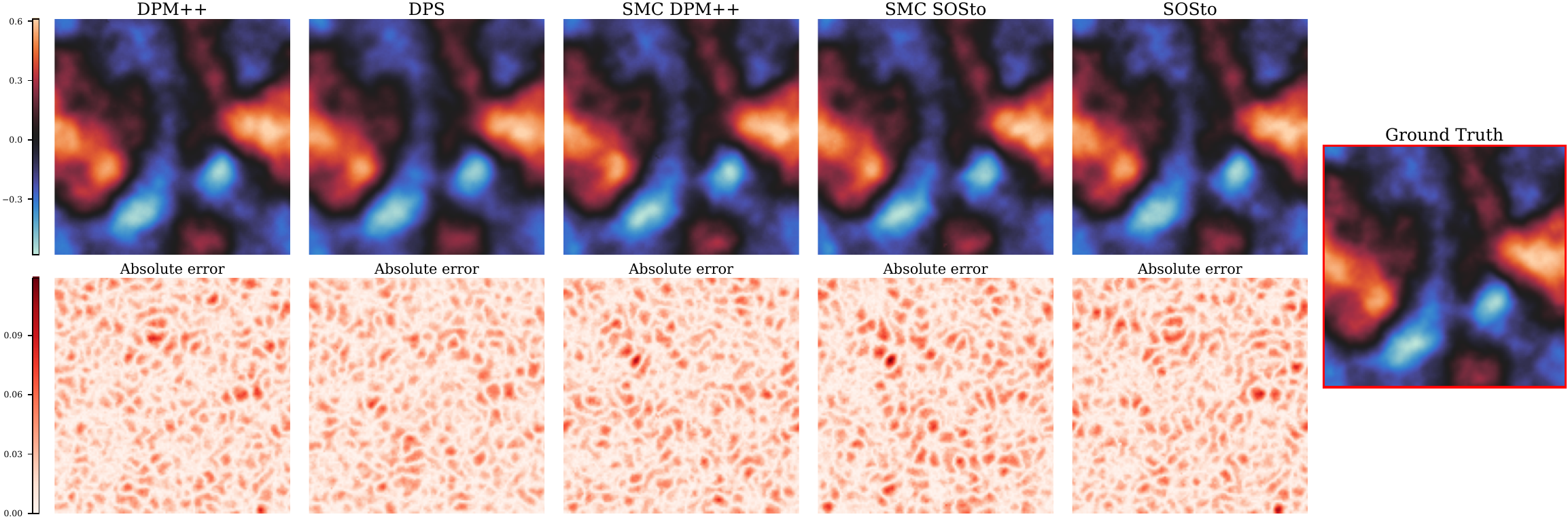}
        \caption{NS-Unbounded}
    \end{subfigure}
    \vspace{0.5em}
    \begin{subfigure}{0.8\textwidth}
        \includegraphics[width=\textwidth]{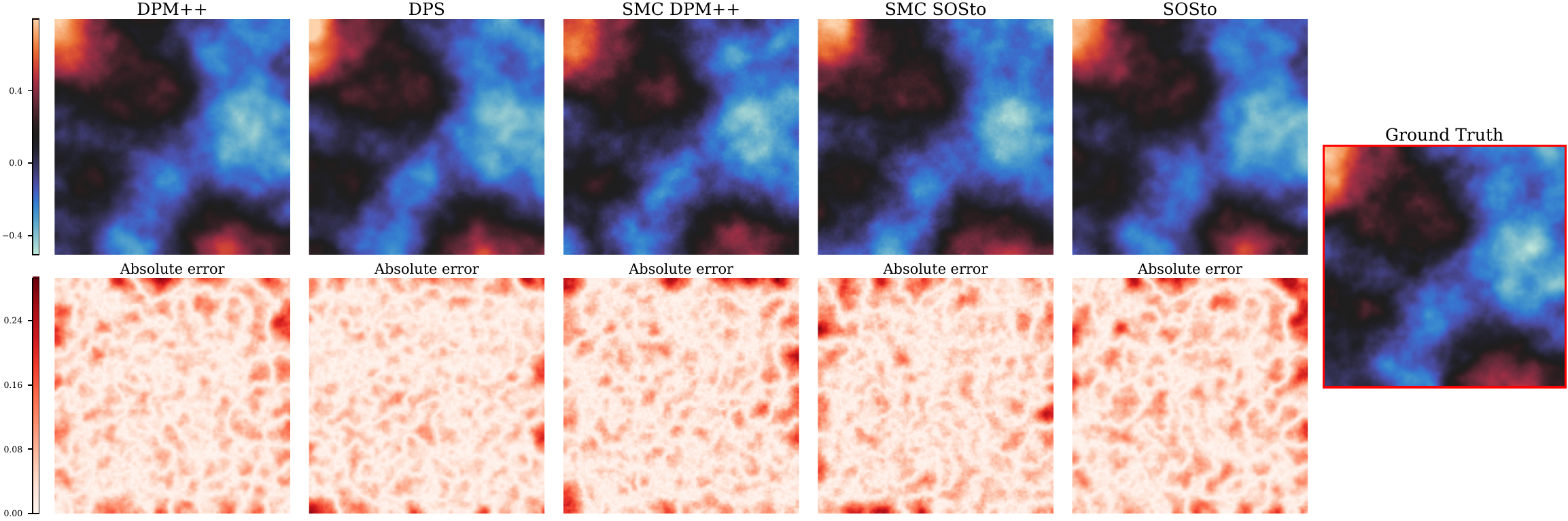}
        \caption{Poisson}
    \end{subfigure}
    \caption{Inverse problem coefficient reconstruction $a$. 500 sampling steps.}
    \label{fig:recon_500_ch0}
\end{figure*}

\begin{figure*}[t]
    \centering
    \begin{subfigure}{0.8\textwidth}
        \includegraphics[width=\textwidth]{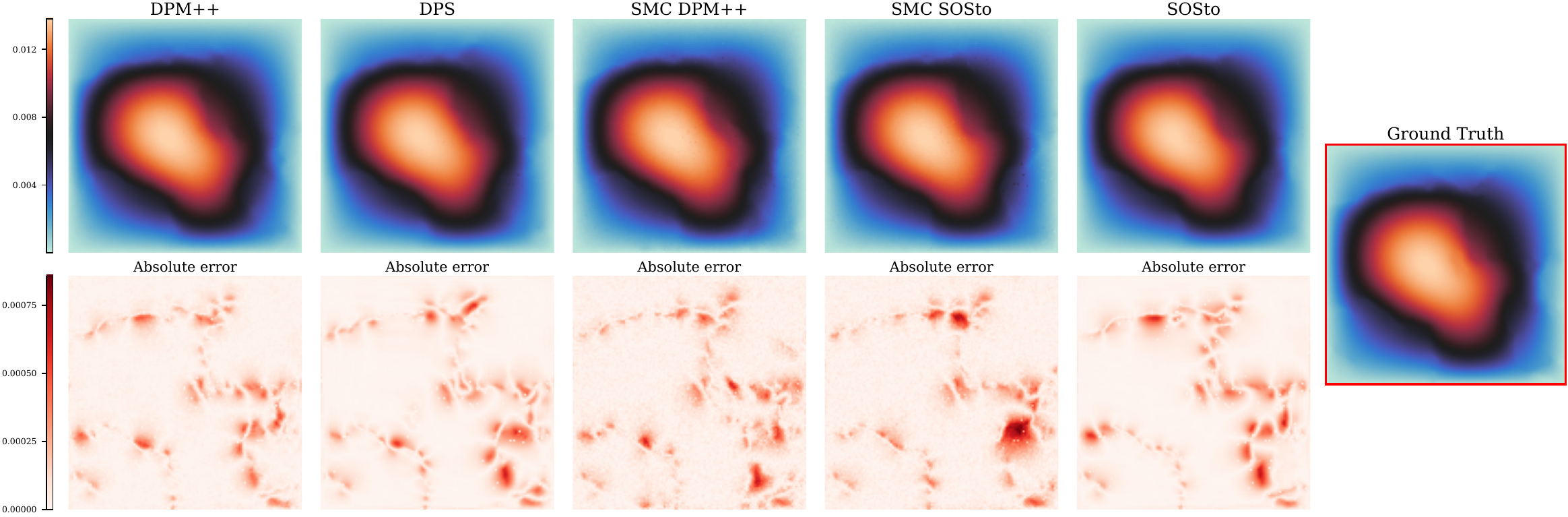}
        \caption{Darcy flow}
    \end{subfigure}
    \vspace{0.5em}
    \begin{subfigure}{0.8\textwidth}
        \includegraphics[width=\textwidth]{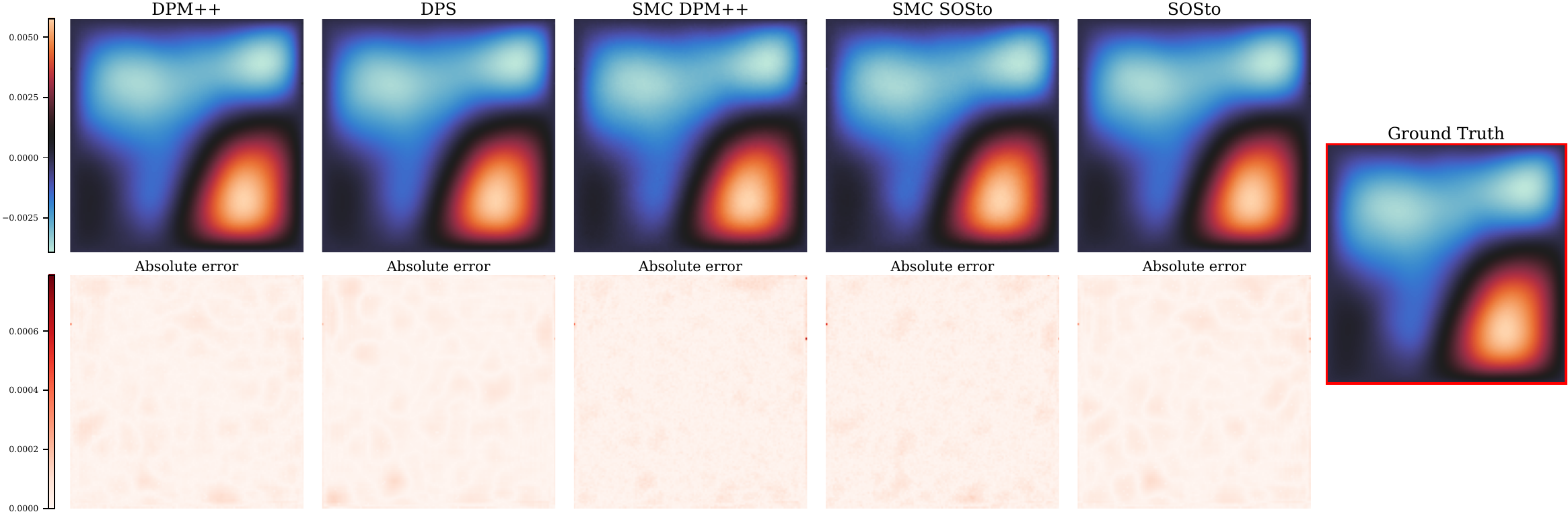}
        \caption{Helmholtz}
    \end{subfigure}
    \vspace{0.5em}
    \begin{subfigure}{0.8\textwidth}
        \includegraphics[width=\textwidth]{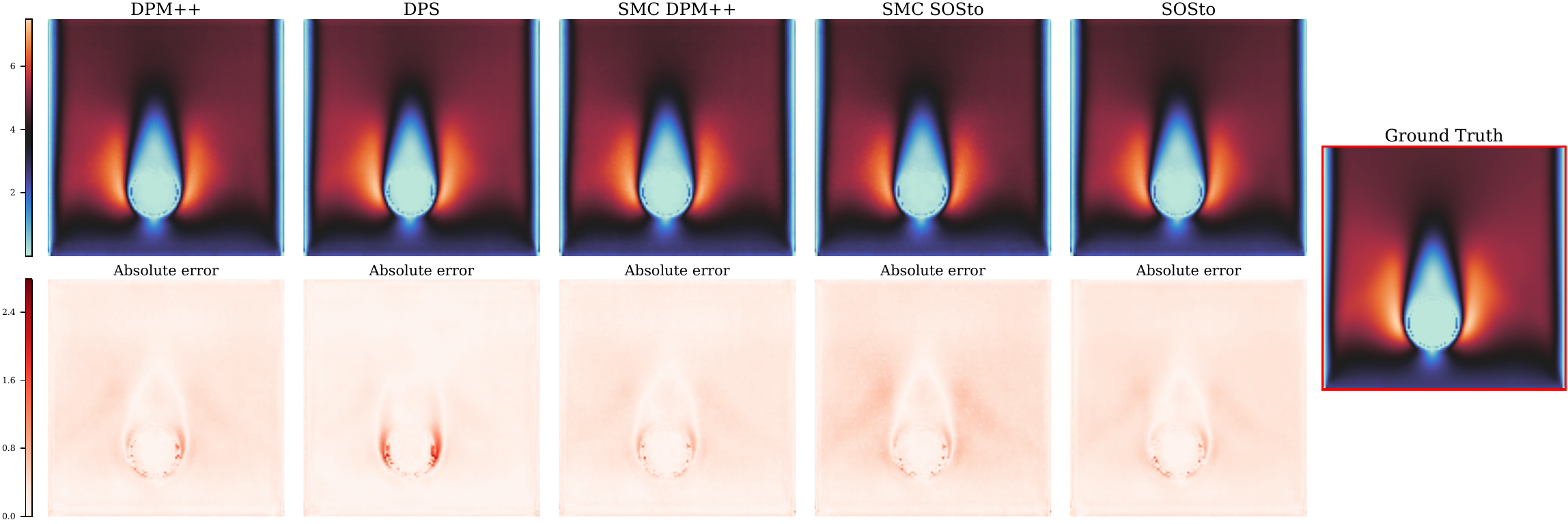}
        \caption{NS-Bounded}
    \end{subfigure}
    \vspace{0.5em}
    \begin{subfigure}{0.8\textwidth}
        \includegraphics[width=\textwidth]{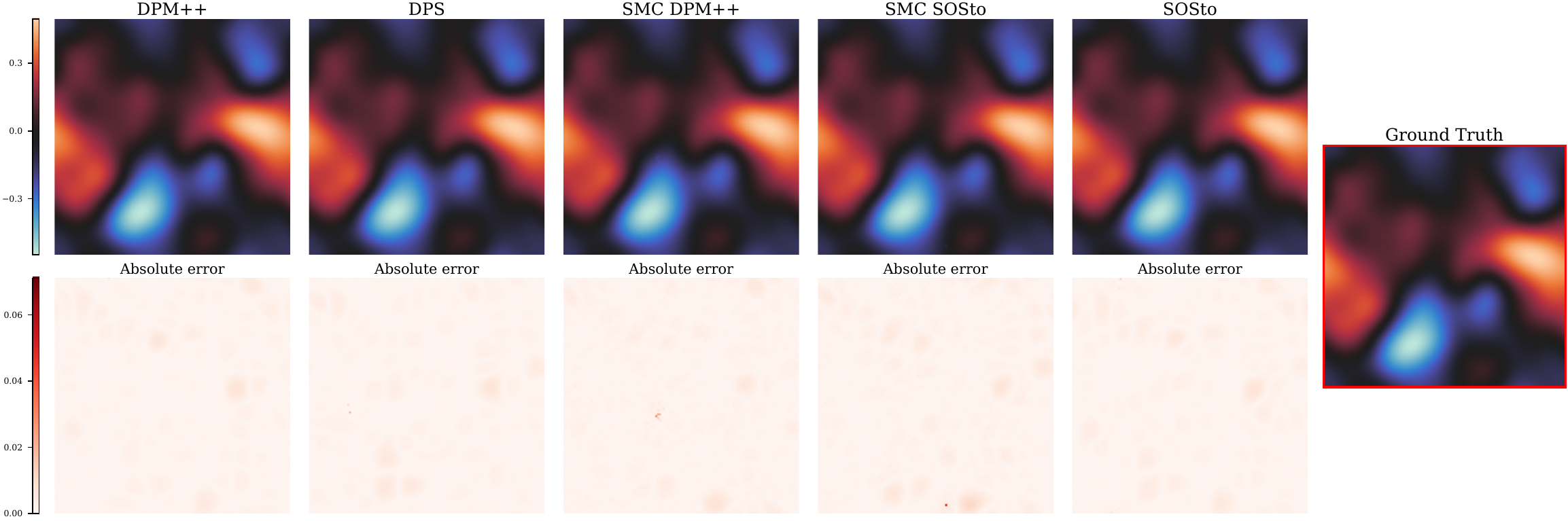}
        \caption{NS-Unbounded}
    \end{subfigure}
    \vspace{0.5em}
    \begin{subfigure}{0.8\textwidth}
        \includegraphics[width=\textwidth]{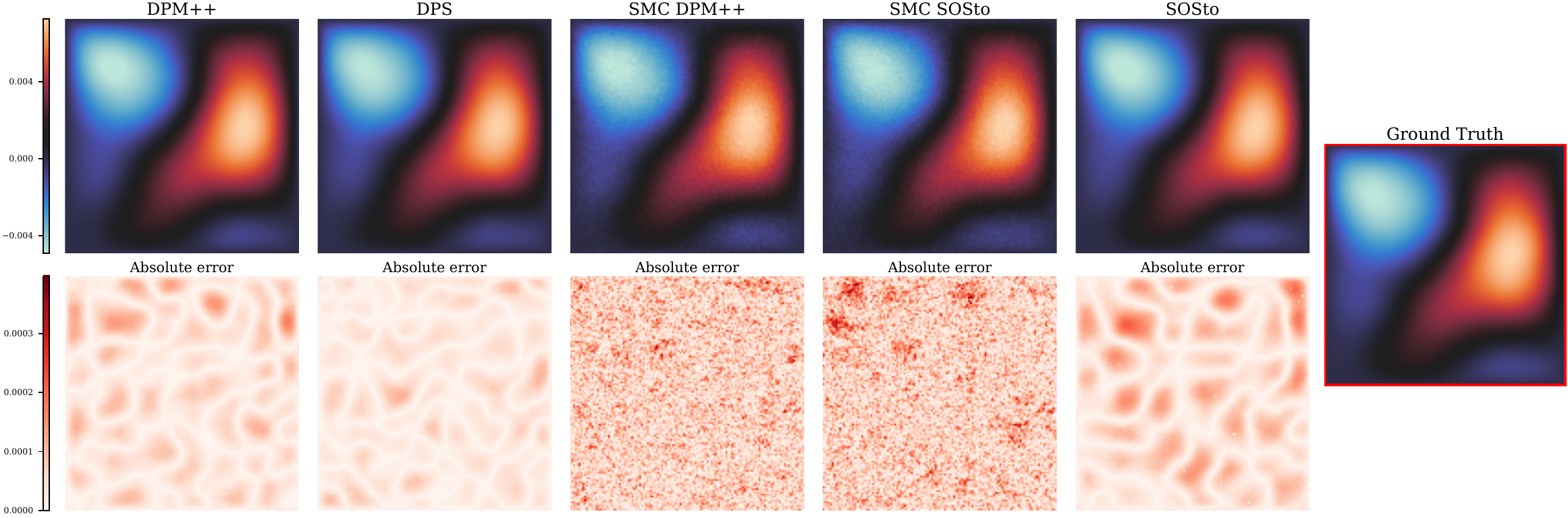}
        \caption{Poisson}
    \end{subfigure}
    \caption{Inverse problem solution field reconstruction $u$. 500 sampling steps.}
    \label{fig:recon_500_ch1}
\end{figure*}

\clearpage




\end{document}